\documentclass{article}

\usepackage[utf8]{inputenc} 
\usepackage[T1]{fontenc}    
\usepackage{hyperref}       
\usepackage{url}            
\usepackage{tabularx}
\usepackage{booktabs}       
\usepackage{amsfonts}       
\usepackage{nicefrac}       
\usepackage{microtype}      
\usepackage{xcolor}         
\usepackage{geometry}
\usepackage{mycommands}
\usepackage{todonotes}
\usepackage{mycommands}
\usepackage{todonotes}
\usepackage{cleveref}
\newcommand{\mc}[1]{\mathcal{#1}}

\newcommand{\bF}{\mathbb{F}}

\newcommand{\peps}{\texttt{PEPS} }

\newcommand{\sample}{\textsf{SAMPLE}}

\newcommand{\eqnmaxlearner}{2C_{3,\ell}^2\sqrt{\log |\X|T_\ell}+\sqrt{2C_{3,\ell}^2T_\ell|\X|\log(T_\ell\ell^2)}+2C_{3,\ell}^2\sum_{t=1}^{T_\ell} \gamma_t}

\newcommand{\ptvsptilde}{\frac{1}{T_\ell}\sum_{t=1}^{T_\ell} \FF_{\theta\sim p_{t}}\left[\norm{\theta-\hat{\theta}_t}_{A(\tilde{\lambda}_t)}^2\right]-\frac{1}{T_\ell}\sum_{t=1}^{T_\ell} \FF_{\theta\sim \tilde{p}_{t}}\bigbrak{\left\|\theta-\hat{\theta}_t\right\|_{A(\tilde{\lambda}_t)}^2}\leq \frac{2C_{3,\ell}^2T_0(\ell)}{T_\ell}}

\newcommand{\martingaleseq}{\frac{1}{T_\ell}\sum_{t=1}^{T_\ell} \FF_{\theta\sim \tilde{p}_{t}}\bigbrak{\left\|\theta-\hat{\theta}_t\right\|_{A(\tilde{\lambda}_t)}^2}-\frac{1}{T_\ell}\sum_{t=1}^{T_\ell} \FF_{\theta\sim \tilde{p}_{t}}\bigbrak{\left\|\theta-\hat{\theta}_t\right\|_{x_tx_t^\T}^2}\leq \sqrt{\frac{2C_{1,\ell}\log \ell^2}{T_\ell}}}

\newcommand{\minlearner}{\sqrt{\frac{C_{3,\ell}^2d\log(T_\ell C_{1,\ell})}{T_\ell}}+C_{3,\ell}\sqrt{\frac{2d\beta(T_\ell,\ell^2)}{T_\ell}\log\bigsmile{\frac{d+T_\ell L^2}{d}}}+C_{3,\ell}\sqrt{\frac{2\log(\ell^2)}{T_\ell}}}

\newcommand{\CTPrime}{\frac{T_2(\ell)L^2\beta(T_2(\ell),\ell^2)}{T_\ell}+4d\beta(T_\ell,\ell^2)T_\ell^{-3/4}}

\newcommand{\CTdoubleprime}{\left|\frac{1}{T_\ell}\inf{\theta\in\Theta_{z_*}^c} \left\|\theta-\hat{\theta}_{T_\ell+1}\right\|_{V_{T_\ell}}^2-\frac{1}{T_\ell}\inf{\theta\in\Theta_{z_*}^c}\norm{\theta^*-\theta}_{V_{T_\ell}}^2\right|\leq (C_{3,\ell}+\Delta_{\max})\sqrt{\frac{\beta(T_\ell,\ell^2)}{T_\ell}}}

\newcommand{\CT}{\frac{1}{T_\ell}\sum_{t=1}^{T_\ell} \FF_{\theta\sim \tilde{p}_{t}}\bigbrak{\left\|\theta-\hat{\theta}_{t+1}\right\|_{V_{t}}^2-\left\|\theta-\hat{\theta}_{t}\right\|_{V_{t}}^2}\leq C_{3,\ell}\sqrt{\frac{2d\beta(T_\ell,\ell^2)}{T_\ell}\log\bigsmile{\frac{d+T_\ell L^2}{d}}}+C_{3,\ell}\sqrt{\frac{2\log(\ell^2)}{T_\ell}}}

\title{Optimal Exploration is no harder than Thompson Sampling}

\author{%
  Zhaoqi Li\footnote{Department of Statistics, University of Washington, \texttt{zli9@uw.edu}}%
  \and Kevin Jamieson\footnote{Allen School of Computer Science \& Engineering, University of Washington, \texttt{jamieson@cs.washington.edu}}%
  \and Lalit Jain\footnote{School of Business, University of Washington,  \texttt{lalitj@uw.edu}}
}

\begin{document}

\maketitle

\begin{abstract}
    Given a set of arms $\mc{Z}\subset \mathbb{R}^d$ and an unknown parameter vector $\theta_\ast\in\mathbb{R}^d$, the pure exploration linear bandit problem aims to return $\arg\max_{z\in \mc{Z}} z^{\top}\theta_{\ast}$, with high probability through noisy measurements of $x^{\top}\theta_{\ast}$ with $x\in \mc{X}\subset \mathbb{R}^d$.  
    Existing (asymptotically) optimal methods require either a) potentially costly projections for each arm $z\in \mathcal{Z}$ or b) explicitly maintaining a subset of $\mc{Z}$ under consideration at each time.  
    This complexity is at odds with the popular and simple Thompson Sampling algorithm for regret minimization, which just requires access to a posterior sampling and argmax oracle, and does not need to enumerate $\mc{Z}$ at any point. 
    Unfortunately, Thompson sampling is known to be sub-optimal for pure exploration. 
    In this work, we pose a natural question: is there an algorithm that can explore optimally and only needs the same computational primitives as Thompson Sampling? We answer the question in the affirmative. We provide an algorithm that leverages only sampling and argmax oracles and achieves an exponential convergence rate, with the exponent being the optimal among all possible allocations asymptotically. In addition, we show that our algorithm can be easily implemented and performs as well empirically as existing asymptotically optimal methods. 
\end{abstract}


\section{Introduction}

The pure exploration bandit problem considers a sequential game between a learner with two sets of arms $\mc{X}, \mc{Z}\subset \mathbb{R}^d$ and nature. In each round, the learner chooses an arm $x\in \mc{X}$ and observes a noisy stochastic reward $y = x^{\top}\theta_{\ast} + \epsilon$ where $\theta_{\ast} \in \Theta$ is an unknown parameter vector and $\epsilon$ is assumed to be i.i.d Gaussian noise. 
The goal of the learner is to identify $z_{\ast} =  \arg\max_{z\in \mc{Z}} z^{\top}\theta_{\ast}$ with high probability in a few measurements.
The case of $\mc{X} = \mc{Z}$ is perhaps the most natural case to consider, and has enjoyed a fair amount of attention \cite{soare2014best,fiez2019sequential,degenne2020gamification}.
However, all proposed approaches share a common trait - complexity. Existing optimal algorithms rely on either explicitly enumerating a potentially large subset of $\ZZ$ or periodically solving a convex optimization program at every iteration. Consequently, it prompts us to question: is such complexity indeed indispensable for reaching asymptotic optimality?
 
Maintaining our focus on the specific instance where $\mc{X} = \mc{Z}$, we note that the pure exploration task can be addressed using any readily available regret minimization algorithm.
That is, if an algorithm generates a series of plays $\{x_t\}_{t=1}^T$ such that $\max_{x \in \mc{X}} \sum_{t=1}^T \langle \theta_*, x - x_t \rangle \leq d \sqrt{T}$ then this immediately implies that $\widehat{x}_T$ drawn uniformly from the set $\{x_t\}_{t=1}^T$ is equal to $x_* = \arg\max_{x \in \mc{X}} \langle x, \theta_* \rangle$ with constant probability as soon as $T \geq d^2 / \Delta_{\min}^2$, where $\Delta_{\min} = \min_{x\in \mc{X}, x\neq x_{\ast}} \theta_{\ast}^{\top}(x^{\ast} - x)$.
One popular regret-minimization algorithm is Thompson Sampling (TS). 
 Following its re-emergence from nearly seven decades of relative obscurity, it has rapidly ascended to become the most prevalently applied bandit algorithm in practical scenarios, as per the industrial experience of the authors.
We postulate that its popularity is due to (1) its simplicity to implement, (2) its flexibility to encode side-information in its prior, (3) its computational efficiency, and (4) strong empirical performance.
The algorithm works by maintaining a distribution $p_t$ over $\Theta$ given all observations up to the time $t$, and then plays $x_t = \arg\max_{x \in \mc{X}} \langle x, \theta_t \rangle$ where $\theta_t \sim p_t$. 
Once $y_t = \langle x_t, \theta_* \rangle + \epsilon_t$ is observed, the distribution is updated and the process repeats. 
As we can see, TS only relies on the ability to sample from a posterior distribution and compute a maximum inner product (an argmax oracle) - both operations which have been heavily studied and optimized.
Unfortunately, TS is known to be sub-optimal for the pure exploration linear bandits problem due to its greedy exploration strategy. 
Indeed, there exist instances of $\mc{X}$ and $\theta_*$ for which the sample complexity of TS to identify the best arm scales \emph{quadratically} in the optimal sample complexity achieved by other algorithms \cite{soare2014best}. 
Even for regret minimization, it is know that TS is far from optimal from an instance-dependent perspective \cite{lattimore2017end}.
But yet, due to its many favorable properties it is still the go-to algorithm in practice. 

This paper aims to answer the following fundamental theoretical question: \emph{Is there an algorithm that enjoys asymptotically optimal exploration that does not need to explicitly enumerate $\mc{Z}$ and only relies on posterior sampling and an argmax oracle?}
We achieve this goal by not striving too far from the Thompson sampling algorithm itself and only assuming access to a sampling oracle and arg-max oracle. 
In fact, our proposed algorithm can be viewed as a generalization of Top-Two Thompson Sampling for the standard multi-armed bandit game \cite{russo2016simple} to the richer linear setting. At each iteration $t$, we maintain a sampling distribution centered at $\widehat{\theta}_t$ (a least squares estimator computed after $t$ samples), and get a sample $\theta_t$ whose best arm is different than that of $\widehat{\theta}_t$ using a sampling oracle.
Once such a $\theta_t$ is found, we update an online learner maintaining a distribution over $\mc{X}$ with rewards $\|\theta_t - \widehat{\theta}_t\|_{x x ^\top}^2$. 
We prove that $\P(\widehat{z}_t \neq z_*)$ decreases at an exponential rate, with an optimal exponent among all possible fixed budget allocations. We also demonstrate that our method is not only theoretically sound by achieving an optimal sample complexity given oracle access, but is also computationally efficient empirically.



\subsection{Problem Setting and Notation}

We first define the linear bandit setting. Let $\X,\ZZ\in\R^d$ be two sets of arms and $\Theta\subset \R^d$ be the parameter space. At time $t$, we draw an action $x_t\in\X$, and receive the reward $y_t=x_{t}^\T\theta_*+\epsilon_t$ where $\theta_*\in\Theta$ and $\epsilon_t$ is i.i.d. Gaussian noise. The choice of arm $x_t$ at time $t$ is dependent on the filtration generated by $\{(x_s, y_s)\}_{s=1}^{t-1}$; furthermore, we denote the conditional probability given this filtration be $\P_{\theta}$.

\textbf{Goal:} We are interested in the best-arm identification task, i.e. we would like to find $z_*:=\arg\max_{z\in\ZZ} z^\T\theta_*$ with high probability, while minimizing the number of measurements taken in $\X$. 

We make the following assumption on the parameters that we will discuss further in Section~\ref{sec:guarantees}. 
\begin{assumption}\label{assump:Theta}
$\Theta$ is closed and bounded, with a non-empty interior. 
\end{assumption}
\begin{assumption}
Assume that $\max_{x}\norm{x}_2\leq L$. 
\end{assumption}
\begin{assumption}
Assume that $\operatorname{span}(\ZZ)\subset \operatorname{span}(\X)$ and the optimal arm $z_*\in\ZZ$ is unique. 
\end{assumption}



\paragraph{Notation.} For any matrix $A\in \mathbb{R}^{d\times d}$, we define the norm $\|x\|^2_A : = x^{\top}A x$. Given a set $\mc{S}$, we define the simplex $\triangle_{\mc{S}} := \{\lambda\in \mathbb{R}_{\geq 0}^{|\mc{S}|}: \sum_{i=1}^{|\mc{S}|} \lambda_i = 1\}$. Finally, given a (multivariate) normal distribution $\mathcal{N}(\theta, \Sigma^{-1})$ on $\mathbb{R}^d$ and some set $\Theta$, we define the truncated normal distribution, denoted as $\operatorname{TN}(\theta, \Sigma^{-1};\Theta)$, to be the normal distribution restricted on $\Theta$. For some $\lambda\in\triangle_\X$, we define $A(\lambda):=\sum_{x\in\X}\lambda_x xx^\T$. We define $\Delta_{\max}:=\max_{x\in\X}\max_{\theta,\theta'\in\Theta}|x^\T(\theta-\theta')|$. We define the constants used in the algorithm as $C_{3,\ell}=\Delta_{\max}+L^2\sqrt{d\log(T_\ell\ell^2)}$. The precise definition is in Appendix~\ref{sec:tableofnotations}.




\section{Motivating Our Approach}\label{sec:posterior}



Among all adaptive algorithms, it is known that for every $\theta_* \in \Theta$ there exists a $\lambda \in \triangle_{\mc{X}}$ such that sampling $x_1,x_2,\dots,\overset{i.i.d.}{\sim} \lambda$  achieves the optimal sample complexity in the fixed confidence setting \cite{soare2014best,fiez2019sequential,degenne2020gamification}.
Specifically, for any $\Theta \subset \R^d$ and $\mc{X},\mc{Z} \subset\R^d$ define
\begin{align}\label{eq:taustar}
    \tau^* := \max_{\lambda \in \triangle_{\mc{X}}} \min_{\theta \in \Theta_{z_*}^c} \tfrac{1}{2}\| \theta - \theta_* \|_{A(\lambda)^{-1}}^2
\end{align}
where $\Theta^c_{z_{\ast}} = \{\theta\in \Theta:\exists z\in \mc{Z}, z^{\top}\theta \geq z_{\ast}^{\top}\theta\}$.
Then it is known that to identify $z_*$ with probability at least $1-\delta$, the expected sample complexity of any algorithm scales as $(\tau^*)^{-1} \log(2.4/\delta)$.
Moreover, sampling according to the $\lambda$ that achieves the maximum, when paired with an appropriate stopping time, achieves the optimal sample complexity asymptotically.
As our setting is more naturally analyzed in the so-called fixed budget setting, we next state a result that can be viewed as a generalization of the result of \cite{russo2016simple} originally stated for the multi-armed bandit setting. Note that this is not a lower bound for the traditional fixed budget setting in multi-armed bandits~\cite{karnin2013almost}, since we only allow fixed $\lambda$ not adapting to the observations.
\begin{theorem}\label{thm:lower_bound}
Fix $\Theta = \R^d$ and any $\theta_* \in \Theta$.
For some $\lambda$ consider a procedure that draws $x_1,\dots,x_T \sim \lambda$, then observes $y_t = \langle x_t, \theta_* \rangle + \epsilon_t$ for each $t$ with $\epsilon_t \sim \mc{N}(0,1)$, and then computes $\widehat{z}_T = \arg\max_{z \in \mc{Z}} \langle z , \widehat{\theta}_T \rangle$ where $\widehat{\theta}_T = \arg\min_{\theta \in \Theta} \sum_{t=1}^T \| y_t - \langle \theta, x_t \rangle \|_{2}^2$.
Then for any $\lambda \in \triangle_{\mc{X}}$ we have
\begin{align*}
    \limsup_{T \rightarrow \infty} -\frac{1}{T}\log\Big( \P_{\theta_*,x_t \sim \lambda}( \widehat{z}_T \neq z_* ) \Big) \leq \tau^*.
\end{align*}
\end{theorem}

The quantity $\tau^{\ast}$ is naturally interpreted from a hypothesis-testing lens. Given a fixed sampling distribution $\lambda$, note that $\E_{x\sim \lambda} KL(\mc{N}(\theta^{\top}x,1)||\mc{N}(\theta_{\ast}^{\top}{x},1)) = \frac{1}{2}\|\theta - \theta_{\ast}\|_{A(\lambda)}^2$. Thus the min-max problem above aims to construct the distribution $\lambda$ which maximizes the smallest KL divergence between $\theta$ and any alternative with a different best-arm. 
As noticed by many authors, this can be translated into a game-theoretic language. The $\max$-player chooses a distribution over the set of possible measurements $\mc{X}$. At the same time, the $\min$-player chooses an alternative $\theta$ whose best arm is not $z_{\ast}$ in an attempt to fool the $\lambda$-player. This lower bound intuitively suggests a strategy for algorithm designers: devise a sampling method that ensures the resultant allocation aligns with the aforementioned objective.


In this pursuit (discussed extensively in Section~\ref{sec:related_works}) the game-theoretic perspective has been directly exploited by several works to give asymptotically optimal algorithms. 
The approaches of these works differ in detail but are similar in spirit and are motivated by the following oracle strategy that has access to $\theta_{\ast}$. At each time, the $\max$-player utilizes a no-regret online learner, such as exponential weights \cite{bubeck2011introduction}, to set $\lambda_{t+1}$ based on an estimate of the best-response of the $\min$-player, namely  $\min_{\theta\in \Theta^c_{z_{\ast}}} \|\theta - \theta_{\ast}\|^2_{A(\lambda_t)}$. 
This guarantees that 
$$\max_{\lambda\in \triangle_{\mc{X}}}\min_{\theta\in \Theta^c_{z_{\ast}}} \|\theta - \theta_{\ast}\|^2_{A(\lambda)} - \sum_{t=1}^{T} \min_{\theta\in \Theta^c_{z_{\ast}}} \|\theta - \theta_{\ast}\|^2_{A(\lambda_t)} \leq o(T)$$
which by a standard Jensen's inequality argument is sufficient to ensure that $\frac{1}{T}\sum_{t=1}^T \lambda_t$ is an approximate solution to the original saddle point problem. 
Then, the arm $x_t$ pulled is sampled from $\lambda_t$ at each time (or a deterministic tracking strategy is used). 


The main computational challenge in this approach is that obtaining the best-response can be rather involved. The alternative set can be decomposed as a union of intersections of a convex set with a halfspace: $\Theta^c_{z_{\ast}} = \cup_{z \neq z_{\ast}} \Theta\cap \{\theta\in \mathbb{R}^d: z^{\top}\theta \geq z_{\ast}^{\top}\theta \}$. Thus computing the best-response involves computing $|\mc{Z}|$-many projections onto convex sets. For small values of $|\mc{Z}|$, this may be feasible. 
However, this computation may be onerous if $|\mc{Z}|$ is large or the projection step is very expensive, for example, in many combinatorial bandit settings such as shortest path problems in a graph \cite{chen2017nearly}.
As another example, in practical recommendation systems where $\mc{Z}$ represents items to be recommended, $|\mc{Z}|$ may be in the millions. Thus computing $|\mc{Z}|$ many projections under latency constraints may be impossible, even though Thompson Sampling can easily recommend good items~\cite{biswas2019seeker}. In addition, for both settings, there may be no easy closed-form expression for the projection. 

Our method is based on the following equivalent formulation of $\tau^{\ast}$. By linearizing the $\min$ over alternatives with a distribution over $\Theta_{z_{\ast}}^c$, we can apply Sion's minimax theorem:
\begin{align*}
    &\max_{\lambda\in\triangle_\X}\inf{\theta\in\Theta_{z_\ast}^c} \frac{1}{2}\norm{\theta-\theta_\ast}_{A(\lambda)}^2 \\
    &= \max_{\lambda\in\triangle_\X}\min_{p\in\triangle(\Theta_{z_\ast}^c)} \E_{\theta\sim p}\left[\frac{1}{2}\norm{\theta-\theta_\ast}_{A(\lambda)}^2\right]\\
    &= \min_{p\in\triangle(\Theta_{z_\ast}^c)}\max_{\lambda\in\triangle_\X} \E_{\theta\sim p}\left[\frac{1}{2}\norm{\theta-\theta_\ast}_{A(\lambda)}^2\right],
\end{align*}
where $\triangle(\Theta_{z_\ast}^c)$ denotes the set of distribution over the alternative set $\Theta_{z_\ast}^c$. 
This replaces the projections with an expectation over a distribution on $\Theta^{c}_{z_\ast}$. 
At first glance, the situation may seem worse - we have gone from finitely many projections to needing to maintain a distribution over a potentially infinite set! 

However, imagine that $\Theta$ is finite and that we solve this saddle-point problem by maintaining a no-regret learner for the $\max$-player as before, while similarly maintaining a no-regret learner for the $\min$-player. 
Standard results in convex optimization guarantee that the average of the iterates of the two learners converge to a saddle point eventually \cite{liu2022initialization}. 
To be more precise, at each round $t$ we draw an $x_t\sim \lambda_t$ and feed the (stochastic) loss $\sum_{\theta\in \Theta^c_{z_{\ast}}} p_{t,\theta}\|\theta - \theta_{\ast}\|_{x_t x_{t}^{\top}}^{2}$ to the learner for the $\min$-player.
Assuming the $\min$-player learner is exponential weights, then the update is
\begin{align*}
    p_{t+1, \theta} \propto p_{t, \theta}e^{-\eta\|\theta_{\ast} - \theta\|^2_{x_tx_t^{\top}}} \propto e^{-\eta\|\theta_{\ast} - \theta\|^2_{\sum_{s=1}^t x_sx_s^{\top}}}.
\end{align*}
where $\eta$ is an appropriate step-size. Hence, the resulting distribution $p_{t+1}$ is reminiscent of the probability density function of a multivariate normal distribution $N(\theta_{\ast}, \eta^{-1}(\sum_{s=1}^t x_s x_s^{\top})^{-1})$ restricted to $ \Theta^c_{z_{\ast}}$. 
This observation motivates our algorithm - for the $\min$-player we maintain an appropriate normal distribution and at each round, use samples from this distribution to generate a stochastic loss to feed the $\max$-player. \emph{This approach avoids explicitly maintaining $\mathcal{Z}$ or ever needing to compute a projection!} Of course, this discussion has relied on knowledge of $\theta_{\ast}$ and $z_{\ast}$. In the next section, we explain how our algorithm, PEPS, overcomes these restrictions.

\section{Best Arm Identification Through Sampling}
\begin{algorithm}[!ht]
\caption{Pure Exploration with Projection-Free Sampling (PEPS)}
\begin{algorithmic}[1]\label{alg:exp_stochastic}
\REQUIRE Finite set of arms $\mathcal{X} \subset \R^d$, $\mathcal{Z} \subset \R^d$, time horizon $T$, $\eta_{\lambda}, \eta_p, \alpha$
\STATE Define $\lambda^G = \arg\min_{\lambda\in\triangle_\X}\max_{x\in\X}\norm{x}_{A(\lambda)^{-1}}^2$, $\lambda_1=\frac{1}{|\X|}\mathbf{1}$
\STATE Initialize $V_0=I$, $S_0=0$, $p_1 = N(0, V_0)$, $\hat{\theta}_1$ arbitrarily 
\FOR{$t=1,2,\cdots,T$}
\STATE $\gamma_t=t^{-\alpha}$
\STATE \textcolor{blue}{//\texttt{Top Two Sampling}}
\STATE Compute $\hat{z}_{t}=\argmax{z\in\ZZ}\,z^\top \hat{\theta}_{t}$ 
\STATE Sample $\theta_t=\textsf{SAMPLE}(\operatorname{TN}(\hat{\theta}_t,\eta_p^{-1}V_{t-1}^{-1};\Theta_{\hat{z}_t}^c))$
\STATE 
\STATE \textcolor{blue}{//\texttt{Take Sample and Observe Reward}}
\STATE Sample $x_t\sim\tilde{\lambda}_t$ where $\tilde{\lambda}_{t}=(1-\gamma_t)\lambda_{t}+\gamma_t\lambda^G$
\STATE Observe $y_t = \langle \theta_*, x_t \rangle + \epsilon_t$ where $\epsilon_t \sim \mathcal{N}(0,1)$ 
\STATE

\STATE \textcolor{blue}{//\texttt{Update }}
\STATE Update $V_{t}=V_{t-1}+x_tx_t^\T$, $S_t = S_{t-1} + x_t y_t$, and $\hat{\theta}_{t+1}=V_t^{-1} S_t$  
\STATE Update $\lambda_{t+1} \propto \lambda_te^{\eta_\lambda \tilde{g}_{t}}$ where $\tilde{g}_{t,x}=\norm{\theta_t-\hat{\theta}_t}_{xx^\T}^2, \forall x\in \mc{X}$ \label{line:lambda-learner}


\ENDFOR
\STATE Sample $\tilde{\theta}=\textsf{SAMPLE}( \operatorname{TN}(\hat{\theta}_{T+1}, V_T^{-1};\Theta))$
\ENSURE $\arg\max_{z\in \mc{Z}} z^{\top}\tilde\theta$

\end{algorithmic}
\end{algorithm}

\begin{algorithm}
\caption{Doubling trick}
\begin{algorithmic}[1]\label{alg:doubling_trick}
\REQUIRE Finite set of arms $\X\subset \R^d$, $\ZZ\subset \R^d$
\FOR{$l=0,1,\cdots$}
\STATE Set $T_\ell=2^\ell$, 
$\eta_\lambda=\sqrt{\frac{\log |\X|}{C_{3,\ell}^2T_\ell}}$, $\eta_p=\sqrt{\frac{d\log(T_\ell C_{3,\ell})}{C_{3,\ell}^2T_\ell}}$, $\alpha=1/4$
\STATE  $\hat{z}_{\ell} = \operatorname{PEPS}(\X,\ZZ,T_\ell,\eta_{\lambda},\eta_p,\alpha)$ 
\ENDFOR
\ENSURE 
\end{algorithmic}
\end{algorithm}


Our main method \peps is presented in Algorithm~\ref{alg:exp_stochastic}. 
Given a budget of $T$ samples, we repeatedly sample $\theta_t$ utilizing a sampling oracle \sample.
We then sample an $x_t\sim\tilde{\lambda}_t$ where $\tilde{\lambda}_t$ is the distribution $\lambda_t$ maintained by the $\lambda$-learner at time $t$ mixed in with a diminishing amount $\gamma_t$ of the $G$-optimal distribution $\lambda^G$.
After playing $x_t$ and observing a reward $y_t$, \peps updates both the $\lambda_t$ and the estimate $\hat{\theta}_t$ with the covariance. In particular, given samples $\{x_s\}_{s=1}^t$, we let $\hat{\theta}_{t+1}=V_{t}^{-1}S_{t}$ where $V_t=\sum_{s=1}^t x_sx_s^\T$ and $S_t=\sum_{s=1}^t x_sy_s$. 
Algorithm~\ref{alg:exp_stochastic} depends on a finite time horizon $T$. To ensure that our algorithm is anytime and eventually converges to the optimal sampling scheme, we employ an outer loop Algorithm~\ref{alg:doubling_trick} utilizing a doubling scheme. Before we explain the theoretical guarantees, we first detail some of the aspects of the algorithm.

\paragraph{Updating the sampling distribution for $\theta_t$.} Our main innovation is introducing a distribution over $\Theta_{\hat{z}_t}^c$ from which we can sample over. 
In particular, in each round, we sample $\theta_t$ from $\operatorname{TN}(\hat{\theta}_{t},\eta_p^{-1} V_{t-1}^{-1};\Theta_{\hat{z}_t}^c)$, which is a \emph{truncated normal distribution} with support $\Theta^c_{\hat{z}_t}$ \cite{burkardt2014truncated}.

Following the discussion in the Section~\ref{sec:posterior}, it is tempting to see this update as a form of continuous exponential weights \cite{bubeck2011introduction}. However, this is not quite true since the underlying action set $\Theta^c_{\hat{z}_t}$ is changing each round. This creates several technical challenges in the proof. 
Note that similar to previous works, we could have maintained a learner for each $z\in \mc{Z}$ \cite{degenne2020gamification}. However, our approach of maintaining a distribution prevents the need for this additional complexity of enumerating $\mc{Z}$.

From the perspective of exponential weights, $\eta_p$ is a step size: the dependence on $d$ in the numerator comes from the dimension of $\Theta$; and $C_{3,\ell}^2$ is an upper bound on the stochastic loss $\|\theta_{t} - \hat{\theta}_t\|^2_{x_t x_t^{\top}}$ that we guarantee with high probability due to forced exploration and boundedness of $\Theta$. 






We have the following regret guarantee on the online min learner. For notational convenience, in this section, for some set $\S$ with nonempty interior, we let $p_t(\S)=\operatorname{TN}(\hat{\theta}_t,\eta_p^{-1}V_{t-1}^{-1};\S)$ be the truncated normal distribution with support on $\S$. 

\begin{lemma}[informal]\label{lem:p_learner}
In round $T_{\ell}$ of epoch $\ell$ of Algorithm~\ref{alg:doubling_trick}, we have with probability greater than $1-1/\ell^2$,  
\begin{align*}
    &\sum_{t=1}^{T_\ell} \E_{\theta\sim p_{t}(\Theta_{z_\ast}^c)}\bigbrak{\left\|\theta-\hat{\theta}_t\right\|_{x_tx_t^\T}^2}- \inf{\theta\in\Theta_{z_*}^c} \left\|\theta-\theta_{\ast}\right\|_{V_{T_{\ell}}}^2\leq O(d\sqrt{T_\ell}\log(LT_\ell)).
\end{align*}    
\end{lemma}

\paragraph{Sampling Oracle.} Our algorithm involves a sampling oracle that takes samples from a truncated normal distribution. 

\begin{definition}[Sampling oracle (\textsf{SAMPLE})]
The oracle $\textsf{SAMPLE}(p)$ is an algorithm that given some distribution $p$, returns a sample $\theta\sim p$. 
\end{definition}

There are various ways to implement this sampling oracle efficiently. The easiest way is to use rejection sampling. In particular, on line 7, for each round $t$, we repeatedly sample $\theta_t\sim N(\hat{\theta}_{t},\eta_p^{-1}V_{t-1}^{-1})$ until the best-arm of $\arg\max_{z\in \mc{Z}} z^{\top}\theta_t$ is not our current best guess $\hat{z}_t = \arg\max_{z\in \mc{Z}}z^{\top}\hat{\theta}_t$, and on line 17 we repeatedly sample $\tilde{\theta}\sim N(\hat{\theta}_{T+1},V_T^{-1})$ until $\tilde\theta\in\Theta$. Regarding the computation cost of rejection sampling, we suffer from some of the same challenges as Top-two sampling algorithms, which empirically work well in practice \cite{russo2016simple}. From a practical perspective, the rejection sampling step is only computationally costly if it requires many draws from the posterior to find a $\theta$ in the alternative $\Theta_{\hat{z}_t}^c$. However, note that if we draw $O(1/\nu)$ vectors and none of them are in the alternative $\Theta_{\hat{z}_t}^c$, by Markov's inequality, this arm they all agree on is the best arm with probability $1-\nu$. Thus, as soon as it becomes computationally costly to sample an alternative, the problem is basically solved. We demonstrate empirically that the computational complexity is not at all onerous in Section~\ref{sec:experiments} and Appendix~\ref{sec:supp_plots}. Also, we note that our focus is on the query complexity given an effective way to sample, not the complexity of sampling from the distribution itself. Since the sampling oracle only returns one sample at the end, our algorithm still achieves an asymptotically optimal \textit{sample complexity} even if we draw $O(1/\nu)$ vectors inside the oracle. 

Moreover, we remark that sampling from truncated normal distributions is a well-explored practice across statistics and machine learning, especially when sampling in a convex set. A variety of efficient methods such as Gibbs and hit-and-run procedures are available for this purpose \cite{Devr86, murphy2013machine, li2015efficient,laddha2023convergence}. In particular, the hit-and-run algorithm ensures one gets a sample in the convex set with probability $1-\nu$ in $O(d^3\log(1/\nu))$ samples in the worst case \cite{lovasz1999hit}. Furthermore, novel approaches have improved the efficiency of traditional rejection techniques, especially when dealing with a convex support of the truncated normal distribution \cite{maatouk2016new}.

\paragraph{Update for $\lambda_t$.} To update $\lambda_t$, which corresponds to the action of our $\max$-player,  we employ an exponential weighted learner (Hedge) over the set of actions $\mc{X}$. The reward vector $\tilde{g}_t \in \mathbb{R}^{|\mc{X}|}$ is stochastic with expectation $\E \tilde{g}_{t,x} = \E_{\theta\sim p_t(\Theta_{\hat{z}_t}^c)}\norm{\theta-\hat{\theta}_t}_{xx^{\top}}^2$  conditioning on the history of the algorithm $\{(x_s, y_s, \theta_s)\}_{s=1}^{t-1}$, and is bounded in high probability. 
We show that if we choose $\alpha=\frac{1}{4}$ and let $\tilde{\Delta}_{\max}$ be an upper bound on the loss function, we have the following regret guarantee:

\medskip

\begin{lemma}[informal]\label{lem:l_learner}
In round $T_{\ell}$ of epoch $\ell$ of Algorithm~\ref{alg:doubling_trick}, we have with probability greater than $1-1/\ell^2$, 
\begin{align*}
    &\max_{\lambda\in\triangle_\X}\sum_{t=1}^{T_\ell} \E_{\theta\sim p_{t}(\Theta_{\hat{z}_t}^c)}\norm{\theta-\hat{\theta}_t}_{A(\lambda)}^2-\sum_{t=1}^{T_\ell} \E_{\theta\sim p_{t}(\Theta_{\hat{z}_t}^c)}\norm{\theta-\hat{\theta}_t}_{A(\lambda_t)}^2\leq O\left(\sqrt{(d +\tilde{\Delta}_{\max})T_\ell\log \ell}\right).
\end{align*}
\end{lemma}

\paragraph{Forced Exploration with $\mathbf{G}$-optimal Design.}
To ensure adequate sampling in all directions, in each round we mix in some amount of the $G$-optimal distribution, denoted as $\lambda^G := \arg\min_{\lambda\in \Delta_{\mc{X}}}\max_{x\in \mc{X}} \|x\|^2_{A(\lambda)^{-1}}$.
This ensures that $\max_{x\in\X} \| \widehat{\theta}_t - \theta \|_{x x^\top}$ is bounded for any $\theta\in \Theta$ and $\hat{z}_t$ is eventually $z_{\ast}$ with probability 1. 
The rate at which the mixture of this distribution decays as $t^{-\alpha}$, for any $0 < \alpha < 1/2$, so it has no effect on asymptotic performance. 
We note that thanks to the implicit anti-concentration properties of sampling $\theta_t$ from a multivariate Gaussian, this step is probably unnecessary and just an artifact of the analysis~\cite{agrawal2017near}.


\paragraph{Argmax Oracle} One advantage of our approach that is most reminiscient of Thompson Sampling is the calculation of $\hat{z}_{t}$ at the start of each epoch. In practice, if we have an efficient $\arg\max$-oracle, this calculation can be computationally efficient and does not require maintaining $\mc{Z}$. 
By exploiting $\arg\max$ oracles, we can tractably solve problems like shortest-path and matchings, even in settings where $|\mc{Z}|$ is super-exponential in $d$ \cite{katz2020empirical}.

\paragraph{Doubling Trick} 
As presented, the regret guarantees for Lemmas~\ref{lem:p_learner} and \ref{lem:l_learner} require fixed step sizes $\eta_{\lambda}, \eta_p$. To overcome this need for a fixed step size, we use a doubling trick and restart the algorithm every $2^\ell$ samples \cite{shalev2012online}.  We believe the use of the doubling trick is purely a theoretical restriction and a more careful analysis could provide an anytime algorithm with no restarts.

\subsection{Theoretical Guarantees}\label{sec:guarantees}

Our main result is the following guarantee on Algorithm~\ref{alg:doubling_trick}.
\begin{theorem}\label{thm:PEPS}
    With probability $1$, 
    \begin{align*}
        \lim_{\ell\rightarrow \infty} -\frac{1}{T_\ell}\log\P_{\theta \sim \pi_{\ell}}(\hat{z}_{\ell}\neq z_{\ast}) = \tau^{\ast},
    \end{align*}
    where $\pi_{\ell}:= N(\hat{\theta}_{T_{\ell}}, V_{T_{\ell}-1}^{-1})$ restricted to $\Theta$.
\end{theorem}

Thus our algorithm guarantees that asymptotically the probability that we do not identify the optimal arm decays at the rate of $e^{-T \tau^{\ast}}$, with $\tau^*$ being the optimal exponent as given in Theorem~\ref{thm:lower_bound}. 
Such guarantees on the probability of a sampled arm are similar to those in the Bayesian best-arm literature, namely \cite{russo2016simple, jourdan2022top}. 
In these works, a posterior distribution is maintained and they guarantee that the posterior probability that a non-optimal arm is sampled converges at an exponential rate, with the best possible exponent among all allocation rules. 
We provide a similar guarantee here for linear bandits.
We provide a small sketch of the proof now. A full proof is in the Appendix. 

\begin{proof}[Proof sketch]
We say that $a_n\doteq b_n$ if $\frac{1}{n}\log(a_n/b_n)\to 0$ as $n\to\infty$. We focus on a fixed round $\ell$ of Algorithm~\ref{alg:doubling_trick}. Using the fact that the expectation of the empirical log-likelihood ratio (conditioned on the data collected) between $\theta_{\ast}$ and some $\theta\in \Theta$ is the KL divergence between them, we can show using a Laplace Approximation 
\[\P_{\theta \sim \pi_{\ell}}(\hat{z}_{\ell}\neq z_{\ast})\doteq  \exp\bigsmile{-T_\ell\inf{\theta\in\Theta_{z_\ast}^c}\frac{1}{2}\norm{\theta-\theta_\ast}_{A(\overline{e}_{T_\ell})}^2}.\]
where $\overline{e}_{T_\ell}=\frac{1}{T_\ell}\sum_{t=1}^{T_\ell} e_{x_t}$. Letting $\overline{p}_{T_\ell}=\frac{1}{T_\ell}\sum_{t=1}^{T_\ell} p_{t}(\Theta_{\hat{z}_t}^c)$, we have 
\begin{align*}
    &\max_{\lambda\in\triangle_\X}\E_{\theta\sim \overline{p}_{T_\ell}}\norm{\hat{\theta}_{t}-\theta}_{A(\lambda)}^2-\min_{p\in\triangle(\Theta_{z_*}^c)}\E_{\theta\sim p}\norm{\hat{\theta}_{t}-\theta}_{A(\overline{e}_{T_\ell})}^2\\
    &= \max_{\lambda\in\triangle_\X}\frac{1}{T_\ell}\sum_{t=1}^{T_\ell}\E_{\theta\sim p_{t}(\Theta_{\hat{z}_t}^c)}\norm{\hat{\theta}_{t}-\theta}_{A(\lambda)}^2-\frac{1}{T_\ell}\sum_{t=1}^{T_\ell} \E_{\theta\sim p_{t}(\Theta_{\hat{z}_t}^c)}\norm{\theta-\hat{\theta}_{t}}_{A(\lambda_t)}^2\tag{regret for $\max$ learner}\\
    &\ \ +\frac{1}{T_\ell}\sum_{t=1}^{T_\ell} \E_{\theta\sim p_{t}(\Theta_{\hat{z}_t}^c)}\norm{\theta-\hat{\theta}_{t}}_{A(\lambda_t)}^2-\frac{1}{T_\ell}\sum_{t=1}^{T_\ell} \E_{\theta\sim p_{t}(\Theta_{z_*}^c)}\left\|\theta-\hat{\theta}_{t}\right\|_{x_tx_t^\T}^2\tag{error when $\hat{z}_t\neq z_{\ast}$}\\
    &\ \ +\frac{1}{T_\ell}\sum_{t=1}^{T_\ell} \E_{\theta\sim p_{t}(\Theta_{z_*}^c)}\left\|\theta-\hat{\theta}_{t}\right\|_{x_tx_t^\T}^2-\frac{1}{T_\ell}\inf{\theta\in\Theta_{z_*}^c} \left\|\theta-\theta_{\ast}\right\|_{V_{T_\ell}}^2.\tag{regret for the $\min$ learner}
\end{align*}
The regret guarantees in Lemmas~\ref{lem:p_learner} and \ref{lem:l_learner} ensure the first and third sum are $o(1)$ and so go to 0 as $T_\ell\rightarrow \infty$. The fact that $p_{t}(\Theta_{\hat{z}_t}^c)$ is equal to $p_{t}(\Theta_{z_*}^c)$ for large enough $t$ ensures that the middle term similarly goes to 0. Combining all terms and the fact that $\hat{\theta}_t$ is close to $\theta_\ast$ guarantees that for any $\epsilon >0$ there is a sufficiently large $\ell$ such that $\max_{\lambda\in\triangle_\X}\E_{\theta\sim \overline{p}_{T_\ell}}\norm{\theta_*-\theta}_{A(\lambda)}^2-\min_{p\in\triangle(\Theta_{z_*}^c)}\E_{\theta\sim p}\norm{\theta_*-\theta}_{A(\overline{e}_{T_\ell})}^2 \leq \epsilon$,
which using minimax duality implies that $$\inf{\theta\in\Theta_{z_*}^c}\frac{1}{2}\norm{\theta-\theta_\ast}_{A(\overline{e}_{T_\ell})}^2 \geq \max_{\lambda\in\triangle_\X}\min_{p\in\triangle(\Theta_{z_*}^c)}\E_{\theta\sim p}\left[\norm{\theta_*-\theta}_{A(\lambda)}^2\right] - \epsilon.$$ 
Since the first term on the right-hand side is $\tau^*$, we have shown that $\inf{\theta\in\Theta_{z_*}^c}\frac{1}{2}\norm{\theta-\theta_*}_{A(\overline{e}_{T_\ell})}^2\geq \tau^* - \epsilon$. Since by definition $\tau^*\geq \inf{\theta\in\Theta_{z_*}^c}\frac{1}{2}\norm{\theta-\theta_*}_{A(\overline{e}_{T_\ell})}^2$, choosing $\epsilon\to 0$ concludes the proof that
$\P_{\theta \sim \pi_{\ell}}(\hat{z}_{\ell}\neq z_{\ast})\doteq  \exp\bigsmile{-T_\ell\inf{\theta\in\Theta_{z_\ast}^c}\frac{1}{2}\norm{\theta-\theta_\ast}_{A(\overline{e}_{T_\ell})}^2}=\exp(-T_\ell \tau^*)$.
\end{proof}


\paragraph{Remark: Stopping times.}
Note that we are not providing a guarantee on the expected stopping time for any finite $\delta$. Existing asymptotically optimal approaches which guarantee a finite stopping time in high probability, e.g. \cite{degenne2020gamification}, utilize a generalized log-likelihood-ratio test of the form
\[\max_{z\in \mc{Z}} \min_{\theta\in \Theta_{\hat{z}_t}^c} \|\theta - \hat{\theta}_t\|_{V_t} \geq \beta(t,\delta)\]
where $\beta(t, \delta) = O(\sqrt{d\log((T+\|\theta_{\ast}\|_2)/\delta)})$ is an anytime confidence bound controlling the deviations of $\|\theta - \hat{\theta}_t\|_{V_t}$ \cite{abbasi2011improved}. As a result, their algorithms saturate the lower bound for an expected stopping time,  i.e. $\lim\sup_{\delta\rightarrow \infty} \E[\tau_{\delta}]/\log(1/\delta) \leq (\tau^{\ast})^{-1}$. Unfortunately, this GLRT stopping rule itself requires a projection onto each element of $\mc{Z}$. We leave it as an open question whether an algorithm can be developed which is asymptotically optimal, requires no explicit projection, and has a finite expected stopping time in high probability.

\paragraph{Remark: Bounded assumptions on $\Theta$.}
We assume $\Theta$ is closed and bounded. The boundedness assumption is needed since we would like to control that for each $\theta\in\Theta$, the rewards $x^\T\theta$ to be bounded for all arms $x\in\X$, which is used in our regret analysis  for each learner. Learning algorithms such as AdaHedge~\cite{de2014follow} avoid the need for bounded rewards and we leave it as a future research direction to remove this condition.

\section{Related Work}\label{sec:related_works}

\paragraph{Pure Exploration Linear Bandits} The pure exploration linear bandit problem was introduced in the seminal work of Soare et al \cite{soare2014best}. In recent years, there has been renewed interest in this problem due to its ability to capture many best-arm-identification and pure exploration settings. 
Following the experimental design approach first considered by \cite{soare2014best}, several different algorithmic frameworks were considered \cite{tao2018best,xu2018fully,karnin2013almost}. 

One of the first algorithms to achieve matching instance-optimal upper and lower bounds (within logarithmic factors) for the case of $\mathbb{R}^d$ was by \cite{fiez2019sequential} and depends on an elimination scheme. Shortly after, several works proposed asymptotically optimal algorithms. The first of these methods utilized the track and stop approach given in \cite{jedra2020optimal}, which fully solves the $\tau^*$ objective of Equation~\ref{eq:taustar} using a plug-in estimator $\hat{\theta}_t$ at each round.  Due to the computational difficulty of this, several works proposed alternatives that iteratively updated the sampling distribution in each round. This includes the game theoretic viewpoint we utilize first proposed by \cite{degenne2020gamification, degenne2019non}, and a novel modification of Frank-Wolfe by \cite{wang2021fast}. Other works have augmented these approaches by providing elimination schemes to reduce the set of alternative $\mc{Z}$ that need to be considered each round. \cite{zaki2022improved} proposes a hybrid approach combining the elimination from \cite{fiez2019sequential} and \cite{degenne2020gamification} to remove the condition that $\Theta$ needs to be bounded. \cite{tirinzoni2022elimination} provide an elimination approach where they carefully exploit properties of $\mc{Z}$. Finally, we mention that the pure exploration problem has also been considered in the generalized linear bandit (logistic) settings in \cite{kazerouni2021best, jun2021improved}. Future work could explore extending sampling methods to these settings.

\paragraph{Oracle Based Approaches} As discussed before, if $\mc{Z}$ is a large or combinatorial set, it may be impossible to maintain and appropriate oracles are needed. \cite{katz2020empirical} considers the linear combinatorial setting for matroid-like classes e.g.  shortest-path, top-k, and bipartite matching. By exploiting ideas similar to \cite{fiez2019sequential}, they provide an algorithm utilizing the argmax oracle to achieve near optimal sample complexity. A recent work by \cite{li2022instance} reduces optimal policy learning in agnostic contextual bandits to pure exploration and provides a method analogous to \cite{agarwal2014taming} which only relies on cost-sensitive classification.

\paragraph{Top Two Methods} Our approach is perhaps most reminiscent of the Top-Two Thompson Sampling (TTTS) algorithm for best-arm identification in multi-armed bandits\footnote{i.e. the arms are standard basis vectors $\mc{X}=\mc{Z} = \{e_1, \cdots, e_d\}\in \mathbb{R}^d$ and $\Theta = [0,1]^d$} of \cite{russo2016simple}. Similar to Thompson sampling~\cite{russo2018tutorial}, TTTS maintains a posterior distribution over the means of the arms, and at each round samples a mean vector from the distribution and chooses the arm with the highest sampled mean. It then continues to sample mean vectors, until one is returned whose highest mean is different from the previous found one. Both arms are then pulled. As discussed in the introduction, our algorithm is similar in spirit - we sample until finding a parameter vector whose best-arm is different from our current estimate and then we utilize these vectors to update our learners. Top-two algorithms for multi-armed bandits perform well in practice and have been extensively studied in Bayesian and frequentist settings under various assumptions on noise \cite{qin2017improving, shang2020fixed, jourdan2022top, qin2022adaptivity,lee2023thompson}. However, they often depend on a parameter $\beta$, and only achieve a weaker notion of $\beta$-optimality. Our work is the first to propose and analyze an asymptotically optimal Top-two algorithm for the general linear bandit setting. We remark that the LinGapE algorithm \cite{xu2018fully} also uses a top-two approach and tends to perform well empricially, however it is unknown whether it is asymptotically optimal. 


\paragraph{Online Learning and Thompson Sampling} Finally we remark that the connection between Thompson Sampling and online learning has been previously explored in the early work of \cite{li2013generalized}. This work focuses on the regret setting. Other works in the regret setting have explored connections between information-theoretic analysis of Thompson sampling and online stochastic mirror descent algorithms \cite{lattimore2021mirror, zimmert2019connections}. We hope that our work provides a strong step in this direction for the structured pure exploration literature. 

\section{Experiments}\label{sec:experiments}

In the following, we provide some preliminary experiments to demonstrate the performance of Algorithm~\ref{alg:exp_stochastic}.  Note that the contribution of this paper is primarily theoretical - our goal is to demonstrate that asymptotically optimal algorithms for pure exploration can rely purely on sampling oracles. We hope that the preliminary experiments we provide encourage further exploration of this line of thinking and lead to algorithms that can be as easy to apply as Thompson sampling in practice.

With this in mind, we ran the following modification of some of the algorithms of the previous section. Firstly, we eschewed the doubling trick and instead just ran \peps directly for a fixed horizon side $T$. Secondly, for the $\max$-learner we made use of AdaHedge which is able to use an adaptive step size. Finally, we set $\eta_p = 1$. Though our algorithm only has theoretical guarantees over a bounded set $\Theta$, we believe that this is primarily  a limitation of our analysis and so we set $\Theta = \mathbb{R}^d$. We also remove the forced $G$-optimal exploration for the same reason. For the sampling oracle, we use rejection sampling method because of its simplicity. We demonstrate empirically that the computation cost is not onerous. We plot the number of rejection steps used each round along with clock time per iteration for our method in Appendix~\ref{sec:supp_plots}. Further details on our experimental setup and additional evaluations are also in Appendix~\ref{sec:supp_plots}. 

\begin{table*}[t]
\centering
\begin{tabular}{|l|lll|lll|lll|}
\hline
        & \multicolumn{3}{c|}{Soare's instance \cite{soare2014best}}                                                                             & \multicolumn{3}{c|}{Sphere}                                 & \multicolumn{3}{c|}{TopK}                                                                  \\ \hline
$\delta$   & \multicolumn{1}{l|}{0.1}                & \multicolumn{1}{l|}{0.05}               & 0.01               & \multicolumn{1}{l|}{0.1} & \multicolumn{1}{l|}{0.05} & 0.01 & \multicolumn{1}{l|}{0.2}  & \multicolumn{1}{l|}{0.1}                 & 0.05                \\ \hline
PEPS    & \multicolumn{1}{l|}{1027}                & \multicolumn{1}{l|}{1606}               & 3326               & \multicolumn{1}{l|}{294} & \multicolumn{1}{l|}{476}  & 794  & \multicolumn{1}{l|}{7326} & \multicolumn{1}{l|}{14188}               & 22518               \\ \hline
LinGame & \multicolumn{1}{l|}{828}                & \multicolumn{1}{l|}{1500}               & 2688               & \multicolumn{1}{l|}{186} & \multicolumn{1}{l|}{282}  & 638  & \multicolumn{1}{l|}{8838} & \multicolumn{1}{l|}{29963} & >30000 \\ \hline
LinTS   & \multicolumn{1}{l|}{\textgreater{}5000} & \multicolumn{1}{l|}{\textgreater{}5000} & \textgreater{}5000 & \multicolumn{1}{l|}{431} & \multicolumn{1}{l|}{\textgreater{}1000}  & \textgreater{}1000 & \multicolumn{1}{l|}{N/A}  & \multicolumn{1}{l|}{N/A}                 & N/A                 \\ \hline
\end{tabular}
\medskip
\caption{The number of samples needed for $\P_{\theta\sim\pi_\ell}(\hat{z}_\ell=z_*)>1-\delta$ for various algorithms}
    \label{tab:id_rate}
\end{table*}

The main algorithms we compare to are Thompson Sampling \cite{russo2018tutorial} and LinGame \cite{degenne2020gamification}. LinGame is based on the two-player game strategy with best-response detailed in Section~\ref{sec:posterior}. For a fair comparison, we run LinGame without stopping. The goal of our experiments was to demonstrate that sampling and no-projection algorithms can be competitive against algorithms that explicitly project. From this perspective, we did not consider algorithms that eliminate. For a more extensive empirical comparison of existing algorithms, please see \cite{tirinzoni2022elimination}. We include an oracle strategy and a comparison to LinGapE \cite{xu2018fully} in Appendix~\ref{sec:supp_plots}.

In summary, our algorithm achieves a similar performance compared to LinGame while beating LinTS in Soare and Sphere instances. For Top-k instance, our algorithm beats both LinGame and LinTS. Note that our algorithm is the first algorithm that relies purely on just sampling oracles and our theoretical analysis is only asymptotic, the experimental results are satisfactory since they show that our algorithm works decently well in practice. Now we detail the setting for each instance. 



\begin{figure}[!htb]
    \centering
    \includegraphics[scale=0.2]{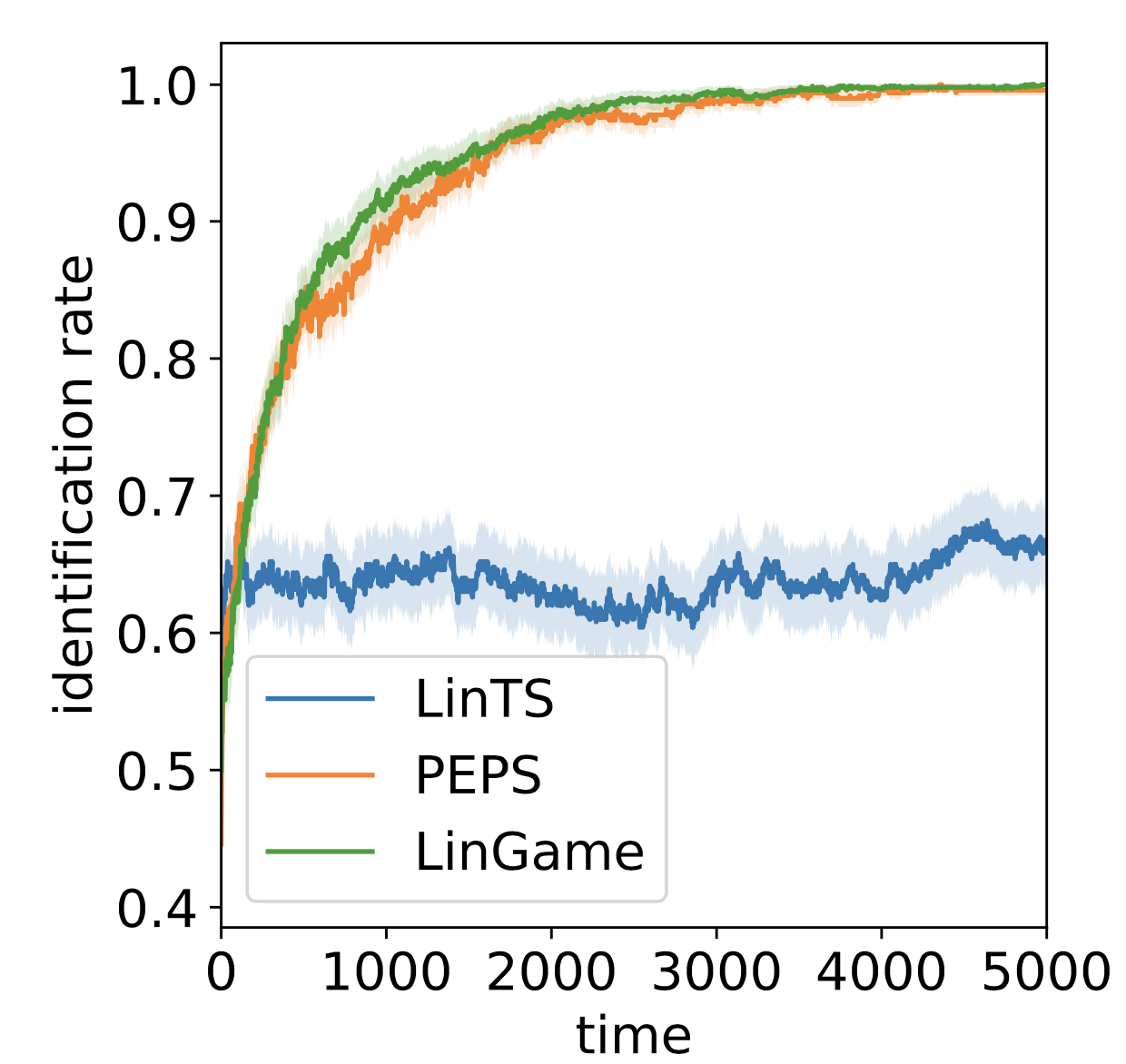}
    \includegraphics[scale=0.2]{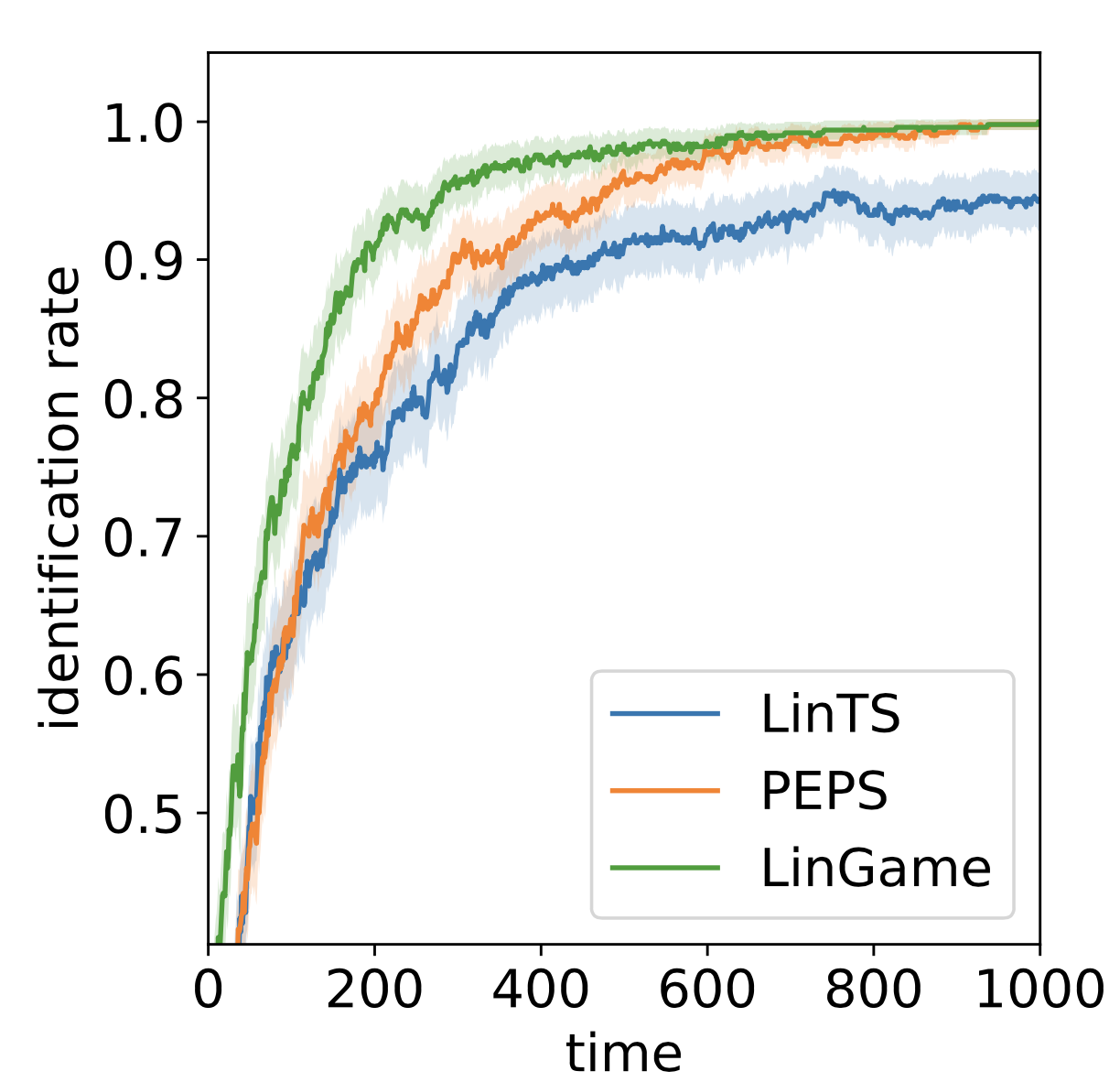}
    \includegraphics[scale=0.16]{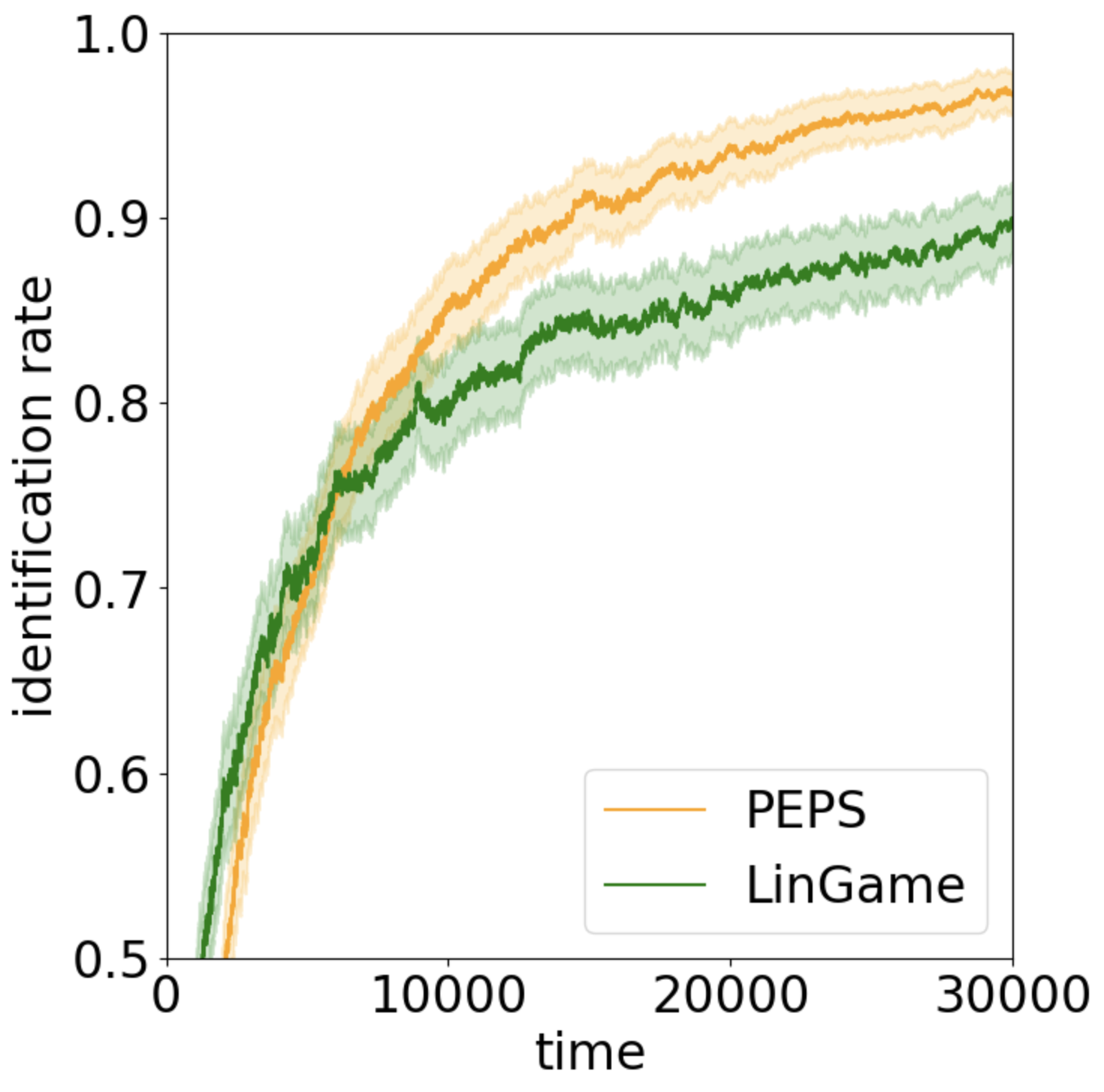}
    \caption{Best-arm identification rate for PEPS, LinGame, and Thompson sampling algorithms under three instances: Soare instance with $\omega=0.1$, sphere instance with $d=6$ and $|\X|=20$, and Top-k instance with $d=12$ and $k=3$. We ran 500 repetitions for all instances. Confidence intervals with plus or minus two standard errors are shown.} 
    \label{fig:id_rate_all}
\end{figure}


\paragraph{Soare's Instance \cite{soare2014best}.} The first instance we consider is the standard benchmark linear bandit instance described in \cite{soare2014best}. 
In this instance, the arm set $\X\subset\R^2$ with $|\X|=3$. The first two arms are $x_1=e_1,x_2=e_2\subset \mathbb{R}^2$, the canonical basis vectors, and an informative arm $x_{3}=(\cos(\omega), \sin(\omega))$. The true parameter is $\theta_*=(1,0)\in\R^d$. 

In this problem, the optimal arm is always $x_1$. However, when the angle $\omega$ is small, it becomes challenging to distinguish the interfering arm $x_{d+1}$ from $x_1$. An effective sampling strategy would pull arm $x_2$ instead of $x_1$ to reduce uncertainty between $x_1$ and $x_{d+1}$ effectively. However, Thompson sampling will tend to pull $x_1$, which will take much longer to distinguish between the two competing arms. The experiments were carried out on a problem instance with $d=2$ and $\omega=0.1$. Our algorithm achieves a similar performance compared with LinGame and beats LinTS. 


\paragraph{Sphere.}
Following~\cite{tao2018best,degenne2020gamification}, we also consider a linear bandit instance where the arm set $\X\subset B^d:=\{x\in\R^d: \norm{x}_2=1\}$ is randomly drawn from a unit sphere of dimension $d$. For the true parameter, we select the two arms, $x$ and $x'$, that are closest to each other, and define $\theta_\ast=x+0.01(x'-x)$, ensuring that $x$ is the best arm. In our experiment, we run the three algorithms on a problem instance with $d=6$ and $|\X|=20$. 
As we can see, our algorithm still outperforms Thompson sampling and is competitive with LinGame. 

\paragraph{Top-k.} The third instance we consider is the top-k combinatorial bandit problem where the goal is to identify the top-k means. In the linear setting, this can be expressed as $\mc{X} = \{e_1, \cdots, e_d\}\subset\mathbb{R}^d$ and $\mc{Z} = \{e_{i_1} + \cdots + e_{i_k}: i_1, \cdots, i_k \in \binom{[d]}{k}\}\subset\mathbb{R}^d$, i.e. $\mc{X}$ is the standard basis and $\mc{Z}$ is the set of indicator vectors of subsets of size $k$. Then, the best arm in this new arm set $\ZZ$ corresponds to the top-k arms in $\X$, which is the goal of top-k identification. Then we run BAI algorithms on this new arm set. We take $\theta = [1, .95, .90, \cdots, 1-.05i, \cdots ]\in \mathbb{R}^d$. As we can see, our algorithm outperforms LinGame in this instance. 

We also present Table~\ref{tab:id_rate} describing the number of samples needed to reach a $1-\delta$ idenfication rate for various $\delta$ values. Note that we do not run Thompson sampling for the Top-k instance (it is not defined when $\mc{X} \ne \mc{Z}$ so we put N/A there, and $>n$ in the table means that the algorithm fails to achieve $1-\delta$ for the $n$ iterations we run in the experiment. We can see that our algorithm, PEPS, achieves an $1-\delta$ best-arm identification probability for all $\delta$ in all instances, with a rate similar to LinGame, outperforming LinTS in all three instances. 
\section{Conclusion}
In this paper, we present the first sampling-based projection-free algorithm for pure exploration in linear bandits. Our algorithm only relies on a sampling oracle and an argmax oracle, so our algorithm is tractable in various settings.  
We show that our algorithm is asymptotically optimal in the sense that the probability that we do not identify the optimal arm decays exponentially with the optimal rate for a fixed allocation. 
We provide experiments demonstrating that our algorithm beats Thompson sampling and has competitive performance against benchmark algorithms such as LinGame \cite{degenne2020gamification} in various problem instances.  Our current approach has various limitations: for example, we need to assume that $\Theta$ is bounded. However, we hope that this work opens a line of investigation into better sampling-based algorithms for effective exploration. 

\bibliographystyle{plain}
\bibliography{refs}


\appendix

\tableofcontents
\newpage

\section{Notations and general description}\label{sec:tableofnotations}
In the following, we let the index $t$, $1\leq t\leq T_{\ell}$ denote the timestep in round $\ell$ for any $\ell$. Throughout this section we will make use of the filtration $\mc{F}_t = \{(x_s, \theta_s, y_s)\}_{s=1}^{t-1}$ defined in any round. The table below summarizes the notations used in the proof. 

\begin{table}[!htb]
\begin{tabularx}{\textwidth}{@{}XX@{}}
\toprule
  $\overline{p}_{T_\ell}=\frac{1}{T_\ell}\sum_{t=1}^{T_\ell} p_t$ & Average of $p$ at the end of round $\ell$ \\
  $\overline{e}_{T_\ell}=\frac{1}{T_\ell}\sum_{t=1}^{T_\ell} e_{x_t}$ & Empirical probability of arms pulled at the end of round $\ell$ \\
  $\pi_{\ell} \sim N(\hat{\theta}_{T_\ell+1},\eta_p^{-1}V_{T_\ell}^{-1})$ restricted on $\Theta$ & The distribution $\theta$ is sampled from at the end of round $\ell$ \\
  $\Delta_{\min}=\min_{x\ne x^*} (x^*-x)^\T\theta^*$ & minimum gap \\
  $T_2(\ell)=\max_{x\in\X}\left(\frac{6\sqrt{\log(|\X|T_\ell\ell^2)}}{\lambda^G_x}\right)^4$ & a time after which each arm gets sufficiently number of pulls \\
  {\tiny $T_0(\ell)=\max\left\{\left(\frac{d\beta(t,\ell^2)\max_{z\in\ZZ} \norm{z}_1}{\Delta_{\min}}\right)^{4/3}, T_2(\ell)+1\right\}$}   & a time after which we have $\hat{z}_t=z_*$ with high probability \\
  $\ell_0  := \min\{\ell: T_{\ell} \geq T_0(\ell)^{3/2}\}$ & minimum round number such that we have guarantee of convergence with high probability \\
  $L$ & upper bound on $\max_{x\in \mc{X}} \|x\|_2$ \\
  $B$ & upper bound on $\|\theta_*\|_2$ \\
  $B_{\mc{X}}$ & $\max_{x\in \mc{X}}\max_{\theta\in \Theta} x^{\top}\theta$ \\
  $\Delta_{\max}$ & $\max_{x\in\X}\max_{\theta,\theta'\in\Theta}|x^\T(\theta-\theta')|$\\
  $\beta(t, 1/\delta) = B+\sqrt{2\log(1/\delta)+d\log\left(\frac{d+tL^2}{d}\right)}$ & anytime confidence bound for $\norm{\hat{\theta}_t-\theta^*}_{V_{t-1}}^2$\\
  $C_{1,\ell} = \Delta_{\max}+L^2\beta(T_{\ell},\ell^2)$ & an upper bound on $\max_{x\in \X}\max_{t\leq T_{\ell}} |\langle x, \hat{\theta}_t\rangle|$ \\
  $C_{3,\ell} =B_{\mc{X}}+\Delta_{\max}+L^2\beta(T_{\ell},\ell^2)$ & an upper bound on $\max_{x\in \X}\max_{\theta\in \Theta}\max_{t\leq T_{\ell}} |\langle x, \theta-\hat{\theta}_t\rangle|$ \\
\bottomrule
\end{tabularx}
\caption{Table of constants and upper bounds used in the proof}
\end{table}

Let $N_{t,x}$ denote the number of times arm $x$ gets pulled at time $t$. We then define several good events needed to guarantee the performance of PEPS at round $\ell$. 
\begin{align*}
    \EE_{1,\ell}&=\bigcup_{t=1}^{T_\ell}\left\{\norm{\hat{\theta}_t-\theta^*}_{V_{t-1}}^2\leq \beta(t, \ell^2)\right\},\\
    \EE_{2,\ell}&=\bigcup_{t=1}^{T_\ell}\left\{\max_{x\in\X}|x^\T\hat{\theta}_t|\leq C_{1,\ell}\right\},\\
    \EE_{3,\ell}&=\bigcup^{T_\ell}_{t\geq T_2 }\bigcup_{x\in \mc{X}} \mc{G}_{t,x}\text{ where }\mc{G}_{t,x} = \{V_t\geq t^{3/4}A(\lambda^G)\}, \forall t\geq T_2, x\in \mc{X}\\
    \EE_{4,\ell}&= \cup_{t\geq T_0}\mathbf{1}\{\hat{z}_t  = z_{\ast}\}\\
\end{align*}
Throughout the proof we also define for some random variable $x\in\X$ with $x\sim p$ and some function $f(x)$, 
\[\FF_{x\sim p}[f(x)]=\sum_{x\in\X}p_x f(x).\]

The rest of the supplement is organized as follows. In Section~\ref{sec:proof_lower_bound}, we present a proof of the lower bound stated in Theorem~\ref{thm:lower_bound}. Section~\ref{sec:supp_plots} provides more experimental results.

In Section~\ref{sec:main_thm}, we prove the main theorem (Theorem~\ref{thm:PEPS}) stated in the paper by combining a saddle-point convergence argument with a guarantee on the likelihood ratio. We tackle the latter in Section~\ref{sec:likelihood_ratio}, where we provide we relate the empirical probability of finding the best-arm at the end of a round of PEPS to the likelihood ratio. 
In Section~\ref{sec:one_round_PEPS}, we show the saddle point approximation and provide a guarantee on how well $\tau^{\ast}$ is approximated after one round of PEPS. 
This argument depends on 
\begin{itemize}
\item Section~\ref{sec:max_learner} and \ref{sec:min_learner} which provide regret guarantees on the $\max$ and $\min$ learners. 
\item Section~\ref{sec:approx_guarantee} provides lemmas bounding terms related to the approximation error of $\hat{\theta}_{T_\ell}$ to $\theta^*$. 
\item Section~\ref{sec:sampling} formally shows that after certain rounds each arm gets enough samples. 
\item Section~\ref{sec:bounds_true} shows that good events needed to guarantee performance of PEPS happen with high probability. 
\end{itemize}
Finally, Section~\ref{sec:tech_lemma} provides some technical lemmas used in the proof. 

\section{Proof of Theorem \ref{thm:lower_bound}}\label{sec:proof_lower_bound}
\begin{theorem}
Fix $\Theta = \R^d$ and any $\theta_* \in \Theta$.
For some $\lambda$ consider a procedure that draws $x_1,\dots,x_T \sim \lambda$, then observes $y_t = \langle x_t, \theta_* \rangle + \epsilon_t$ with $\epsilon_t \sim \mc{N}(0,1)$, and then computes $\widehat{z}_T = \arg\max_{z \in \mc{Z}} \langle z , \widehat{\theta}_T \rangle$ where $\widehat{\theta}_T = \arg\min_{\theta \in \Theta} \sum_{t=1}^T \| y_t - \langle \theta, x_t \rangle \|_{2}^2$.
Then for any $\lambda \in \triangle_{\mc{X}}$ we have
\begin{align*}
    \limsup_{T \rightarrow \infty} -\frac{1}{T}\log\Big( \P_{\theta_*,x_t \sim \lambda}( \widehat{z}_T \neq z_* ) \Big) \leq \tau^*.
\end{align*}
\end{theorem}

\begin{proof}
Assume that $\{z-z_*\}_{z \in \mc{Z}}$ span $\R^d$. 
Otherwise, discard the components of $\mc{X}$ and $\theta_*$ that are orthogonal to the span of $\{z-z_*\}_{z \in \mc{Z}}$ and reparameterize in the subspace spanned by $\{z-z_*\}_{z \in \mc{Z}}$.
We can then work in this reparameterized space, so without loss of generality we can assume  $\{z-z_*\}_{z \in \mc{Z}}$ span $\R^d$. 

Furthermore, assume that $\mc{X}$ spans $\R^d$. If this were not true, then there could be a component of $\theta_*$ that is orthogonal to the span of $\mc{X}$ which makes $z_*$ not identifiable since we assumed $\{z-z_*\}_{z \in \mc{Z}}$ spans $\R^d$.
That is, if $\theta_*^\perp$ is the projection of $\theta_*$ onto the subspace orthogonal to the span of $\mc{X}$, then $\langle z - z_*, \theta_*^\perp \rangle$ could be arbitrarily large but no measurement could detect $\theta_*^\perp$.

Putting the two assumptions together, we conclude that there exists a $\lambda \in \triangle_{\mc{X}}$ such that $A(\lambda) \succ 0$ (equivalently, $\lambda_{\min}(A(\lambda)) > 0$) and $\max_{z \in \mc{Z}} \|z-z_*\|_{A(\lambda)^{-1}} < \infty$.
Fix any $\lambda$ satisfying such conditions.
Define the event $G_\lambda = \{\sum_{t=1}^T x_t x_t \succeq A(\lambda)T(1 - g_{\lambda,T} ) \}$ for some $g_{\lambda,T} = o(T)$ sequence to be defined next.

By applying matrix Chernoff to the random matrices $\{ \frac{1}{T} A(\lambda)^{-1} x_t x_t^\top \}_t$ we have for any $\epsilon \in [0,1)$ that
\begin{align*}
    \P\Big( \frac{1}{T} \sum_{t=1}^T x_t x_t^\top \succeq A(\lambda) (1 - \epsilon) \Big) \geq 1 - d \exp( - \epsilon^2 / 2 R )
\end{align*}
where $R = \max_t \lambda_{\max}( \frac{1}{T} A(\lambda)^{-1} x_t x_t^\top )$.
Observe that 
\begin{align*}
    \lambda_{\max}( \frac{1}{T} A(\lambda)^{-1} x_t x_t^\top ) &\leq \| \frac{1}{T} A(\lambda)^{-1} x_t x_t^\top \|_2 \\
    &\leq L^2 / \lambda_{\min}(A(\lambda)) T.
\end{align*}
So taking $\epsilon = g_{\lambda,T} = \sqrt{ \frac{2 L^2 \lambda_{\min}(A(\lambda))^{-1} \log(d T) }{T} }$ we have that $\P(G_\lambda) \geq 1 - 1/T$ whenever $g_{\lambda,T} < 1$ which holds for sufficiently large $T$.

Now, for any $\{x_t \}_{t=1}^T$ that span $\R^d$ (will be guaranteed by event $G_\lambda$) we have that
\begin{align*}
    \widehat{\theta}_T &= \arg\min_{\theta \in \Theta} \sum_{t=1}^T \| y_t - \langle \theta, x_t \rangle \|_{2}^2 \\
    &= \Big( \sum_{t=1}^T x_t x_t^\top \Big)^{-1} \sum_{t=1}^T x_t y_t \\
    &= \theta_* + \Big( \sum_{t=1}^T x_t x_t^\top \Big)^{-1} \sum_{t=1}^T x_t \epsilon_t \\
    &= \theta_* +  \Big( \sum_{t=1}^T x_t x_t^\top \Big)^{-1/2} \eta
\end{align*}
where the last line holds with inequality in distribution for $\eta \sim \mc{N}(0,I_d)$.
We conclude that for any $z$ that $\langle \widehat{\theta}_T - \theta_* , z - z_* \rangle$ is a zero-mean Gaussian random variable with variance
\begin{align*}
     \sigma_{z,\lambda}^2 :=& \E[ \langle \widehat{\theta}_T - \theta_* , z - z_* \rangle^2] \\
     =& \E[ \langle \Big( \sum_{t=1}^T x_t x_t^\top \Big)^{-1/2} \eta, z - z_* \rangle^2 ] \\
     =& ( z - z_* )^\top \Big( \sum_{t=1}^T x_t x_t^\top \Big)^{-1} ( z - z_* ).
\end{align*}
Thus, on $G_\lambda$ we have that $\sigma_{z,\lambda}^2
     \leq  \frac{1}{T( 1- g_{\lambda,T})} \|z - z_* \|_{ A(\lambda)^{-1}}^2$.

Consequently,
\begin{align*}
    \P_{\theta_*}( \widehat{z}_T \neq z_* ) &= \P_{\theta_*}( \bigcup_{z \in \mc{Z} \setminus z_*} \{ \widehat{z}_T = z, z \neq z_* \}) \\ &\geq \max_{z \in \mc{Z} \setminus z_*}\P_{\theta_*}(  \widehat{z}_T = z, z \neq z_* ) \\
    &= \max_{z \in \mc{Z} \setminus z_*}\P_{\theta_*}(   \langle \widehat{\theta}_T , z - z_* \rangle \geq 0 ) \\
    &= \max_{z \in \mc{Z} \setminus z_*}\P_{\theta_*}(   \langle \widehat{\theta}_T - \theta_* , z - z_* \rangle \geq \langle \theta_*, z - z_* \rangle ) \\
    &\geq \max_{z \in \mc{Z} \setminus z_*}\E_{ \{x_t\} \sim \lambda} \E_{\theta_*}\big[ \mathbf{1}\{ G_\lambda \} \mathbf{1}\big\{ \langle \widehat{\theta}_T - \theta_* , z - z_* \rangle \geq \langle \theta_*, z - z_* \rangle  \big\} \big| \{x_t\} \big]  \Big] \\
    &= \max_{z \in \mc{Z} \setminus z_*}\P_{ \{x_t\} \sim \lambda}(G_\lambda)  \P_{\eta_1 \sim \mc{N}(0,1)}\big( \eta_1 \sigma_{z,\lambda} \geq \langle \theta_*, z - z_* \rangle \big).
\end{align*}
Using the fact that 
\begin{align*}
    \P_{\eta_1 \sim \mc{N}(0,1)}\big( \eta_1 \geq s ) &= \int_{x = s}^\infty \frac{1}{\sqrt{2\pi}} e^{-x^2/2} dx > (\frac{1}{s} - \frac{1}{s^3})  \frac{1}{\sqrt{2\pi}} e^{-s^2/2}
\end{align*}
for positive $s$, we conclude that
\begin{align*}
    &\P_{\theta_*}( \widehat{z}_T \neq z_* )\\
    &\geq \max_{z \in \mc{Z} \setminus z_*}\P_{ \{x_t\} \sim \lambda}(G_\lambda)  \P_{\eta_1 \sim \mc{N}(0,1)}\big( \eta_1 \sigma_{z,\lambda} \geq \langle \theta_*, z - z_* \rangle \big) \\
    &\geq \mathbf{1}\{ g_{\lambda,T} < 1 \} (1-\frac{1}{T}) \max_{z \in \mc{Z} \setminus z_*} (\frac{\sigma_{z,\lambda}}{\langle \theta_*, z - z_* \rangle } - \frac{\sigma_{z,\lambda}^3}{\langle \theta_*, z - z_* \rangle ^3})  \frac{1}{\sqrt{2\pi}} e^{-\frac{\langle \theta_*, z - z_* \rangle^2}{\sigma_{z,\lambda}^2}/2} \\
    &\geq \max_{z \in \mc{Z} \setminus z_*} \mathbf{1}\{ g_{\lambda,T} < 1, \tfrac{\langle \theta_*, z - z_* \rangle^2}{\sigma_{z,\lambda}^2} \geq 2 \} (1-\frac{1}{T}) \frac{\sigma_{z,\lambda}}{\langle \theta_*, z - z_* \rangle }  \frac{1}{\sqrt{8\pi}} e^{-\frac{\langle \theta_*, z - z_* \rangle^2}{\sigma_{z,\lambda}^2}/2} \\
    &\geq \max_{z \in \mc{Z} \setminus z_*} \mathbf{1}\{ g_{\lambda,T} < 1, \tfrac{T( 1- g_{\lambda,T})  \langle \theta_*, z - z_* \rangle^2}{\|z - z_* \|_{ A(\lambda)^{-1}}^2 } \geq 2 \} (1-\frac{1}{T})   \tfrac{ \|z - z_* \|_{ A(\lambda)^{-1}}^2}{ T( 1- g_{\lambda,T}) \langle \theta_*, z - z_* \rangle^2}   \frac{1}{\sqrt{8\pi}} e^{- \frac{ T( 1- g_{\lambda,T}) \langle \theta_*, z - z_* \rangle^2}{ \|z - z_* \|_{ A(\lambda)^{-1}}^2}/2}.
\end{align*}
Thus, because $g_{\lambda,T} = o(T)$ and $\tfrac{ \|z - z_* \|_{ A(\lambda)^{-1}}^2}{ \langle \theta_*, z - z_* \rangle^2} < \infty$ we have that
\begin{align*}
    \limsup_{T \rightarrow \infty} -\frac{1}{T}\log\Big(\P_{\theta_*,x_t \sim \lambda}( \widehat{z}_T \neq z_* ) \Big) &\leq \frac{ \langle \theta_*, z - z_* \rangle^2}{ \|z - z_* \|_{ A(\lambda)^{-1}}^2}/2 \\
    &= \min_{\theta \in \Theta_{z_*}^c} \|\theta-\theta_*\|_{A(\lambda)}^2 /2 \\
    &\leq \max_{\lambda \in \triangle_{\mc{X}}} \min_{\theta \in \Theta_{z_*}^c} \|\theta-\theta_*\|_{A(\lambda)}^2 /2 
    =  \tau^*
\end{align*}
where the second line uses the fact that $\Theta = \R^d$.
\end{proof}

\section{Proof of the Main Theorem}\label{sec:main_thm}

\begin{theorem}\label{thm:asymp_pos_rate}
Under Algorithm~\ref{alg:exp_stochastic} and \ref{alg:doubling_trick} and Assumption~\ref{assump:Theta}, we have the sampling distribution satisfies with probability 1, 
\[\lim_{\ell\to\infty}-\frac{1}{T_{\ell}}\log\pi_\ell(\Theta_{z_*}^c)=\tau^*.\]
\end{theorem}
\begin{proof}

By Theorem~\ref{thm:good_events}, we have that for $\ell\geq \ell_0$, $\P(\EE^c_\ell)\leq \frac{5}{\ell^2}$. Also, since $T_\ell=2^{\ell}$, and $T_0(\ell)$ only scales logarithmically in $\ell$, so $\ell_0<\infty$. Therefore, $\sum_{\ell=1}^\infty \P(\EE_\ell)<\infty$. By Borel-Cantelli, we have 
\[\P\bigsmile{\limsup_{n\to\infty}\,\EE^c_\ell}=0.\]
Note that $\limsup_{\ell\to\infty} \EE_\ell=\bigcap_{\ell=1}^\infty\bigcup_{k=\ell}^\infty \EE_k$, this implies that the probability that infinitely many of them occur is zero, which means that $\EE_{\ell}$ eventually holds for sufficiently large $\ell$ with probability 1. However, under $\EE_{\ell}$ we have
\begin{align*}
    \pi_\ell(\Theta_{z_*}^c) &= \frac{\int_{\Theta_{z_*}^c} \pi_\ell(\theta)d\theta}{\int_{\Theta} \pi_\ell(\theta)d\theta} = \frac{\int_{\Theta_{z_*}^c} \pi_\ell(\theta)/\pi_\ell(\theta^*)d\theta}{\int_{\Theta} \pi_\ell(\theta)/\pi_\ell(\theta^*)d\theta}\\
    &\doteq \frac{\int_{\Theta_{z_*}^c} e^{-\frac{T_{\ell}}{2}\norm{\theta-\theta^*}_{A(\overline{e}_{T_{\ell}})}^2}d\theta}{\int_{\Theta} e^{-\frac{T_{\ell}}{2}\norm{\theta-\theta^*}_{A(\overline{e}_{T_{\ell}})}^2}d\theta}\tag{by $\EE_{\ell}$}\\
    &\doteq  e^{-T_{\ell}\inf{\theta\in\Theta_{z_*}^c}\frac{1}{2}\norm{\theta-\theta^*}_{A(\overline{e}_{T_{\ell}})}^2}\tag{Lemma~\ref{lem:Laplace_approx} and $\inf{\theta\in\Theta} \norm{\theta-\theta^*}_{A(\lambda)}^2=0$ for any $\lambda$}.
\end{align*}
This implies that there exists some $\epsilon_\ell'\to 0$ such that 
\[\left|-\frac{1}{T_\ell}\log \pi_\ell(\Theta_{z_*}^c)- \inf{\theta\in\Theta_{z_*}^c} \frac{1}{2}\norm{\theta-\theta^*}_{A(\overline{e}_{T_\ell})}^2\right|\leq \epsilon_\ell'.\]

Under $\EE_{6,\ell}$, there exists some sequence $\epsilon_\ell\to 0$ such that 
\[\tau^*-\inf{\theta\in\Theta_{z_*}^c} \frac{1}{2}\norm{\theta-\theta^*}_{A(\overline{e}_{T_\ell})}^2\leq \epsilon_\ell.\]
Since 
\[\tau^*=\max_{\lambda\in\triangle_\X}\inf{\theta\in\Theta_{z_*}^c} \frac{1}{2}\norm{\theta-\theta^*}_{A(\lambda)}^2\geq \inf{\theta\in\Theta_{z_*}^c} \frac{1}{2}\norm{\theta-\theta^*}_{A(\overline{e}_{T_\ell})}^2,\]
combining the above three displays, we have under $\EE_{\ell}$, 
\[\left|-\frac{1}{T_\ell}\log \pi_\ell(\Theta_{z_*}^c)-\tau^*\right|\leq \epsilon_\ell+\epsilon_\ell',\]
where $\epsilon_\ell+\epsilon_\ell'\to 0$ as $\ell\to\infty$. Combining this with the fact that $\P\bigsmile{\limsup_{\ell\to\infty}\,\EE_\ell}=0$, we have with probability 1, 
\[\lim_{\ell\to\infty}-\frac{1}{T_\ell}\log \pi_\ell(\Theta_{z_*}^c)=\tau^*.\]
\end{proof}

\begin{theorem}\label{thm:good_events}
In round $\ell$ for $\ell \geq \ell_0$, define
\begin{align*}
    \EE_{5,\ell}&= \left\{\sup_{\theta\in\Theta} \frac{1}{T_\ell}\left|\log\frac{\pi_{T_\ell}(\theta^*)}{\pi_{T_\ell}(\theta)}-\frac{T_\ell}{2}\norm{\theta-\theta^*}_{A(\overline{e}_{T_\ell})}^2\right|\leq \kappa_\ell\right\}\\
    \EE_{6,\ell}&= \left\{\left|\max_{\lambda\in\triangle_\X}\inf{\theta\in\Theta_{z_*}^c} \frac{1}{2}\norm{\theta-\theta^*}_{A(\lambda)}^2- \inf{\theta\in\Theta_{z_*}^c} \frac{1}{2}\norm{\theta-\theta^*}_{A(\overline{e}_{T_\ell})}^2\right|\leq \epsilon_\ell\right\}
\end{align*}    
with $\epsilon_{\ell} \to 0$ and $\kappa_\ell\to 0$ as $\ell\to\infty$. Define $\EE_{\ell} = \EE_{5,\ell} \cap  \EE_{6,\ell}$. Then $\P(\EE_{\ell})\geq 1-5/\ell^2$.
\end{theorem}
\begin{proof}
We first summarize the guarantees for the probabilities of events below. For $\ell\geq \ell_0$, we have
\begin{itemize}
    \item from Lemma~\ref{lem:dual_gap_bound}, we have that $\P(\EE_{6,\ell}|\EE_{1,\ell}\cap\EE_{2,\ell}\cap\EE_{3,\ell} \cap\EE_{4,\ell}) \geq 1-1/\ell^2$ with choice of $\epsilon_\ell=O(T_\ell^{-1/4})$;
    \item from Lemma~\ref{lem:EE_1_true}, $\P(\EE_{1,\ell}) \geq 1-1/\ell^2$;
    \item by Lemma~\ref{lem:EE_2_true}, $\EE_{2,\ell}$ is true under $\EE_{3,\ell}\cap \EE_{1,\ell}$;
    \item by Lemma~\ref{lem:xteq}, $\P(\EE_{4,\ell}|\EE_{1,\ell}) \geq 1-1/\ell^2$;
    \item by Lemma~\ref{lem:likelihood_ratio} with $\kappa_\ell=O(T_\ell^{-1/2})$, $\P(\EE_{5,\ell})\geq 1-1/\ell^2$;
    \item by Lemma~\ref{lem:sampling}, $\P(\EE_{3,\ell})\geq 1-1/\ell^2$.
\end{itemize}
Note that $\EE_{\ell}\supset \EE_{1,\ell}\cap \EE_{2,\ell}\cap \EE_{3,\ell}\cap \EE_{4,\ell}\cap \EE_{5,\ell}\cap \EE_{6,\ell}$, and so 
\begin{align*}
    \EE_{\ell}^c &\subset \EE_{1,\ell}^c\cup \EE_{2,\ell}^c\cup \EE_{3,\ell}^c\cup \EE_{4,\ell}^c\cup\EE_{5,\ell}^c\cup \EE_{6,\ell}^c\\
    &= \EE_{1,\ell}^c\cup (\EE_{2,\ell}^c\cap \EE_{1,\ell}\cap \EE_{3,\ell})\cup \EE_{3,\ell}^c\cup (\EE_{4,\ell}^c\cap \EE_{1,\ell})\cup\EE_{5,\ell}^c\cup (\EE_{6,\ell}^c\cap \EE_{1,\ell}\cap \EE_{2,\ell}\cap\EE_{3,\ell}\cap \EE_{4,\ell}).
\end{align*}
Therefore, for $\ell\geq \ell_0$, 
\begin{align*}
    &\P(\EE_{\ell}^c)\\
    &\leq \P(\EE_{1,\ell}^c)+ \P(\EE_{2,\ell}^c\cap \EE_{1,\ell}\cap \EE_{3,\ell})+\P(\EE_{3,\ell}^c)+ \P(\EE_{4,\ell}^c\cap \EE_{1,\ell})+\P(\EE_{5,\ell}^c)+\P(\EE_{6,\ell}^c\cap \EE_{1,\ell}\cap \EE_{2,\ell}\cap\EE_{3,\ell}\cap \EE_{4,\ell})\\
    &\leq \P(\EE_{1,\ell}^c)+ \P(\EE_{2,\ell}^c| \EE_{1,\ell}\cap \EE_{3,\ell})\P(\EE_{1,\ell}\cap \EE_{3,\ell})+\P(\EE_{3,\ell}^c)+ \P(\EE_{4,\ell}^c| \EE_{1,\ell})\P(\EE_{1,\ell})\\
    &\quad+\P(\EE_{5,\ell}^c)+\P(\EE_{6,\ell}^c|\EE_{1,\ell}\cap \EE_{2,\ell}\cap\EE_{3,\ell}\cap \EE_{4,\ell})\P(\EE_{1,\ell}\cap \EE_{2,\ell}\cap\EE_{3,\ell}\cap \EE_{4,\ell})\\
    &\leq \P(\EE_{1,\ell}^c)+ \P(\EE_{2,\ell}^c| \EE_{1,\ell}\cap \EE_{3,\ell})+\P(\EE_{3,\ell}^c)+ \P(\EE_{4,\ell}^c| \EE_{1,\ell})+\P(\EE_{5,\ell}^c)+\P(\EE_{6,\ell}^c|\EE_{1,\ell}\cap \EE_{2,\ell}\cap\EE_{3,\ell}\cap \EE_{4,\ell})\\
    &\leq \frac{5}{\ell^2}. 
\end{align*}
Therefore, $\P(\EE_\ell)\geq 1-\frac{5}{\ell^2}$. 
\end{proof}

\subsection{Guarantees on the Likelihood Ratio}\label{sec:likelihood_ratio}
\begin{lemma}\label{lem:likelihood_ratio}
We have with probability at least $1-1/\ell^2$, 
\[\sup_{\theta\in\Theta} \frac{1}{T_{\ell}}\left|\log\frac{\pi_{\ell}(\theta)}{\pi_{\ell}(\theta^*)}-\frac{T_{\ell}}{2}\norm{\theta-\theta^*}_{A(\overline{e}_{T_{\ell}})}^2\right|\leq \Delta_{\max}\sqrt{\frac{2d\log\bigsmile{\frac{(d+T_{\ell}L^2)\ell^2}{d}}}{T_{\ell}}}.\]
Which implies that $\frac{\pi_\ell(\theta)}{\pi_\ell(\theta^*)} \doteq e^{-T_\ell\norm{\theta-\theta^*}_{A(\overline{e}_{T_{\ell}})}^2}$.
\end{lemma}
\begin{proof}
Throughout the following we set $T := T_{\ell}$. Recall that $\pi_\ell(\theta)=\mathcal{N}(\hat{\theta}_{T+1},V_{T}^{-1})$ restricted on $\Theta$, which means that for each $\theta\in\Theta$, 
\begin{align*}
    \pi_\ell(\theta)=\frac{\exp\bigsmile{-\frac{1}{2}\norm{\theta-\hat{\theta}_{T+1}}_{V_T}^2}}{\int_{\Theta} \exp\bigsmile{-\frac{1}{2}\norm{\theta'-\hat{\theta}_{T+1}}_{V_T}^2}\,d\theta'}.
\end{align*}
Since the denominator is independent of $\theta$, this means that
\begin{align*}
    \frac{\pi_\ell(\theta)}{\pi_\ell(\theta^*)} &= \exp\left(-\frac{1}{2}\bigsmile{\norm{\theta-\hat{\theta}_{T+1}}_{V_{T}}^2-\norm{\theta^*-\hat{\theta}_{T+1}}_{V_{T}}^2}\right)
\end{align*}
where 
\begin{align*}
    &\norm{\theta^*-\hat{\theta}_{T+1}}_{V_{T}}^2-\norm{\theta-\hat{\theta}_{T+1}}_{V_{T}}^2\\
    &= \norm{\theta^*}_{V_{T}}^2-2(\theta^*)^\T V_{T}\hat{\theta} + \norm{\hat{\theta}_{T+1}}_{V_{T}}^2-\norm{\hat{\theta}_{T+1}}_{V_{T}}^2+2(\hat{\theta}_{T+1})^\T V_{T}\theta-\norm{\theta}_{V_{T}}^2\\
    &= \left\|\theta^*\right\|_{V_T}^2-2\left(\theta^*\right)^{\top} V_T\left(\theta^*+V_T^{-1} \sum_{s=1}^T \epsilon_s x_s\right)+2 \theta^{\top} V_T\left(\theta^*+V_T^{-1} \sum_{s=1}^T \epsilon_s x_s\right)-\norm{\theta}_{V_T}^2\\
    &= \norm{\theta^*}_{V_{T}}^2-2\norm{\theta^*}_{V_{T}}^2-2(\theta^*)^\T \left(\sum_{s=1}^T\epsilon_sx_s\right) +2(\theta^*)^\T V_{T}\theta+2 \theta^\T \left(\sum_{s=1}^T\epsilon_sx_s\right)-\norm{\theta}_{V_{T}}^2\\
    &= -\norm{\theta^*-\theta}_{V_{T}}^2-2\inner{\theta^*-\theta}{\sum_{s=1}^{T} \epsilon_sx_s}\\
    &= -\norm{\theta^*-\theta}_{V_{T}}^2-2\sum_{s=1}^T \epsilon_sx_s^\T(\theta^*-\theta).
\end{align*} 
Note that 
\begin{align*}
    \sum_{s=1}^T \epsilon_sx_s^\T(\theta^*-\theta) &= \sum_{s=1}^T \epsilon_sx_s^\T V_{T}^{-1/2}V_{T}^{1/2}(\theta^*-\theta)\leq \norm{\sum_{s=1}^T \epsilon_sx_s}_{V_{T}^{-1}}\norm{\theta^*-\theta}_{V_{T}}.
\end{align*}
Also,
\begin{align*}
    \norm{\theta^*-\theta}_{V_{T}} &= \sqrt{\norm{\theta^*-\theta}_{V_{T}}^2} = \sqrt{\sum_{t=1}^T (x_t^\T (\theta^*-\theta))^2} \leq \Delta_{\max}\sqrt{T},
\end{align*}
and since $\E[\epsilon_sx_s|\F_{s-1}]=0$ for all $s$, $\epsilon_sx_s$ is a vector-valued martingale. Then by Theorem 1 of \cite{abbasi2011improved}, with probability greater than $1-\delta$, 
\begin{align*}
    \norm{\sum_{s=1}^T \epsilon_sx_s}_{V_T^{-1}}
    \leq \sqrt{2d\log\bigsmile{\frac{d+TL^2}{d\delta}}} 
\end{align*}
so with probability $1-\delta$, 
\begin{align*}
    \norm{\sum_{s=1}^T \epsilon_sx_s}_{V_T^{-1}}\norm{\theta^*-\theta}_{V_T}\leq \Delta_{\max}\sqrt{T}\sqrt{2d\log\bigsmile{\frac{d+TL^2}{d\delta}}}.
\end{align*}
so for any $\theta\in\Theta$, 
\[\left|\bigsmile{\norm{\theta-\hat{\theta}_{T+1}}_{V_{T}}^2-\norm{\theta^*-\hat{\theta}_{T+1}}_{V_{T}}^2}-\norm{\theta^*-\theta}_{V_T}^2\right|\leq \Delta_{\max}\sqrt{T}\sqrt{2d\log\bigsmile{\frac{d+TL^2}{d\delta}}},\]
which means that
\[\left|\log\frac{\pi_{\ell}(\theta^*)}{\pi_{\ell}(\theta)}-\frac{T}{2}\norm{\theta-\theta^*}_{A(\overline{e}_{T})}^2\right|\leq \Delta_{\max}\sqrt{T}\sqrt{2d\log\bigsmile{\frac{d+TL^2}{d\delta}}}.\]
Taking a supremum over $\theta\in\Theta$ on both sides and taking $\delta=\frac{1}{\ell^2}$ gives the result. 
\end{proof}

\subsection{Guarantee on Saddle-Point Convergence of PEPS in Round $\ell$}\label{sec:one_round_PEPS}

In this section, we present a key result to this proof, which shows that as round $\ell$ gets large, the distribution from PEPS achieves the optimal allocation deduced by $\tau^*$. Fix a round $\ell$. At iteration $t$, let $\tilde{\lambda}_t$ denote the sampling distribution of $x_t$ . The result is stated in the following lemma. In the proof, we decompose the difference into several terms and argue about each piece in subsequent sections. 

\begin{lemma}[Guarantee for PEPS]\label{lem:dual_gap_bound}
On $\EE_{1,\ell} \cap \EE_{2,\ell} \cap\EE_{3,\ell} \cap \EE_{4,\ell}$, for $\ell> \ell_0$ then at the end of epoch $\ell$, we have with probability at least $1-\frac{1}{\ell^2}$, 
\begin{align*}
&\tau^*-\inf{\theta\in\Theta_{z_*}^c}\left[\frac{1}{2}\norm{\theta^*-\theta}_{A(\overline{e}_{T_{\ell}})}^2\right]\leq \epsilon_\ell
\end{align*}
for a sequence $\epsilon_\ell\to 0$ as $\ell\rightarrow\infty$.
\end{lemma}
\begin{proof}
Recall the definition of $\bar{p}_{T_{\ell}}$ and $\bar{e}_{T_{\ell}}$ in Section~\ref{sec:tableofnotations}. We first show that there exists some $\epsilon_\ell$ that goes to zero as $\ell\to\infty$ such that under $\EE_{1,\ell} \cap \EE_{2,\ell} \cap\EE_{3,\ell} \cap \EE_{4,\ell}$, for $\ell>\ell_0$, 
\begin{align*}
&\max_{\lambda\in\triangle_\X}\FF_{\theta\sim \overline{p}_{T_{\ell}}}\left[\frac{1}{2}\norm{\theta^*-\theta}_{A(\lambda)}^2\right]-\min_{p\in\PP(\Theta_{z_*}^c)}\FF_{\theta\sim p}\left[\frac{1}{2}\norm{\theta^*-\theta}_{A(\overline{e}_{T_{\ell}})}^2\right]\leq \epsilon_\ell.
\end{align*}
We have 
\begin{align*}
    &\max_{\lambda\in\triangle_\X}\bF_{\theta\sim \overline{p}_{T_{\ell}}}\left[\norm{\theta^*-\theta}_{A(\lambda)}^2\right]-\min_{p\in\PP(\Theta_{z_*}^c)}\bF_{\theta\sim p}\left[\norm{\theta^*-\theta}_{A(\overline{e}_{T_{\ell}})}^2\right]\\
    &= \max_{\lambda\in\triangle_\X}\bF_{\theta\sim \overline{p}_{T_{\ell}}}\left[\norm{\theta^*-\theta}_{A(\lambda)}^2\right]-\inf{\theta\in\Theta_{z_*}^c}\norm{\theta^*-\theta}_{A(\overline{e}_{T_{\ell}})}^2\\
    &= \max_{\lambda\in\triangle_\X}\bF_{\theta\sim \overline{p}_{T_{\ell}}}\left[\norm{\theta^*-\theta}_{A(\lambda)}^2\right]-\frac{1}{T_{\ell}}\inf{\theta\in\Theta_{z_*}^c} \left\|\theta-\hat{\theta}_{T_{\ell}+1}\right\|_{V_{T_{\ell}}}^2+C_{T_\ell}''\\
\end{align*}
\begin{align*}
    &= \max_{\lambda\in\triangle_\X}\bF_{\theta\sim \overline{p}_{T_{\ell}}}\left[\norm{\theta^*-\theta}_{A(\lambda)}^2\right]-\max_{\lambda\in\triangle_\X}\frac{1}{T_{\ell}}\sum_{t=1}^{T_\ell}\bF_{\theta\sim p_t}\left[\norm{\hat{\theta}_t-\theta}_{A(\lambda)}^2\right]\tag{S1. $C_{T_\ell}'$}\\
    &\quad+\max_{\lambda\in\triangle_\X}\frac{1}{T_{\ell}}\sum_{t=1}^{T_\ell}\FF_{\theta\sim p_t}\left[\norm{\hat{\theta}_t-\theta}_{A(\lambda)}^2\right]-\frac{1}{T_{\ell}}\sum_{t=1}^{T_\ell} \bF_{\theta\sim p_{t}}\left[\norm{\theta-\hat{\theta}_t}_{A(\tilde{\lambda}_t)}^2\right]\tag{S2. regret for $\max$ learner}\\
    &\quad+\frac{1}{T_{\ell}}\sum_{t=1}^{T_\ell} \FF_{\theta\sim p_{t}}\left[\norm{\theta-\hat{\theta}_t}_{A(\tilde{\lambda}_t)}^2\right]-\frac{1}{T_{\ell}}\sum_{t=1}^{T_\ell} \FF_{\theta\sim \tilde{p}_{t}}\bigbrak{\left\|\theta-\hat{\theta}_t\right\|_{A(\tilde{\lambda}_t)}^2}\tag{S3. }\\
    &\quad+\frac{1}{T_{\ell}}\sum_{t=1}^{T_\ell} \FF_{\theta\sim \tilde{p}_{t}}\bigbrak{\left\|\theta-\hat{\theta}_t\right\|_{A(\tilde{\lambda}_t)}^2}-\frac{1}{T_{\ell}}\sum_{t=1}^{T_\ell} \FF_{\theta\sim \tilde{p}_{t}}\bigbrak{\left\|\theta-\hat{\theta}_t\right\|_{x_tx_t^\T}^2}\tag{S4. }\\
    &\quad+\frac{1}{T_\ell}\sum_{t=1}^{T_\ell} \FF_{\theta\sim \tilde{p}_{t}}\bigbrak{\left\|\theta-\hat{\theta}_t\right\|_{x_tx_t^\T}^2}-\frac{1}{T_{\ell}}\inf{\theta\in\Theta_{z_*}^c} \left\|\theta-\hat{\theta}_{T_\ell+1}\right\|_{V_{T_{\ell}}}^2\tag{S5. regret for the $\min$ learner}\\
    &+C_{T_\ell}'',
\end{align*}
where we define 
\begin{align*}
    C_{T_\ell}'&:=\max_{\lambda\in\triangle_\X}\bF_{\theta\sim \overline{p}_{T_\ell}}\left[\norm{\theta^*-\theta}_{A(\lambda)}^2\right]-\max_{\lambda\in\triangle_\X}\frac{1}{T_\ell}\sum_{t=1}^{T_\ell}\bF_{\theta\sim p_t}\left[\norm{\hat{\theta}_t-\theta}_{A(\lambda)}^2\right]\\
    C_{T_\ell}''&=\frac{1}{T_\ell}\inf{\theta\in\Theta_{z_*}^c} \left\|\theta-\hat{\theta}_{T_\ell+1}\right\|_{V_{T_\ell}}^2-\inf{\theta\in\Theta_{z_*}^c}\norm{\theta^*-\theta}_{A(\overline{e}_{T_\ell})}^2.
\end{align*}
We now handle each term separately by referring to the lemma which provides a guarantee.

\begin{itemize}
    \item \textbf{(S1)} By Lemma~\ref{lem:CT'}, under $\EE_{1,\ell} \cap \EE_{2,\ell}$, for $T_\ell\geq T_2(\ell)$, 
    \begin{align*}
        &\max_{\lambda\in\triangle_\X}\bF_{\theta\sim \overline{p}_{T_{\ell}}}\left[\norm{\theta^*-\theta}_{A(\lambda)}^2\right]-\max_{\lambda\in\triangle_\X}\frac{1}{T_\ell}\sum_{t=1}^{T_\ell}\bF_{\theta\sim p_t}\left[\norm{\hat{\theta}_t-\theta}_{A(\lambda)}^2\right]\\
        &\leq\CTPrime,
    \end{align*}
    so for $T_\ell\geq T_2(\ell)^{3/2}$, we have the above is upper bounded by 
    \[O(L^2\beta(T_2(\ell), \ell^2)T_\ell^{-1/2}+4d\beta(T_\ell,\ell^2)T_\ell^{-3/4});\]
    \item \textbf{(S2)} By Lemma~\ref{lem:maxlearner}, we have with probability $1-1/(3\ell^2)$ conditioned on $\EE_{2,\ell}$
    \begin{align*}
        &\max_{\lambda\in\triangle_\X}\sum_{t=1}^{T_\ell} \FF_{\theta\sim p_t, x\sim \lambda}\norm{\theta-\hat{\theta}_t}_{xx^\T}^2-\sum_{t=1}^{T_\ell} \FF_{\theta\sim p_t, x\sim \tilde{\lambda}_t}\norm{\theta-\hat{\theta}_t}_{xx^\T}^2\\
        &\leq \eqnmaxlearner,
    \end{align*}
    so with a choice of $\gamma_t=t^{-\alpha}$ with $\alpha=1/4$, 
    \begin{align*}
        &\max_{\lambda\in\triangle_\X}\frac{1}{T_\ell}\sum_{t=1}^{T_\ell} \FF_{\theta\sim p_t, x\sim \lambda}\norm{\theta-\hat{\theta}_t}_{xx^\T}^2-\frac{1}{T_\ell}\sum_{t=1}^{T_\ell} \FF_{\theta\sim p_t, x\sim \tilde{\lambda}_t}\norm{\theta-\hat{\theta}_t}_{xx^\T}^2\\
        &\leq C_{3,\ell}^2\sqrt{\log |\X|}T_\ell^{-1/2}+\sqrt{2C_{3,\ell}^2\log \ell^2}T_\ell^{-1/2}+\sqrt{2C_{3,\ell}^2|\X|\log(3T\ell^2)}T_\ell^{-1/2}+2C_{3,\ell}^2T_\ell^{-1/4}
    \end{align*}
    \item \textbf{(S3)} By Lemma~\ref{lem:ptversusptilde}, we have conditioned on $\EE_{4,\ell}\cap \EE_{1,\ell}$ for $\ell\geq \ell_0$, 
    \[\ptvsptilde\]
    for $T_\ell\geq T_0(\ell)^{3/2}$, we have the above is bounded by $2C_{3,\ell}^2 T_\ell^{-1/2}$;
    \item \textbf{(S4)} By Lemma~\ref{lem:martingaleseq}, we have with probability $1-1/(3\ell^2)$, conditioned on $\EE_{2,\ell}$, 
    \[\martingaleseq\]
    \item \textbf{(S5)} By Lemma~\ref{lem:min_learner}, we have with probability $1-1/(3\ell^2)$, conditioned on $\EE_{1,\ell}\cap\EE_{2,\ell}$, 
    \begin{align}
        &\frac{1}{T_\ell}\bigbrak{\sum_{t=1}^{T_\ell} \E_{\theta\sim \tilde{p}_{t}}\bigbrak{\left\|\theta-\hat{\theta}_{t}\right\|_{x_tx_t^\T}^2}- \inf{\theta\in\Theta_{z_{\ast}}^c} \left\|\theta-\hat{\theta}_{T_\ell+1}\right\|_{V_{T_\ell}}^2}\nonumber\\
        &\leq \minlearner.\nonumber
    \end{align}
    \item \textbf{($C_{T_{\ell}}''$)} By Lemma~\ref{lem:CT''}, conditioned on $\EE_{1,\ell}\cap \EE_{2,\ell}$, we have 
    \[\CTdoubleprime\]
\end{itemize}
Add them altogether, we get that with probability greater than $1-1/\ell^2$ on $\EE_{1,\ell}\cap \EE_{2,\ell}\cap \EE_{4,\ell}$
\begin{align*}
    &\max_{\lambda\in\triangle_\X}\E_{\theta\sim \overline{p}_{T_{\ell}}}\left[\norm{\theta^*-\theta}_{A(\lambda)}^2\right]-\min_{p\in\PP(\Theta_{z_*}^c)}\E_{\theta\sim p}\left[\norm{\theta^*-\theta}_{A(\overline{\lambda}_{T_{\ell}})}^2\right]\\
    &\leq L^2\beta(T_2(\ell), \ell^2)T_\ell^{-1/2}+4d\beta(T_\ell,\ell^2)T_\ell^{-3/4}\\
    &\quad+ C_{3,\ell}^2\sqrt{\log |\X|}T_\ell^{-1/2}+\sqrt{2C_{3,\ell}^2\log \ell^2}T_\ell^{-1/2}+\sqrt{2C_{3,\ell}^2|\X|\log(3T\ell^2)}T_\ell^{-1/2}+C_{3,\ell}^2T_\ell^{-1/4}\\
    &\quad+2C_{3,\ell}^2T_\ell^{-1/2}+\sqrt{\frac{2C_{1,\ell}\log\ell^2}{T_\ell}}\\
    &\quad+\minlearner\\
    &\quad+(C_{3,\ell}+\Delta_{\max})\sqrt{\frac{\beta(T_\ell,\ell^2)}{T_\ell}}.
\end{align*}

Note that each term approaches zero as $T_\ell\to\infty$. By the choice of $T_\ell=2^{\ell}$ in the algorithm, this implies that there exists some $\epsilon_\ell>0$ with $\epsilon_\ell\to 0$ as $\ell\to\infty$ such that for each $\ell$, 
\begin{equation}\label{eqn:dual_gap}
\max_{\lambda\in\triangle_\X}\FF_{\theta\sim \overline{p}_{T_{\ell}}}\left[\norm{\theta^*-\theta}_{A(\lambda)}^2\right]-\min_{p\in\PP(\Theta_{z_*}^c)}\FF_{\theta\sim p}\left[\norm{\theta^*-\theta}_{A(\overline{e}_{T_{\ell}})}^2\right]\leq\epsilon_\ell.
\end{equation}

Now we show how this result leads to the saddle point convergence. Note that
\begin{align*}
    \max_{\lambda\in\triangle_\X}\FF_{\theta\sim \overline{p}_{T_{\ell}}}\left[\norm{\theta^*-\theta}_{A(\lambda)}^2\right] \geq \max_{\lambda\in\triangle_\X}\min_{p\in\PP(\Theta_{z_*}^c)}\FF_{\theta\sim p}[\norm{\theta^*-\theta}_{A(\lambda)}^2]\geq \min_{p\in\PP(\Theta_{z_*}^c)}\FF_{\theta\sim p}\left[\norm{\theta^*-\theta}_{A(\overline{e}_{T_{\ell}})}^2\right],
\end{align*}
so using Equation~\ref{eqn:dual_gap} we have  
\begin{align*}
    \max_{\lambda\in\triangle_\X}\min_{p\in\PP(\Theta_{z_*}^c)}\FF_{\theta\sim p}[\norm{\theta^*-\theta}_{A(\lambda)}^2] - \min_{p\in\PP(\Theta_{z_*}^c)}\FF_{\theta\sim p}\left[\norm{\theta^*-\theta}_{A(\overline{e}_{T_{\ell}})}^2\right] \leq \epsilon_\ell.
\end{align*}
However, note that 
\[\min_{p\in\PP(\Theta_{z_*}^c)}\FF_{\theta\sim p}\left[\norm{\theta^*-\theta}_{A(\overline{e}_{T_{\ell}})}^2\right]=\inf{\theta\in\Theta_{z_*}^c}\norm{\theta^*-\theta}_{A(\overline{e}_{T_{\ell}})}^2 \]
and $\max_{\lambda\in\triangle_\X}\min_{p\in\PP(\Theta_{z_*}^c)}\FF_{\theta\sim p}[\norm{\theta^*-\theta}_{A(\lambda)}^2]=\tau^*$, we have shown that 
\[\tau^*-\inf{\theta\in\Theta_{z_*}^c}\norm{\theta^*-\theta}_{A(\overline{e}_{T_{\ell}})}^2<\epsilon_\ell.\]
\end{proof}

\subsection{Guarantees on the $\max$-learner}\label{sec:max_learner}

In this section, we show that the $\max$-learner gets sublinear regret as $\ell$ gets large. The key idea is that we mix a diminishing amount of $G$-optimal distribution each round, and we show that by its diminishing nature, the mixing of $G$-optimal distribution keeps the regret sublinear. 

\begin{lemma}\label{lem:maxlearner}
Under $\EE_{\ell,2}$, with the choice of $\eta_\lambda=\sqrt{\frac{\log|\X|}{C_{3,\ell}^4T}}$, we have with probability greater than $1-1/\ell^2$, 
\begin{align*}
    &\max_{\lambda\in\triangle_\X}\sum_{t=1}^{T_\ell} \FF_{\theta\sim p_t, x\sim \lambda}\norm{\theta-\hat{\theta}_t}_{xx^\T}^2-\sum_{t=1}^{T_\ell} \FF_{\theta\sim p_t, x\sim \tilde{\lambda}_t}\norm{\theta-\hat{\theta}_t}_{xx^\T}^2\\
    &\leq \eqnmaxlearner.
\end{align*}
\end{lemma}
\begin{proof}
We first show that the statement is true for some fixed $\lambda$, i.e. we would like to show that with probability $1-\delta$, 
\begin{align*}
    &\sum_{t=1}^{T_\ell} \FF_{\theta\sim p_t, x\sim \lambda}\left[\norm{\theta-\hat{\theta}_t}_{xx^\T}^2\right]-\sum_{t=1}^{T_\ell} \FF_{\theta\sim p_t, x\sim \tilde{\lambda}_t}\left[\norm{\theta-\hat{\theta}_t}_{xx^\T}^2\right]\\
    &\leq C_{3,\ell}^2\sqrt{\log |\X|{T_\ell}}+\sqrt{2C_{3,\ell}^2T_\ell\log(1/\delta)}+2C_{3,\ell}^2\sum_{t=1}^{T_\ell}\gamma_t.
\end{align*}
Let $\F_{t-1}$ be the history up to time $t$. Then for any fixed $\lambda$,
\begin{align*}
    \E_{\theta_t}[\FF_{x\sim \lambda}[\|\hat{\theta}_t - \theta_{t}\|^2_{xx^{\top}}]|\F_{t-1}] = \FF_{\theta\sim p_t, x\sim \lambda}[\|\hat{\theta}_t - \theta\|^2_{xx^{\top}}].
\end{align*}
Thus, setting
\begin{align*}
    X_t &= \FF_{x\sim \tilde{\lambda}_t}\left[\norm{\theta_t-\hat{\theta}_t}_{xx^\T}^2\right] -\FF_{x\sim\tilde{\lambda}_t,\theta\sim p_t}\norm{\theta-\hat{\theta}_t}_{xx^\T}^2 \\
    &- \left[\FF_{x\sim\lambda}\left[\norm{\theta_t-\hat{\theta}_t}_{xx^\T}^2\right]-\FF_{x\sim\lambda,\theta\sim p_t}\norm{\theta-\hat{\theta}_t}_{xx^\T}^2\right]\\
\end{align*}
we see that the $X_t$ form a Martingale difference sequence, i.e. $\E[X_t|\F_{t-1}]=0$. Note that for any $\theta\in\Theta$, 
\begin{align*}
    &\FF_{x\sim\lambda_t}\left[\norm{\theta-\hat{\theta}_t}_{xx^\T}^2\right]= \FF_{x\sim\tilde{\lambda}_t}\left[\norm{\theta-\hat{\theta}_t}_{xx^\T}^2\right]+\gamma_t\bigsmile{\FF_{x\sim\lambda_t}\left[\norm{\theta-\hat{\theta}_t}_{xx^\T}^2\right]-\FF_{x\sim \lambda^G}\left[\norm{\theta-\hat{\theta}_t}_{xx^\T}^2\right]}.
\end{align*}
Since under $\EE_{2,\ell}$, we have for any $x\in\X$, $\theta\in\Theta$, any $t\leq T_\ell$, $\norm{\theta-\hat{\theta}_t}_{xx^\T}^2\leq C_{3,\ell}^2$, we have for any $\theta\in\Theta$, 
\begin{align*}
    &\FF_{x\sim\lambda_t}\left[\norm{\theta-\hat{\theta}_t}_{xx^\T}^2\right]\leq \FF_{x\sim\tilde{\lambda}_t}\left[\norm{\theta-\hat{\theta}_t}_{xx^\T}^2\right]+2C_{3,\ell}^2\gamma_t.
\end{align*}
Then we have 
\begin{align}
    &\sum_{t=1}^{T_\ell} \FF_{\theta\sim p_t, x\sim \lambda}\left[\norm{\theta-\hat{\theta}_t}_{xx^\T}^2\right]-\sum_{t=1}^{T_\ell} \FF_{\theta\sim p_t,x\sim\tilde{\lambda}_t}\left[\norm{\theta-\hat{\theta}_t}_{xx^\T}^2\right]\nonumber\\
    &=\sum_{t=1}^{T_\ell} \FF_{x\sim \lambda}\left[\norm{\theta_t-\hat{\theta}_t}_{xx^\T}^2\right]-\sum_{t=1}^{T_\ell} \FF_{x\sim\tilde{\lambda}_t}\left[\norm{\theta_t-\hat{\theta}_t}_{xx^\T}^2\right]\nonumber\\
    &\quad-\left[\sum_{t=1}^{T_\ell} \FF_{x\sim\lambda}\left[\norm{\theta_t-\hat{\theta}_t}_{xx^\T}^2\right]-\sum_{t=1}^{T_\ell} \FF_{x\sim\tilde{\lambda}_t}\left[\norm{\theta_t-\hat{\theta}_t}_{xx^\T}^2\right]\right.\nonumber\\
    &\quad-\left.\bigbrak{\sum_{t=1}^{T_\ell} \FF_{\theta\sim p_t,x\sim\lambda}\norm{\theta-\hat{\theta}_t}_{xx^\T}^2-\sum_{t=1}^{T_\ell} \FF_{x\sim\tilde{\lambda}_t,\theta\sim p_t}\norm{\theta-\hat{\theta}_t}_{xx^\T}^2}\right]\nonumber\\
    &\leq\sum_{t=1}^{T_\ell} \FF_{x\sim \lambda}\left[\norm{\theta_t-\hat{\theta}_t}_{xx^\T}^2\right]-\sum_{t=1}^{T_\ell} \FF_{x\sim\lambda_t}\left[\norm{\theta_t-\hat{\theta}_t}_{xx^\T}^2\right]\nonumber\\
    &\quad-\left[\sum_{t=1}^{T_\ell} \FF_{x\sim\lambda}\left[\norm{\theta_t-\hat{\theta}_t}_{xx^\T}^2\right]-\sum_{t=1}^{T_\ell} \FF_{x\sim\tilde{\lambda}_t}\left[\norm{\theta_t-\hat{\theta}_t}_{xx^\T}^2\right]\right.\nonumber\\
    &\quad-\left.\bigbrak{\sum_{t=1}^{T_\ell} \FF_{\theta\sim p_t,x\sim\lambda}\norm{\theta-\hat{\theta}_t}_{xx^\T}^2-\sum_{t=1}^{T_\ell} \FF_{x\sim\tilde{\lambda}_t,\theta\sim p_t}\norm{\theta-\hat{\theta}_t}_{xx^\T}^2}\right]+2C_{3,\ell}^2\sum_{t=1}^{T_\ell}\gamma_t\label{eqn:max_learner_decomp}
\end{align}
Note that 
\begin{align*}
    &\sum_{t=1}^{T_\ell} \FF_{x\sim\lambda}\left[\norm{\theta_t-\hat{\theta}_t}_{xx^\T}^2\right]-\sum_{t=1}^{T_\ell} \FF_{x\sim\tilde{\lambda}_t}\left[\norm{\theta_t-\hat{\theta}_t}_{xx^\T}^2\right]\\
    &\quad-\bigbrak{\sum_{t=1}^{T_\ell} \FF_{\theta\sim p_t,x\sim\lambda}\norm{\theta-\hat{\theta}_t}_{xx^\T}^2-\sum_{t=1}^{T_\ell} \FF_{x\sim\tilde{\lambda}_t,\theta\sim p_t}\norm{\theta-\hat{\theta}_t}_{xx^\T}^2}=\sum_{t=1}^{T_\ell} X_t. 
\end{align*}
We know that under $\EE_{2,\ell}$, we have for any $x\in\X$, $\theta\in\Theta$, any $t\leq T_\ell$, $\norm{\theta-\hat{\theta}_t}_{xx^\T}^2\leq C_{3,\ell}^2$. Then, for any $t$, $|X_t|\leq 4C_{3,\ell}^2$, so by Azuma-Hoeffding, with probability $1-\delta$, $\sum_{t=1}^{T_\ell} X_t\leq \sqrt{8C_{3,\ell}^2 {T_\ell}\log(1/\delta)}$. Plugging the above and Lemma~\ref{lem:MW_hat} in Equation~\ref{eqn:max_learner_decomp} gives us
\begin{align*}
    &\sum_{t=1}^{T_\ell} \FF_{\theta\sim p_t, x\sim \lambda}\left[\norm{\theta-\hat{\theta}_t}_{xx^\T}^2\right]-\sum_{t=1}^{T_\ell} \FF_{\theta\sim p_t,x\sim\tilde{\lambda}_t}\left[\norm{\theta-\hat{\theta}_t}_{xx^\T}^2\right]\\
    &\leq C_{3,\ell}^2\sqrt{\log |\X|{T_\ell}}+\sqrt{2C_{3,\ell}^2T_\ell\log(1/\delta)}+2C_{3,\ell}^2\sum_{t=1}^{T_\ell}\gamma_t.
\end{align*}
This result holds for any $\lambda$, but in particular we want it to hold for the $\lambda$ which maximizes the reward, so we perform a covering argument on $\lambda$. 

We take an $\epsilon$-cover $\S_{\epsilon}$ of $\triangle_\X$ in $\norm{\cdot}_1$. Then, we know that for any $\lambda\in\triangle_\X$, there is some $\lambda'\in \S_\epsilon$ such that $\norm{\lambda-\lambda'}_1\leq \epsilon$. Let $w_t(\lambda):=\FF_{\theta\sim p_t, x\sim \lambda}\norm{\theta-\hat{\theta}_t}_{xx^\T}^2$. Then, note that for any $t$ and  $\lambda_1,\lambda_2\in\triangle_\X$, 
\begin{align*}
    w(\lambda_1)-w(\lambda_2) &= \FF_{\theta\sim p_t, x\sim \lambda_1}\norm{\theta-\hat{\theta}_t}_{xx^\T}^2 - \FF_{\theta\sim p_t, x\sim \lambda_2}\norm{\theta-\hat{\theta}_t}_{xx^\T}^2\\
    &= \FF_{\theta\sim p_t}\sum_{x}([\lambda_1]_x-[\lambda_2]_x)(x^\top (\theta-\hat{\theta}_t))^2\\
    &\leq C_{3,\ell}^2\FF_{\theta\sim p_t}\sum_{x}([\lambda_1]_x-[\lambda_2]_x)\\
    &= C_{3,\ell}^2\norm{\lambda_1-\lambda_2}_1,
\end{align*}
so $w_t(\lambda)$ is $C_{3,\ell}^2$-Lipschitz for any $t$. Then, assuming that $\bar{\lambda}\in\triangle_\X$ satisfies that 
\[\sum_{t=1}^{T_\ell} \FF_{\theta\sim p_t, x\sim \bar{\lambda}}\norm{\theta-\hat{\theta}_t}_{xx^\T}^2=\max_{\lambda\in\triangle_\X}\sum_{t=1}^{T_\ell} \FF_{\theta\sim p_t, x\sim \lambda}\norm{\theta-\hat{\theta}_t}_{xx^\T}^2,\]
we can find some $\lambda_0\in \S_{\epsilon}$ such that $\norm{\lambda_0-\bar{\lambda}}\leq \epsilon$, so by Lipschitzness of $w_t$ for any $t$, we have 
\begin{align*}
    &\max_{\lambda\in\triangle_\X}\sum_{t=1}^{T_\ell} \FF_{\theta\sim p_t, x\sim \lambda}\norm{\theta-\hat{\theta}_t}_{xx^\T}^2 - \max_{\lambda\in \S_\epsilon}\sum_{t=1}^{T_\ell} \FF_{\theta\sim p_t, x\sim \lambda}\norm{\theta-\hat{\theta}_t}_{xx^\T}^2\\
    &= \sum_{t=1}^{T_\ell} \FF_{\theta\sim p_t, x\sim \bar{\lambda}}\norm{\theta-\hat{\theta}_t}_{xx^\T}^2 - \max_{\lambda\in \S_\epsilon}\sum_{t=1}^{T_\ell} \FF_{\theta\sim p_t, x\sim \lambda}\norm{\theta-\hat{\theta}_t}_{xx^\T}^2\\
    &\leq \sum_{t=1}^{T_\ell} \FF_{\theta\sim p_t, x\sim \bar{\lambda}}\norm{\theta-\hat{\theta}_t}_{xx^\T}^2 - \sum_{t=1}^{T_\ell} \FF_{\theta\sim p_t, x\sim \lambda_0}\norm{\theta-\hat{\theta}_t}_{xx^\T}^2\\
    &\leq C_{3,\ell}^2T_\ell\epsilon.
\end{align*}
Also, let $K=|\X|$. Denote $B_1^K$ as the $l_1$ ball with dimension $K$. We know that for $\epsilon\leq 1$, $N(B_1^K, \norm{\cdot}_1, \epsilon)\leq \left(\frac{3}{\epsilon}\right)^K$. Since $\triangle_\X\subset B_1^K$, we have the covering number \[N(\triangle_\X, \norm{\cdot}_1, \epsilon)\leq N(B_1^k, \norm{\cdot}_1, \epsilon)\leq\left(\frac{3}{\epsilon}\right)^K.\] 
Therefore, $|\S_\epsilon|\leq \left(\frac{3}{\epsilon}\right)^K$. By union bounding over all $\lambda\in\S_\epsilon$, we have with probability at least $1-\delta$, 
\begin{align*}
    &\max_{\lambda\in \S_\epsilon}\sum_{t=1}^{T_\ell} \FF_{\theta\sim p_t, x\sim \lambda}\norm{\theta-\hat{\theta}_t}_{xx^\T}^2-\sum_{t=1}^{T_\ell} \FF_{\theta\sim p_t, x\sim \lambda_t}\norm{\theta-\hat{\theta}_t}_{xx^\T}^2\\
    &\leq C_{3,\ell}^2\sqrt{\log |\X|{T_\ell}}+\sqrt{2C_{3,\ell}^2T_\ell\log(1/(\delta |\S_{\epsilon}|))}+2C_{3,\ell}^2\sum_{t=1}^{T_\ell} \gamma_t\\
    &\leq C_{3,\ell}^2\sqrt{\log |\X|{T_\ell}}+\sqrt{2C_{3,\ell}^2T_\ell|\X|\log(3/(\epsilon\delta))}+2C_{3,\ell}^2\sum_{t=1}^{T_\ell} \gamma_t.
\end{align*}
Combining two displays gives us 
\begin{align*}
    &\max_{\lambda\in\triangle_\X}\sum_{t=1}^{T_\ell} \FF_{\theta\sim p_t, x\sim \lambda}\norm{\theta-\hat{\theta}_t}_{xx^\T}^2-\sum_{t=1}^{T_\ell} \FF_{\theta\sim p_t, x\sim \lambda_t}\norm{\theta-\hat{\theta}_t}_{xx^\T}^2\\
    &\leq C_{3,\ell}^2\sqrt{\log |\X|{T_\ell}}+\sqrt{2C_{3,\ell}^2T_\ell|\X|\log(3/(\delta\epsilon))}+2C_{3,\ell}^2\sum_{t=1}^{T_\ell} \gamma_t+ C_{3,\ell}^2 {T_\ell}\epsilon.
\end{align*}
Taking $\epsilon=1/\sqrt{{T_\ell}}$ and $\delta=1/\ell^2$ gives us the result. 
\end{proof}

\begin{lemma}\label{lem:MW_hat}
Under $\EE_{2,\ell}$, with the choice of $\eta=\sqrt{\frac{\log|\X|}{C_{3,\ell}^4T_\ell}}$, we have for any $\lambda$, 
\[\sum_{t=1}^{T_\ell} \norm{\theta_t-\hat{\theta}_t}_{A(\lambda)}^2-\sum_{t=1}^{T_\ell} \norm{\theta_t-\hat{\theta}_t}_{A(\lambda_t)}^2\leq C_{3,\ell}^2\sqrt{\log |\X|{T_\ell}}.\]
\end{lemma}
\begin{proof}
Let $\ell_t(\lambda)=-\norm{\theta_t-\hat{\theta}_t}_{A(\lambda)}^2$. Then we have 
\[[\nabla_\lambda \ell_t(\lambda_t)]_x=-\norm{\theta_t-\hat{\theta}_t}_{xx^\T}^2=\tilde{g}_{t,x}.\]
Since 
\begin{align*}
    \max_{t\in[T_\ell]}\|\tilde{g}_t\|_\infty=\max_{t\in[T_\ell],x\in\X}\norm{\theta_t-\hat{\theta}_t}_{xx^\T}^2\leq C_{3,\ell}^2, 
\end{align*}
by the guarantee of exponentiated gradient algorithm \cite{orabona2019modern}, we have that for any $\lambda$, 
\[\sum_{t=1}^{T_\ell} [\ell_t(\lambda_t)-\ell_t(\lambda)]\leq \frac{\log |\X|}{\eta}+\frac{\eta {T_\ell}}{2} C_{3,\ell}^4.\]
Plugging in the definition of $\ell_t(\lambda)$, we have 
\begin{align*}
    \sum_{t=1}^{T_\ell} \norm{\theta_t-\hat{\theta}_t}_{A(\lambda)}^2-\sum_{t=1}^{T_\ell} \norm{\theta_t-\hat{\theta}_t}_{A(\lambda_t)}^2\leq \frac{\log |\X|}{\eta}+\frac{\eta {T_\ell}}{2} C_{3,\ell}^4.
\end{align*}
Choosing $\eta=\sqrt{\frac{\log|\X|}{C_{3,\ell}^4T_\ell}}$, we have 
\begin{align*}
    \sum_{t=1}^{T_\ell} \norm{\theta_t-\hat{\theta}_t}_{A(\lambda)}^2-\sum_{t=1}^{T_\ell} \norm{\theta_t-\hat{\theta}_t}_{A(\lambda_t)}^2\leq C_{3,\ell}^2\sqrt{\log |\X|{T_\ell}}. 
\end{align*}
\end{proof}

\subsection{Guarantees on the $\min$-learner}\label{sec:min_learner}
In this section, we show that the $\min$-learner gets sublinear regret as $\ell$ gets large. For the $\min$ learner, we see that the update for the sampling distribution is very similar to the continuous exponential weights updates \cite{bubeck2011introduction}. The difference between our setting and continuous exponential weights is that the space $\Theta_{\hat{z}_t}^c$ is changing each time, so we potentially have a changing action space each time. To overcome this challenge, we first analyze the regret guarantee when we assume access to the true alternative in Lemma~\ref{lem:min_learner}, and use Lemma~\ref{lem:xteq} to argue that the estimate $\Theta_{\hat{z}_t}^c$ is good enough. We state the following guarantee for the $\min$-learner.

\begin{lemma}\label{lem:min_learner}
On event $\EE_{\ell, 1}\cap \EE_{\ell, 2}$, with probability $1-1/\ell^2$,  
\begin{align}
    &\frac{1}{{T_\ell}}\bigbrak{\sum_{t=1}^{T_\ell} \E_{\theta\sim \tilde{p}_{t}}\bigbrak{\left\|\theta-\hat{\theta}_{t}\right\|_{x_tx_t^\T}^2}- \inf{\theta\in\Theta_{z_{\ast}}^c} \left\|\theta-\hat{\theta}_{{T_\ell}+1}\right\|_{V_{{T_\ell}}}^2}\nonumber\\
    &\leq \minlearner.\nonumber
\end{align}
\end{lemma}
\begin{proof}
We begin by a bound that will be useful in our exponential weights analogy. At iteration $t$, we apply Hoeffding's lemma with the following upper bound given $\EE_{\ell, 1}\cap\EE_{\ell, 2}$ and Lemma~\ref{lem:recurs_LS},
\begin{align*}
    &\E_{\theta\sim \tilde{p}_{t}}\bigbrak{\left\|\theta-\hat{\theta}_{t}\right\|_{x_tx_t^\T}^2+\left\|\theta-\hat{\theta}_{t+1}\right\|_{V_{t}}^2-\left\|\theta-\hat{\theta}_{t}\right\|_{V_{t}}^2}\\
    &\leq C_{3,\ell}^2+\E_{\theta\sim \tilde{p}_{t}}\bigbrak{\left\|\theta-\hat{\theta}_{t+1}\right\|_{V_{t-1}}^2-\left\|\theta-\hat{\theta}_{t}\right\|_{V_{t-1}}^2} \tag{$\EE_{\ell, 2}$}\\
    &\leq C_{3,\ell}^2 + 2C_{3,\ell}(C_{1,\ell}+1)\tag{Lemma~\ref{lem:recurs_LS}}\\
    &\leq 4C_{3,\ell}^2.
\end{align*}

At round $t > 1$, we define $W_t=\int_{\theta\in\Theta_{z_{\ast}}^c} \exp\bigsmile{-\eta_{p}\norm{\theta-\hat{\theta}_{t}}_{V_{t-1}}^2}d\theta$ and $W_1$ being a uniform distribution on $\Theta_{z_*}^c$. Then 
{\footnotesize
\begin{align*}
    &\log\frac{W_{t+1}}{W_{t}} \\
    &= \log\frac{\int_{\theta\in\Theta_{z_{\ast}}^c} \exp\bigsmile{-\eta_{p}\left\|\theta-\hat{\theta}_{t+1}\right\|_{V_t}^2}d\theta}{\int_{\theta\in\Theta_{z_*}^c} \exp\bigsmile{-\eta_{p}\left\|\theta-\hat{\theta}_{t}\right\|_{V_{t-1}}^2}d\theta}\\
    &= \log\frac{\int_{\theta\in\Theta_{z_*}^c} \exp\bigsmile{-\eta_{p}\left\|\theta-\hat{\theta}_{t+1}\right\|_{V_{t}}^2-\eta_{p}\left\|\theta-\hat{\theta}_{t}\right\|_{x_tx_t^\T}^2+\eta_{p}\left\|\theta-\hat{\theta}_{t}\right\|_{x_tx_t^\T}^2+\eta_{p}\left\|\theta-\hat{\theta}_{t}\right\|_{V_{t-1}}^2-\eta_{p}\left\|\theta-\hat{\theta}_{t}\right\|_{V_{t-1}}^2}d\theta}{\int_{\theta\in\Theta_{z_*}^c} \exp\bigsmile{-\eta_{p}\left\|\theta-\hat{\theta}_{t}\right\|_{V_{t-1}}^2}d\theta}\\
    &= \log\frac{\int_{\theta\in\Theta_{z_*}^c} \exp\bigsmile{-\eta_{p}\left\|\theta-\hat{\theta}_{t}\right\|_{x_tx_t^\T}^2-\eta_{p}\left\|\theta-\hat{\theta}_{t+1}\right\|_{V_{t}}^2+\eta_{p}\left\|\theta-\hat{\theta}_{t}\right\|_{V_t}^2-\eta_{p}\left\|\theta-\hat{\theta}_{t}\right\|_{V_{t-1}}^2}d\theta}{\int_{\theta\in\Theta_{z_*}^c} \exp\bigsmile{-\eta_{p}\left\|\theta-\hat{\theta}_{t}\right\|_{V_{t-1}}^2}d\theta}\\
    &\leq -\eta_{p} \E_{\theta\sim p_{t}(\Theta_{z_*}^c)}\bigbrak{\left\|\theta-\hat{\theta}_{t}\right\|_{x_tx_t^\T}^2+\left\|\theta-\hat{\theta}_{t+1}\right\|_{V_{t}}^2-\left\|\theta-\hat{\theta}_{t}\right\|_{V_{t}}^2}+\frac{\eta_{p}^2\cdot 4C_{3,\ell}^2}{8}
\end{align*}
}
where the inequality follows from the Hoeffding inequality $\ln \mathbb{E} e^{s X} \leq s \mathbb{E} X+\frac{s^2(a-b)^2}{8}$. By telescoping, we have 
\begin{align*}
    \log \frac{W_{T_\ell+1}}{W_1} & =\ln \frac{W_{T_\ell+1}}{W_{T_\ell}}+\ln \frac{W_{T_\ell}}{W_{{T_\ell}-1}}+\cdots+\ln \frac{W_2}{W_1} \\
    & \leq-\eta_{p}\sum_{t=1}^{T_\ell} \E_{\theta\sim p_{t}(\Theta_{z_*}^c)}\bigbrak{\left\|\theta-\hat{\theta}_{t}\right\|_{x_tx_t^\T}^2+\left\|\theta-\hat{\theta}_{t+1}\right\|_{V_{t}}^2-\left\|\theta-\hat{\theta}_{t}\right\|_{V_{t}}^2}+\frac{{T_\ell}\eta_{p}^2C_{3,\ell}^2}{2}.
\end{align*}
On the other hand, let $\tilde{\theta}=\arg\inf{\theta\in\Theta_{z_{\ast}}^c}\left\|\theta-\hat{\theta}_{{T_\ell}+1}\right\|_{V_{{T_\ell}}}^2$. Let $w_{t}(\theta)=\exp\bigsmile{-\eta_{p}\norm{\theta-\hat{\theta}_{t}}_{V_{t-1}}^2}$. Let $\mc{N}_\gamma:=\{(1-\gamma)\tilde{\theta}+\gamma\theta,\theta\in\Theta_{z_{\ast}}^c\}$ for $\gamma > 0$ that we choose later. We have 
\begin{align*}
\log \frac{W_{{T_\ell}+1}}{W_1} & =\log \left(\frac{\int_{\theta \in \Theta_{z_{\ast}}^c} \exp\bigsmile{-\eta_{p}\left\|\theta-\hat{\theta}_{{T_\ell}+1}\right\|_{V_{{T_\ell}}}^2} d \theta}{\int_{\theta \in \Theta_{z_{\ast}}^c} 1 d \theta}\right) \\
& \geq \log \left(\frac{\int_{\theta \in \mathcal{N}_\gamma} \exp\bigsmile{-\eta_{p}\left\|\theta-\hat{\theta}_{{T_\ell}+1}\right\|_{V_{{T_\ell}}}^2} d \theta}{\int_{\theta \in \Theta_{z_{\ast}}^c} 1 d \theta}\right) \\
& \geq \log \left(\frac{\int_{\theta \in \gamma\Theta_{z_{\ast}}^c} \exp\bigsmile{-\eta_{p}\left\|(1-\gamma)\tilde{\theta}+\theta-\hat{\theta}_{{T_\ell}+1}\right\|_{V_{{T_\ell}}}^2} d \theta}{\int_{\theta \in \Theta_{z_{\ast}}^c} 1 d \theta}\right) \\
&= \log \left(\frac{\int_{\theta \in \Theta_{z_{\ast}}^c} \gamma^d\exp\bigsmile{-\eta_{p}\left\|(1-\gamma)\tilde{\theta}+\gamma\theta-\hat{\theta}_{{T_\ell}+1}\right\|_{V_{{T_\ell}}}^2} d \theta}{\int_{\theta \in \Theta_{z_{\ast}}^c} 1 d \theta}\right) \\
&= \log \left(\frac{\int_{\theta \in \Theta_{z_{\ast}}^c} \gamma^d\exp\bigsmile{-\eta_{p}\left\|(1-\gamma)\tilde{\theta}+\gamma\theta-\hat{\theta}_{{T_\ell}+1}\right\|_{V_{{T_\ell}}}^2} d \theta}{\int_{\theta \in \Theta_{z_{\ast}}^c} 1 d \theta}\right) \\
&\geq \log \left(\frac{\int_{\theta \in \Theta_{z_{\ast}}^c} \gamma^d\exp\bigsmile{-\eta_{p}\left((1-\gamma)\left\|\tilde{\theta}-\hat{\theta}_{{T_\ell}+1}\right\|_{V_{{T_\ell}}}^2+\gamma\left\|\theta-\hat{\theta}_{{T_\ell}+1}\right\|_{V_{{T_\ell}}}^2\right)} d \theta}{\int_{\theta \in \Theta_{z_{\ast}}^c} 1 d \theta}\right) \\
&\geq \log \left(\frac{\int_{\theta \in \Theta_{z_{\ast}}^c} \gamma^d\exp\bigsmile{-\eta_{p}\left((1-\gamma)\left\|\tilde{\theta}-\hat{\theta}_{{T_\ell}+1}\right\|_{V_{{T_\ell}}}^2+\gamma\left\|\theta-\hat{\theta}_{{T_\ell}+1}\right\|_{V_{{T_\ell}}}^2\right)} d \theta}{\int_{\theta \in \Theta_{z_{\ast}}^c} 1 d \theta}\right)\\
&\geq \log \left(\frac{\int_{\theta \in \Theta_{z_{\ast}}^c} \gamma^d\exp\bigsmile{-\eta_{p}\left(\left\|\tilde{\theta}-\hat{\theta}_{{T_\ell}+1}\right\|_{V_{{T_\ell}}}^2+\gamma T_\ell C_{1,\ell}\right)} d \theta}{\int_{\theta \in \Theta_{z_{\ast}}^c} 1 d \theta}\right)\\
&= -\eta_{p}\inf{\theta\in\Theta_{z_{\ast}}^c}\norm{\theta-\hat{\theta}_{{T_\ell}+1}}_{V_{T_\ell}}^2+d\log\gamma-\eta_{p}\gamma T_\ell C_{1,\ell}.
\end{align*}
where the last inequality follows from the fact that for any $\theta\in\Theta$, 
\[\left\|\theta-\hat{\theta}_{{T_\ell}+1}\right\|_{V_{{T_\ell}}}^2=\sum_{t=1}^{T_\ell} (x_t^\T (\theta-\hat{\theta}_{{T_\ell}+1})^2\leq T_\ell C_{3,\ell}^2\]
under $\EE_{2,\ell}$. Combining the two displays gives us 
\begin{align*}
    &-\eta_p\inf{\theta\in\Theta_{z_{\ast}}^c}\norm{\theta-\hat{\theta}_{{T_\ell}+1}}_{V_{T_\ell}}^2+d\log\gamma-\eta_p\gamma T_\ell C_{1,\ell}\\
    &\leq-\eta_{p}\sum_{t=1}^{T_\ell} \E_{\theta\sim p_{t}(\Theta_{z_*}^c)}\bigbrak{\left\|\theta-\hat{\theta}_{t}\right\|_{x_tx_t^\T}^2+\left\|\theta-\hat{\theta}_{t+1}\right\|_{V_{t}}^2-\left\|\theta-\hat{\theta}_{t}\right\|_{V_{t}}^2}+\frac{{T_\ell}\eta_p^2C_{3,\ell}^2}{2}.
\end{align*}
Rearranging, we have 
\begin{align*}
    &\sum_{t=1}^{T_\ell} \E_{\theta\sim p_{t}(\Theta_{z_*}^c)}\bigbrak{\left\|\theta-\hat{\theta}_{t}\right\|_{x_tx_t^\T}^2+\left\|\theta-\hat{\theta}_{t+1}\right\|_{V_{t}}^2-\left\|\theta-\hat{\theta}_{t}\right\|_{V_{t}}^2} - \inf{\theta\in\Theta_{z_{\ast}}^c} \left\|\theta-\hat{\theta}_{{T_\ell}+1}\right\|_{V_{{T_\ell}}}^2\\
    &\leq \frac{\eta_p C_{3,\ell}^2T_\ell}{2}+\frac{d\log (1/\gamma)}{\eta_p}+\gamma T_\ell C_{1,\ell}.
\end{align*}
By choosing $\gamma=\frac{1}{T_\ell C_{1,\ell}}$ and $\eta_p=\sqrt{\frac{d\log(T_\ell C_{1,\ell})}{C_{3,\ell}^2T_\ell}}$, we have 
{
\footnotesize
\[\sum_{t=1}^{T_\ell} \E_{\theta\sim p_{t}(\Theta_{z_*}^c)}\bigbrak{\left\|\theta-\hat{\theta}_{t}\right\|_{x_tx_t^\T}^2+\left\|\theta-\hat{\theta}_{t+1}\right\|_{V_{t}}^2-\left\|\theta-\hat{\theta}_{t}\right\|_{V_{t}}^2} - \inf{\theta\in\Theta_{z_*}^c} \left\|\theta-\hat{\theta}_{{T_\ell}+1}\right\|_{V_{{T_\ell}}}^2\leq\sqrt{T_\ell C_{3,\ell}^2d\log(T_\ell C_{1,\ell})},\]
}
so
{
\footnotesize
\[\frac{1}{{T_\ell}}\bigbrak{\sum_{t=1}^{T_\ell} \E_{\theta\sim p_{t}(\Theta_{z_*}^c)}\bigbrak{\left\|\theta-\hat{\theta}_{t}\right\|_{x_tx_t^\T}^2+\left\|\theta-\hat{\theta}_{t+1}\right\|_{V_{t}}^2-\left\|\theta-\hat{\theta}_{t}\right\|_{V_{t}}^2} - \inf{\theta\in\Theta_{z_{\ast}}^c} \left\|\theta-\hat{\theta}_{{T_\ell}+1}\right\|_{V_{{T_\ell}}}^2}\leq \sqrt{\frac{C_{3,\ell}^2d\log(T_\ell C_{1,\ell})}{{T_\ell}}}.\]
}
In other words, 
\begin{align}
    &\frac{1}{{T_\ell}}\bigbrak{\sum_{t=1}^{T_\ell} \E_{\theta\sim \tilde{p}_{t}}\bigbrak{\left\|\theta-\hat{\theta}_{t}\right\|_{x_tx_t^\T}^2}- \inf{\theta\in\Theta_{z_{\ast}}^c} \left\|\theta-\hat{\theta}_{{T_\ell}+1}\right\|_{V_{{T_\ell}}}^2}\nonumber\\
    &\leq \sqrt{\frac{C_{3,\ell}^2d\log(T_\ell C_{1,\ell})}{{T_\ell}}}+\frac{1}{{T_\ell}}\sum_{t=1}^{T_\ell} \E_{\theta\sim \tilde{p}_{t}}\bigbrak{\left\|\theta-\hat{\theta}_{t+1}\right\|_{V_{t}}^2-\left\|\theta-\hat{\theta}_{t}\right\|_{V_{t}}^2}.\nonumber
\end{align}
By Lemma~\ref{lem:CT}, we have with probability $1-1/\ell^2$, 
\[\CT.\] 
Combining the above two displays gives us with probability $1-1/\ell^2$, 
\begin{align}
    &\frac{1}{{T_\ell}}\bigbrak{\sum_{t=1}^{T_\ell} \E_{\theta\sim \tilde{p}_{t}}\bigbrak{\left\|\theta-\hat{\theta}_{t}\right\|_{x_tx_t^\T}^2}- \inf{\theta\in\Theta_{z_{\ast}}^c} \left\|\theta-\hat{\theta}_{{T_\ell}+1}\right\|_{V_{{T_\ell}}}^2}\nonumber\\
    &\leq \minlearner.\nonumber
\end{align}
\end{proof}

\subsection{Approximation Guarantees}\label{sec:approx_guarantee}

In this section, we present several technical lemmas bounding the terms related to the approximation error of $\hat{\theta}_{t}$ to $\theta^*$ in each iteration $t$. More specifically, these lemmas show upper bound on the terms in the decomposition in the proof of lemma \ref{lem:dual_gap_bound}. 
\begin{lemma}[S4]\label{lem:martingaleseq}
Under $\EE_{2,\ell}$, with probability $1-1/\ell^2$, 
\[\martingaleseq.\]
\end{lemma}
\begin{proof}
Define 
$M_t=\FF_{\theta\sim\tilde{p}_t}\bigbrak{\norm{\theta-\hat{\theta}_t}_{x_tx_t^\T}^2}$. Note that $$\E_{x_t}[M_t|\F_{t-1}]=\FF_{\theta\sim \tilde{p}_{t}}\bigbrak{\left\|\theta-\hat{\theta}_t\right\|_{A(\tilde{\lambda}_t)}^2},$$ so $\tilde{M}_t=M_t-\FF_{\theta\sim \tilde{p}_{t}}\bigbrak{\left\|\theta-\hat{\theta}_t\right\|_{A(\tilde{\lambda}_t)}^2}$ is a mean-zero martingale. Also, under $\EE_{2,\ell}$, $|M_t|\leq C_{1,\ell}$, then by Azuma-Hoeffding, we have with probability at least $1-\frac{1}{\ell^2}$, $\sum_{t=1}^{T_\ell} \tilde{M}_t\leq \sqrt{2C_{1,\ell}{T_\ell}\log \ell^2}$, so 
\[\martingaleseq.\]
\end{proof}

\begin{lemma}[$C_{T_\ell}$]\label{lem:CT}
Under $\EE_{1,\ell}\cap \EE_{2,\ell}$, with probability $1-1/\ell^2$, 
\[\CT.\] 
\end{lemma}
\begin{proof}
We first consider some round $t$ and some $\theta$. By Lemma~\ref{lem:recurs_LS}, 
\begin{align*}
    &\left\|\theta-\hat{\theta}_{t}\right\|_{V_{t-1}}^2-\left\|\theta-\hat{\theta}_{t+1}\right\|_{V_{t-1}}^2\leq 2C_{3,\ell}(y_t - x_t^{\top}\hat{\theta}_t).
\end{align*}

Therefore, 
\begin{align}
    &\frac{1}{{T_\ell}}\sum_{t=1}^{T_\ell} \FF_{\theta\sim \tilde{p}_{t}}\bigbrak{\left\|\theta-\hat{\theta}_{t}\right\|_{V_{t-1}}^2-\left\|\theta-\hat{\theta}_{t+1}\right\|_{V_{t-1}}^2}\leq \frac{2C_{3,\ell}}{{T_\ell}}\sum_{t=1}^{T_\ell} (y_{t} - x_{t}^{\top}\hat{\theta}_{t}).\label{eqn:reg_LinUCB}
\end{align}
Now, note that 
\begin{align*}
    y_t-x_t^\T\hat{\theta}_{t} &= x_t^\T (\theta^*-\hat{\theta}_{t}) + \epsilon_t\\
    &\leq \norm{x_t}_{V_{t-1}^{-1}}\norm{\theta^*-\hat{\theta}_{t}}_{V_{t-1}} + \epsilon_t\\
    &\leq \norm{x_t}_{V_{t-1}^{-1}}\sqrt{\beta(t,\ell^2)} + \epsilon_t.\tag{by $\EE_{1,\ell}$}
\end{align*}
Note that since $\epsilon_t\sim N(0,1)$ is 1-subGaussian, by Azuma-Hoeffding, we have with probability $1-1/\ell^2$, 
\[\sum_{t=1}^{T_\ell} \epsilon_t\leq \sqrt{2T_\ell\log(\ell^2)}.\]

By summing it from $1$ to ${T_\ell}$, we have under $\EE_{1,\ell}$, with probability $1-1/\ell^2$, 
\begin{align*}
    \sum_{t=1}^{{T_\ell}} (y_{t} - x_{t}^{\top}\hat{\theta}_{t}) &\leq \sum_{t=1}^{{T_\ell}} \sqrt{\beta(t,\ell^2)}\norm{x_t}_{V_{t-1}^{-1}}+\sum_{t=1}^{T_\ell} \epsilon_t\\
    &\leq \sum_{t=1}^{{T_\ell}} \sqrt{\beta(t,\ell^2)}\norm{x_t}_{V_{t-1}^{-1}}+\sqrt{2T_\ell\log(\ell^2)}\\
    &\leq \sqrt{{T_\ell}\sum_{t=1}^{{T_\ell}} \beta(t,\ell^2)\norm{x_t}_{V_{t-1}^{-1}}^2}+\sqrt{2T_\ell\log(\ell^2)}\tag{by Cauchy-Schwarz}\\
    &\leq \sqrt{{T_\ell}\beta({T_\ell},\ell^2)\sum_{t=1}^{{T_\ell}} \norm{x_t}_{V_{t-1}^{-1}}^2}+\sqrt{2T_\ell\log(\ell^2)}\tag{by Cauchy-Schwarz}\\
    &\leq \sqrt{{T_\ell}\beta({T_\ell},\ell^2)2d\log\bigsmile{\frac{d+T_\ell L^2}{d}}}+\sqrt{2T_\ell\log(\ell^2)}\tag{by Elliptical potential lemma \cite{abbasi2011improved}}.
\end{align*}
Plugging this in Equation~\ref{eqn:reg_LinUCB} gives the result. 
\end{proof}

\begin{lemma}[$C_{T_\ell}'$]\label{lem:CT'} 
Under $\EE_{1,\ell}\cap \EE_{2,\ell}$, we have
\begin{align*}
    &\max_{\lambda\in\triangle_\X}\frac{1}{{T_\ell}}\sum_{t=1}^{T_\ell}\bF_{\theta\sim p_t}\left[\norm{\theta^*-\theta}_{A(\lambda)}^2\right]-\max_{\lambda\in\triangle_\X}\frac{1}{{T_\ell}}\sum_{t=1}^{T_\ell}\bF_{\theta\sim p_t}\left[\norm{\hat{\theta}_t-\theta}_{A(\lambda)}^2\right]\\
    &\leq\CTPrime.
\end{align*}
\end{lemma}
\begin{proof}
We have 
\begin{align*}
    &\max_{\lambda\in\triangle_\X}\frac{1}{{T_\ell}}\sum_{t=1}^{T_\ell}\bF_{\theta\sim p_t}\left[\norm{\theta^*-\theta}_{A(\lambda)}^2\right]-\max_{\lambda\in\triangle_\X}\frac{1}{{T_\ell}}\sum_{t=1}^{T_\ell}\bF_{\theta\sim p_t}\left[\norm{\hat{\theta}_t-\theta}_{A(\lambda)}^2\right]\\
    &\leq \max_{\lambda\in\triangle_\X}\frac{1}{{T_\ell}}\sum_{t=1}^{T_\ell}\bF_{\theta\sim p_t}\left[\norm{\theta^*-\theta}_{A(\lambda)}^2-\norm{\hat{\theta}_t-\theta}_{A(\lambda)}^2\right].
\end{align*}
We fix some $\theta$ and $\lambda$. Note that 
\begin{align*}
    &\norm{\theta^*-\theta}_{A(\lambda)}^2-\norm{\hat{\theta}_t-\theta}_{A(\lambda)}^2\\
    &= (\theta^*+\hat{\theta}_t-2\theta)^\T A(\lambda)(\theta^*-\hat{\theta}_t)\\
    &= \sum_{x\in\X}\lambda_x(\theta^*+\hat{\theta}_t-2\theta)^\T xx^\T(\theta^*-\hat{\theta}_t)\\
    &\leq \max_{x\in\X}(\theta^*+\hat{\theta}_t-2\theta)^\T xx^\T(\theta^*-\hat{\theta}_t)\\
    &\leq (C_{3,\ell} + \Delta_{\max})\max_{x\in\X}x^\T(\theta^*-\hat{\theta}_t).
\end{align*}
Therefore, 
\begin{align}
    &\max_{\lambda\in\triangle_\X}\frac{1}{{T_\ell}}\sum_{t=1}^{T_\ell}\bF_{\theta\sim p_t}\left[\norm{\theta^*-\theta}_{A(\lambda)}^2-\norm{\hat{\theta}_t-\theta}_{A(\lambda)}^2\right]\leq (C_{3,\ell} + \Delta_{\max}) \max_{x\in\X}\frac{1}{{T_\ell}}\sum_{t=1}^{T_\ell} \inner{\hat{\theta}_t-\theta^*}{x}.\label{eqn:max_diff}
\end{align}
By Lemma~\ref{lem:xthetahat}, under $\EE_{3,\ell}\cap \EE_{1,\ell}$, for any $t\geq T_2(\ell)+1$, we have for any $x\in\X$, 
\[\inner{x}{\hat{\theta}_t-\theta^*}\leq \frac{d}{t^{3/4}}\beta(t,\ell^2).\]
Also, by Lemma~\ref{lem:EE_2_true}, under $\EE_{1,\ell}$, we have for any $t\geq 1$, 
\[\inner{x}{\hat{\theta}_t-\theta^*}\leq L^2\beta(t,\ell^2).\]
Therefore, 
\begin{align*}
    &\max_{x\in\X}\frac{1}{{T_\ell}}\sum_{t=1}^{T_\ell} \inner{\hat{\theta}_t-\theta^*}{x}\\
    &\leq \max_{x\in\X}\frac{1}{{T_\ell}}\left[\sum_{t=1}^{T_2(\ell)} \inner{\hat{\theta}_t-\theta^*}{x}+\sum_{t=T_2(\ell)+1}^{{T_\ell}} \inner{\hat{\theta}_t-\theta^*}{x}\right]\\
    &\leq \frac{1}{{T_\ell}}\left[T_2(\ell)L^2\beta(T_2(\ell),\ell^2)+\sum_{t=T_2(\ell)+1}^{{T_\ell}}\frac{d}{t^{3/4}}\beta(t,\ell^2)\right]\tag{by Lemma~\ref{lem:EE_2_true} and~\ref{lem:xthetahat}}\\
    &\leq \frac{1}{{T_\ell}}\left[T_2(\ell)L^2\beta(T_2(\ell),\ell^2)+d\beta({T_\ell},\ell^2)\int_{t=T_2(\ell)}^{{T_\ell}}t^{-3/4}dt\right]\\
    &= \frac{1}{{T_\ell}}\left[T_2(\ell)L^2\beta(T_2(\ell),\ell^2)+d\beta({T_\ell},\ell^2)(4T_\ell^{1/4}-4T_2(\ell)^{1/4})\right]\\
    &\leq \frac{T_2(\ell)L^2\beta(T_2(\ell),\ell^2)}{{T_\ell}}+4d\beta({T_\ell},\ell^2){T_\ell}^{-3/4}.
\end{align*}
Plugging this in Equation~\ref{eqn:max_diff} gives us 
\begin{align}
    &\max_{\lambda\in\triangle_\X}\frac{1}{{T_\ell}}\sum_{t=1}^{T_\ell}\bF_{\theta\sim p_t}\left[\norm{\theta^*-\theta}_{A(\lambda)}^2-\norm{\hat{\theta}_t-\theta}_{A(\lambda)}^2\right]\leq \CTPrime.\nonumber
\end{align}
\end{proof}

\begin{lemma}[$C_{T_\ell}''$]\label{lem:CT''}
Assume that $\Theta$ is closed. Then, we have under $\EE_{1,\ell}\cap\EE_{2,\ell}$, 
\[\CTdoubleprime.\]
\end{lemma}
\begin{proof}
Let $\theta_1:=\arg\inf{\theta\in\Theta_{z_*}^c}\norm{\theta-\hat{\theta}_{{T_\ell}+1}}_{V_{T_\ell}}^2$ and $\theta_2:=\arg\inf{\theta\in\Theta_{z_*}^c}\norm{\theta-\theta^*}_{V_{T_\ell}}^2$. We have
\begin{align*}
    &\inf{\theta\in\Theta_{z_*}^c} \left\|\theta-\hat{\theta}_{{T_\ell}+1}\right\|_{V_{{T_\ell}}}^2-\inf{\theta\in\Theta_{z_*}^c}\norm{\theta^*-\theta}_{V_{{T_\ell}}}^2\\
    &\leq \norm{\hat{\theta}_{{T_\ell}+1}-\theta_2}_{V_{{T_\ell}}}^2-\norm{\theta^*-\theta_2}_{V_{{T_\ell}}}^2\\
    &=\left( \norm{\hat{\theta}_{{T_\ell}+1}-\theta_2}_{V_{{T_\ell}}}-\norm{\theta^*-\theta_2}_{V_{{T_\ell}}}\right)\left(\norm{\hat{\theta}_{{T_\ell}+1}-\theta_2}_{V_{{T_\ell}}}+\norm{\theta^*-\theta_2}_{V_{{T_\ell}}}\right).\\
    &\leq \norm{\hat{\theta}_{{T_\ell}+1}-\theta_*}_{V_{{T_\ell}}}\left(\norm{\hat{\theta}_{{T_\ell}+1}-\theta_2}_{V_{{T_\ell}}}+\norm{\theta^*-\theta_2}_{V_{{T_\ell}}}\right).
\end{align*}

Note that under $\EE_{2,\ell}$, 
\begin{align*}
    \norm{\hat{\theta}_{{T_\ell}+1}-\theta_1}_{V_{{T_\ell}}}&= \sqrt{\sum_{t=1}^{T_\ell} (x_t^\T(\hat{\theta}_{{T_\ell}+1}-\theta_1))^2}\leq C_{3,\ell}\sqrt{{T_\ell}};\\
    \norm{\theta^*-\theta_2}_{V_{{T_\ell}}}&= \sqrt{\sum_{t=1}^{{T_\ell}} (x_t^\T(\theta^*-\theta_2))^2}\leq \Delta_{\max}\sqrt{{T_\ell}}.
\end{align*}
Therefore, 
\begin{align*}
    &\inf{\theta\in\Theta_{z_*}^c} \left\|\theta-\hat{\theta}_{{T_\ell}+1}\right\|_{V_{{T_\ell}}}^2-\inf{\theta\in\Theta_{z_*}^c}\norm{\theta^*-\theta}_{V_{{T_\ell}}}^2\\
    &\leq (C_{3,\ell}+\Delta_{\max})\sqrt{{T_\ell}}\norm{\hat{\theta}_{{T_\ell}+1}-\theta^*}_{V_{{T_\ell}}}\\
    &\leq (C_{3,\ell}+\Delta_{\max})\sqrt{{T_\ell}\beta({T_\ell},\ell^2)}\tag{by $\EE_{1,\ell}$}.
\end{align*}
\end{proof}


We use the above lemma to bound the term that relates $\tilde{p}_t$ to $p_t$. 
\begin{lemma}[$\tilde{p}_t$ to $p_t$]\label{lem:ptversusptilde}
Under $\EE_{2,\ell}\cap\EE_{4,\ell}$ for ${T_\ell}\geq T_0$, 
\[\ptvsptilde.\]
\end{lemma}
\begin{proof}
Note that $\tilde{p}_t=p_t$ under $\EE_{4,\ell}$, 
\begin{align*}
    &\frac{1}{{T_\ell}}\sum_{t=1}^{{T_\ell}} \bigsmile{\FF_{\theta\sim p_{t}}\left[\norm{\theta-\hat{\theta}_t}_{A(\tilde{\lambda}_t)}^2\right]- \FF_{\theta\sim \tilde{p}_{t}}\bigbrak{\left\|\theta-\hat{\theta}_t\right\|_{A(\tilde{\lambda}_t)}^2}}\\
    &=\frac{1}{{T_\ell}}\sum_{t=1}^{T_0(\ell)} \bigsmile{\FF_{\theta\sim p_{t}}\left[\norm{\theta-\hat{\theta}_t}_{A(\tilde{\lambda}_t)}^2\right]- \FF_{\theta\sim \tilde{p}_{t}}\bigbrak{\left\|\theta-\hat{\theta}_t\right\|_{A(\tilde{\lambda}_t)}^2}}\\
    &\quad+\frac{1}{{T_\ell}}\sum_{t=T_0(\ell)+1}^{{T_\ell}} \bigsmile{\FF_{\theta\sim p_{t}}\left[\norm{\theta-\hat{\theta}_t}_{A(\tilde{\lambda}_t)}^2\right]- \FF_{\theta\sim \tilde{p}_{t}}\bigbrak{\left\|\theta-\hat{\theta}_t\right\|_{A(\tilde{\lambda}_t)}^2}}\\
    &=\frac{1}{{T_\ell}}\sum_{t=1}^{T_0(\ell)} \bigsmile{\FF_{\theta\sim p_{t}}\left[\norm{\theta-\hat{\theta}_t}_{A(\tilde{\lambda}_t)}^2\right]- \FF_{\theta\sim \tilde{p}_{t}}\bigbrak{\left\|\theta-\hat{\theta}_t\right\|_{A(\tilde{\lambda}_t)}^2}}.
\end{align*}
Since for any $\theta\in\Theta$, under $\EE_{2,\ell}$, 
\begin{align*}
    &\norm{\theta-\hat{\theta}_t}_{A(\tilde{\lambda}_t)}^2= \sum_{x\in\X}\tilde{\lambda}_{t,x}\norm{\theta-\hat{\theta}_t}_{xx^\T}^2\leq \max_{x\in\X} \norm{\theta-\hat{\theta}_t}_{xx^\T}^2\leq C_{3,\ell}^2, 
\end{align*}
we have 
\[\frac{1}{{T_\ell}}\sum_{t=1}^{T_0(\ell)} \bigsmile{\FF_{\theta\sim p_{t}}\left[\norm{\theta-\hat{\theta}_t}_{A(\tilde{\lambda}_t)}^2\right]- \FF_{\theta\sim \tilde{p}_{t}}\bigbrak{\left\|\theta-\hat{\theta}_t\right\|_{A(\tilde{\lambda}_t)}^2}}\leq \frac{2C_{3,\ell}^2T_0(\ell)}{{T_\ell}}.\]
\end{proof}

\subsection{Guarantees on sampling and learning the estimate}\label{sec:sampling}
In this section we provide some general guarantees on sampling together with a threshold after which each arm gets enough samples and . Consider a setting where at each time we receive a distribution $\tilde{\lambda} = (1-\gamma_t)\lambda_t + \gamma_t P$ for a fixed distribution $P$. 

\begin{lemma}\label{lem:sampling_general}
Fix a distribution $P$ on $\mc{X}$ with full support. On an event that is true with probability greater than $1-\delta$, for any $0<\alpha<1/2 $ there exists a $T_1 := T_1(\alpha, \delta, T)$ such that for any $t \geq T_1$,
    \begin{align*} 
        V_t \geq  \frac{c}{1-\alpha}A(P)t^{1-\alpha}.
    \end{align*}
\end{lemma}
\begin{proof}
    Fix $x\in \mc{X}$, let $N_{t,x} = \sum_{s=1}^t Z_s$ where $Z_s = 1$ if $x_s = x$ else 0. Then,
    $V_t = \sum_{x\in \mc{X}} \sum_{s=1}^t Z_s xx^{\top}$. We assume that $\gamma_s = 1/s^{\alpha}, s\geq 1$.

    Note that $\P(Z_s = 1|\mc{F}_{s-1}) = (1-\gamma_s)\lambda_{s,x} + \gamma_s P_x$. So for $t > 1$,
    \begin{align*}
        \P\left(\sum_{s=1}^t Z_s \leq c P_x\sum_{s=1}^t \gamma_s\right) 
        &=  \P\left(\sum_{s=1}^t Z_s - (1-\gamma_s)\lambda_{s,x} - \gamma_s P_x  \leq \sum_{s=1}^t c P_x\gamma_s -(1-\gamma_s)\lambda_{s,x} - \gamma_s P_x\right)\\
        &= \P\left(\sum_{s=1}^t Z_s - (1-\gamma_s)\lambda_{s,x} - \gamma_s P_x  \leq \sum_{s=1}^t (c -1) P_x \gamma_s -(1-\gamma_s)\lambda_{s,x} \right)\\
        &\leq \P\left(\sum_{s=1}^t Z_s - (1-\gamma_s)\lambda_{s,x} - \gamma_s P_x  \leq \sum_{s=1}^t (c-1) P_x \gamma_s \right)\\
        &\leq \P\left(\sum_{s=1}^t Z_s - (1-\gamma_s)\lambda_{s,x} - \gamma_s P_x  \leq -\sum_{s=1}^t (1-c) P_x \gamma_s \right)\\
        &\leq \exp\left(-\frac{1}{t}\left( \sum_{s=1}^t (1-c) P_x \gamma_s\right)^2\right) \tag{Azuma-Hoeffding }\\
        &= \exp\left(-\left( \frac{(1-c) P_x}{\sqrt{t}}\sum_{s=1}^t  \gamma_s\right)^2\right) \\
        &\leq \exp\left(-\left( \frac{(1-c) P_x}{\sqrt{t}}\frac{t^{1-\alpha} - 1}{1-\alpha}\right)^2\right) \tag{$\sum_{s=1}^t \frac{1}{s^{\alpha}}\geq\frac{t^{1-\alpha}-1}{1- \alpha}$}\\
        &\leq \exp\left(-\left( (1-c) P_x\frac{t^{1/2-\alpha} - t^{-1/2}}{1-\alpha}\right)^2\right) \\
        &\leq \exp\left(-\left( \frac{(1-c) P_x}{2(1-\alpha)}t^{1/2-\alpha}\right)^2\right)\tag{$t^{1/2-\alpha} - t^{-1/2} > \frac{1}{2}t^{1/2-\alpha}$, $t\geq 2$} \\
        &\leq \exp\left(-\left( \frac{(1-c) P_x}{2(1-\alpha)}\right)^2t^{1-2\alpha}\right)
    \end{align*}    
    This implies that with the sequence $\gamma_s = 1/s^{\alpha}, \alpha < 1/2$ (to ensure $1-2\alpha > 0$), with probability greater than $1-\delta$ we have
    \begin{align*}
        N_{t,x} = \sum_{s=1}^t Z_s \geq c P_x\sum_{s=1}^t \gamma_s\geq \frac{cP_x}{1-\alpha} (t^{1-\alpha} - 1) \quad\text{whenever}\quad t\geq \left(\frac{2(1-\alpha)\sqrt{\log(1/\delta)}}{(1-c)P_x}\right)^{\frac{2}{1-2\alpha}}.
    \end{align*}
\end{proof}
The lemma below states that there exists some time $T_2$ such that all the arms get enough samples. 
\begin{lemma}\label{lem:sampling}
For $T_2(\ell)=\max_{x\in\X}\left(\frac{6\sqrt{\log(|\X|T_\ell\ell^2)}}{\lambda^G_x}\right)^4$, we have 
\begin{align*}
    \P\left(\EE_{3,\ell}\right) \geq 1-1/\ell^2.
\end{align*}
\end{lemma}
\begin{proof}
By Lemma~\ref{lem:sampling_general} with a choice of $c=1-\alpha$, $\alpha=\frac{1}{4}$, $\delta=\frac{1}{|\X|T_\ell\ell^2}$, and $P=\lambda^G$, we have for any $t\geq \left(\frac{2(1-\alpha)\sqrt{\log(1/\delta)}}{(1-c)P_x}\right)^{\frac{2}{1-2\alpha}}=\left(\frac{6\sqrt{\log(|\X|T_\ell\ell^2)}}{\lambda^G_x}\right)^4$, we have $\P(V_t\geq t^{3/4}A(\lambda^G))\geq 1-\frac{1}{|\X|T_\ell\ell^2}$. 
Let $T_2(\ell):=\max_{x\in\X}\left(\frac{6\sqrt{\log(|\X|T_\ell\ell^2)}}{\lambda^G_x}\right)^4$, union bounding for $t\in [T_2,T_\ell]$ and $x\in\X$ gives the result. 
\end{proof}

\begin{lemma}\label{lem:xthetahat}
Under $\EE_{3,\ell}\cap \EE_{1,\ell}$, for any $t\geq T_2(\ell)+1$, we have for any $x\in\X$, 
\[\inner{x}{\hat{\theta}_t-\theta^*}\leq \frac{d}{t^{3/4}}\beta(t,\ell^2).\]
\end{lemma}
\begin{proof}
Let $N_{t,x}$ be the number of times arm $x$ gets pulled at round $t$. By Lemma~\ref{lem:sampling}, for $t\geq T_2(\ell)+1$, under $\EE_{3,\ell}$, we have 
\begin{align*}
    V_{t-1}=\sum_{x\in\X}N_{t-1,x}xx^\T \geq t^{3/4}A(\lambda^G).
\end{align*}
Therefore, for any $x\in\X$, 
\begin{align*}
    \norm{x}_{V_{t-1}^{-1}}^2 &\leq \frac{1}{t^{3/4}}\norm{x}_{A(\lambda^G)^{-1}}^2\leq \frac{d}{t^{3/4}}
\end{align*}
by Kiefer-Wolfowitz. Therefore, under $\EE_{1,\ell}$, for any $x\in\X$, 
\begin{align*}
    \inner{x}{\hat{\theta}_t-\theta^*}&\leq\norm{x}_{V_{t-1}^{-1}}^2\norm{\hat{\theta}_t-\theta^*}_{V_{t-1}}^2\\
    &\leq \frac{d}{t^{3/4}}\norm{\hat{\theta}_t-\theta^*}_{V_{t-1}}^2\\
    &\leq \frac{d}{t^{3/4}}\beta(t,\ell^2). 
\end{align*}

\end{proof}
The following lemma provides a guarantee that we eventually finds $z_{\ast}$. 
\begin{lemma}\label{lem:xteq} For $T_0(\ell)=\max\left\{\left(\frac{d\beta(T_\ell,\ell^2)\max_{z\in\ZZ} \norm{z}_1}{\Delta_{\min}}\right)^{4/3}, T_2(\ell)+1\right\}$, we have $\P(\EE_{4,\ell}|\EE_{1,\ell}\cap\EE_{3,\ell}) \geq 1-1/\ell^2$.
\end{lemma}
\begin{proof}
By Lemma~\ref{lem:xthetahat}, we know that for any $t\geq T_2(\ell)+1$, under $\EE_{1,\ell}\cap \EE_{3,\ell}$ we have for any $x\in\X$, \[\inner{x}{\hat{\theta}_t-\theta^*}\leq \frac{d}{t^{3/4}}\beta(t,\ell^2).\] Since the span of $\ZZ$ is in the subset of $\X$, for any $z\in\ZZ$, we write $z_*-z=\sum_{x\in\X}{\alpha_{z,x}x}$. Then 
\begin{align*}
    (z_*-z)^\T(\theta_*-\hat{\theta}_t) &= \sum_{x\in\X} \alpha_{z,x}x^\T(\theta_*-\hat{\theta}_t)\\
    &\leq \sum_{x\in\X} \alpha_{z,x}\frac{d}{t^{3/4}}\beta(t,\ell^2)\\
    &\leq \max_{z\in\ZZ} \norm{z}_1\frac{d}{t^{3/4}}\beta(t,\ell^2).
\end{align*}

Then, for any $t>\left(\frac{d\beta(t,\ell^2)\max_{z\in\ZZ} \norm{z}_1}{\Delta_{\min}}\right)^{4/3}$, we have 
\[\max_{z\in\ZZ} \norm{z}_1\frac{d}{t^{3/4}}\beta(t,\ell^2)<\Delta_{\min},\]
which implies that for any $z$, 
\begin{align*}
    &(z_*-z)^\T(\theta_*-\hat{\theta}_t)<\Delta_{\min}\\
    \Rightarrow &(z_*-z)^\T(\hat{\theta}_t-\theta_*)>-\Delta_{\min}\\
    \Rightarrow &(z_*-z)^\T\hat{\theta}_t > 0,
\end{align*}
which implies that $\hat{z}_t=z_*$. 
\end{proof}

\section{Bounds and Events that Hold True Each Round}\label{sec:bounds_true}

The following lemma states an anytime confidence bound for the least-squares estimator. It is a restatement of Theorem 20.5 of \cite{lattimore2020bandit} in our setting. 
\begin{lemma}[$\EE_{1,\ell}$]\label{lem:EE_1_true}
With probability $1-1/\ell^2$, for all $t$, we have 
\[\norm{\hat{\theta}_t-\theta^*}_{V_{t-1}}^2\leq B+\sqrt{2\log(\ell^2)+d\log\left(\frac{d+tL^2}{d}\right)}.\]
\end{lemma}
\begin{proof}
    Follows from Theorem 20.5 of \cite{lattimore2020bandit}.
\end{proof}

\begin{lemma}[$\EE_{2,\ell}$]\label{lem:EE_2_true}
Under $\EE_{1,\ell}$, we have for any $x\in\X$ and any $t\in [1,{T_\ell}]$, $\inner{x}{\hat{\theta}_t}\leq \Delta_{\max}+L^2\beta({T_\ell},\ell^2)$. 
\end{lemma}
\begin{proof}
For any $x\in\X$, 
\begin{align*}
    \inner{x}{\hat{\theta}_t} &= \inner{x}{\theta^*}+\inner{x}{\hat{\theta}_t-\theta^*}\\
    &\leq \Delta_{\max}+\norm{x}_{V_{t-1}^{-1}}^2\norm{\hat{\theta}_t-\theta^*}_{V_{t-1}}^2\\
    &\leq \Delta_{\max}+\norm{x}_{V_{t-1}^{-1}}^2\beta(t,\ell^2).\tag{under $\EE_{1,\ell}$}
\end{align*}
Since we have 
\[V_{t-1}=V_0+\sum_{s=1}^{t-1}x_sx_s^\T,\]
for $V_0=I$, we have the minimum eigenvalue 
$\sigma_{\min}(V_{t-1})\geq \sigma_{\min}(V_0)+\sigma_{\min}\left(\sum_{s=1}^{t-1}x_sx_s^\T\right)\geq 1$, so 
\[\sigma_{\max}(V_{t-1}^{-1})=\frac{1}{\sigma_{\min}(V_{t-1})}\leq 1,\]
which implies that 
\[\max_{x\in\X}\norm{x}_{V_{t-1}^{-1}}^2\leq \sigma_{\max}(V_{t-1}^{-1})\max_{x\in\X}\norm{x}_2^2\leq L^2.\]
Therefore, 
\[\inner{x}{\hat{\theta}_t}\leq \Delta_{\max}+L^2\beta(t,\ell^2)\leq \Delta_{\max}+L^2\beta(T_\ell,\ell^2).\]
\end{proof}

\section{Technical Lemmas}\label{sec:tech_lemma}
\begin{lemma}[Recursive Least Squares Guarantee]\label{lem:recurs_LS}
In any round $\ell$, conditional on event $\mc{E}_{1,\ell} \cap \mc{E}_{2,\ell}$, for any $\theta\in\Theta$ and any $t\in [1,T_\ell]$ we have 
\begin{align*}
\left\|\theta-\hat{\theta}_{t+1}\right\|_{V_{t}}^2-\left\|\theta-\hat{\theta}_{t}\right\|_{V_{t}}^2    
&\leq  2C_{3,\ell}(y_t - x_t^{\top}\hat{\theta}_t)\leq  2C_{3,\ell}(C_{1,\ell}+1),
\end{align*}
assuming that all rewards are bounded in $[-1,1]$.
\end{lemma}

\begin{proof}
    We first consider some round $t$ and some $\theta$. Note that $\hat{\theta}_t = V_t^{-1}X_t^{\top} Y_t$. Then

\begin{align*}
    \hat{\theta}_{t+1} 
    &= (V_{t-1} + x_{t}x_{t}^{\top})^{-1}(X_{t-1}^{\top}Y_{t-1} + x_{t}y_{t})\\
    &= \left(V_{t-1}^{-1} - \frac{V_{t-1}^{-1}x_{t}x_{t}^{\top}V_{t-1}^{-1}}{1+x_{t}^{\top}V_{t-1}^{-1}x_{t}}\right)(X_{t-1}^{\top}Y_{t-1} + x_{t}y_{t})\\
    &= \hat{\theta}_t - \frac{V_{t-1}^{-1}x_{t}x_{t}^{\top}\hat{\theta}_{t}}{1+x_{t}^{\top}V_{t-1}^{-1}x_{t}} + V_{t-1}^{-1}x_{t}y_{t}- \frac{V_{t-1}^{-1}x_{t}x_{t}^{\top}V_{t-1}^{-1}x_{t}y_{t}}{1+x_{t}^{\top}V_{t-1}^{-1}x_{t}}\\
    &=\hat{\theta}_t - \frac{V_{t-1}^{-1}x_{t}x_{t}^{\top}\hat{\theta}_{t}}{1+x_{t}^{\top}V_{t-1}^{-1}x_{t}} + 
    \frac{V_{t-1}^{-1}x_{t}y_{t}(1+x_{t}^{\top}V_{t-1}^{-1}x_{t}) - x_{t}^{\top}V_{t-1}^{-1}x_{t} V_{t-1}^{-1}x_{t}y_{t}}{(1+x_{t}^{\top}V_{t-1}^{-1}x_{t})}\\
    &=\hat{\theta}_t - \frac{V_{t-1}^{-1}x_{t}x_{t}^{\top}\hat{\theta}_{t}}{1+x_{t}^{\top}V_{t-1}^{-1}x_{t}} + 
    \frac{V_{t-1}^{-1}x_{t}y_{t}}{(1+x_{t}^{\top}V_{t-1}^{-1}x_{t})}\\
    &=\hat{\theta}_t + \frac{V_{t-1}^{-1}x_{t}(y_t - x_{t}^{\top}\hat{\theta}_{t})}{1+x_{t}^{\top}V_{t-1}^{-1}x_{t}}
\end{align*}

Hence
\[\hat{\theta}_{t+1} - \hat{\theta}_t =  \frac{V_{t-1}^{-1}x_{t}}{1+x_{t}^{\top}V_{t-1}^{-1}x_{t}} (y_{t} - x_{t}^{\top}\hat{\theta}_t) \]
and
\begin{align*}
V_{t}(\hat{\theta}_{t+1} - \hat{\theta}_t) 
&=  \frac{V_{t}V_{t-1}^{-1}x_{t}}{1+x_{t}^{\top}V_{t-1}^{-1}x_{t}} (y_{t} - x_{t}^{\top}\hat{\theta}_t)\\
&=  \frac{(I + x_tx_t^{\top}V_{t-1}^{-1})x_{t}}{1+x_{t}^{\top}V_{t-1}^{-1}x_{t}} (y_{t} - x_{t}^{\top}\hat{\theta}_t)\\
&=  \frac{x_t(1 + x_t^{\top}V_{t-1}^{-1}x_{t})}{1+x_{t}^{\top}V_{t-1}^{-1}x_{t}} (y_{t} - x_{t}^{\top}\hat{\theta}_t)\\
&=  (y_{t} - x_{t}^{\top}\hat{\theta}_t)x_t\\
\end{align*}
Then 
\begin{align*}
    &\left\|\theta-\hat{\theta}_{t+1}\right\|_{V_{t}}^2-\left\|\theta-\hat{\theta}_{t}\right\|_{V_{t}}^2\\
    &= (\hat{\theta}_{t+1}-\hat{\theta}_{t})^\top V_{t}(\hat{\theta}_{t+1}+\hat{\theta}_{t}-2\theta)\\
    &= (y_{t} - x_{t}^{\top}\hat{\theta}_{t}) x_{t}^\top (\hat{\theta}_{t+1}+\hat{\theta}_{t}-2\theta)\\
    &\leq 2C_{3,\ell}(y_{t} - x_{t}^{\top}\hat{\theta}_{t})\\
    &\leq 2C_{3,\ell}(C_{1,\ell}+1)
\end{align*}
assuming all rewards are bounded by $1$.
\end{proof}


\begin{lemma}\label{lem:Laplace_approx}
For any open set $\tilde{\Theta}\subset \Theta$, we have \[\int_{\tilde{\Theta}} \exp\bigsmile{-\frac{{T_\ell}}{2}\bigsmile{\norm{\theta^*-\theta}_{A(\overline{e}_{T_\ell})}^2}}\,d\theta\doteq \exp\bigsmile{-\frac{{T_\ell}}{2}\inf{\theta\in\tilde{\Theta}}\norm{\theta^*-\theta}_{A(\overline{e}_{T_\ell})}^2}.\]
\end{lemma}
\begin{proof}
The following argument is inspired by an analogous one in Lemma 11 of \cite{russo2016simple}. Let $\iota_\ell:=\int_{\tilde{\Theta}} \exp\bigsmile{-\frac{{T_\ell}}{2}\norm{\theta^*-\theta}_{A(\overline{e}_{T_\ell})}^2}\,d\theta$ and $W_{T_\ell}(\theta):=\frac{1}{2}\norm{\theta^*-\theta}_{A(\overline{e}_{T_\ell})}^2$. Also, let $\tilde{\theta}_\ell\in\operatorname{closure}(\tilde{\Theta})$ be a point that attains the infimum, i.e. \[\tilde{\theta}_\ell:=\arg\inf{\theta\in\tilde{\Theta}}\norm{\theta^*-\theta}_{A(\overline{e}_{T_\ell})}^2.\]
Such a point must exist by the continuity of $W_{T_\ell}(\theta)$ and $\operatorname{closure}(\tilde{\Theta})$ being compact. Then, we first observe that 
\[\int_{\tilde{\Theta}} \exp\bigsmile{-\frac{{T_\ell}}{2}\norm{\theta^*-\theta}_{A(\overline{e}_{T_\ell})}^2}\,d\theta\leq \operatorname{Vol}(\tilde{\Theta})\exp\bigsmile{-\frac{T_\ell}{2}\norm{\theta^*-\tilde{\theta}_\ell}_{A(\overline{e}_{T_\ell})}^2},\]
so \[\limsup_{\ell\to\infty} \frac{1}{T_\ell}\log(\iota_\ell)+W_{T_\ell}(\tilde{\theta}_\ell)\leq 0.\]
Second, we fix some arbitrary $\epsilon>0$. Note that for any $\theta,\theta'\in\Theta$, 
\begin{align*}
    |W_{T_\ell}(\theta)-W_{T_\ell}(\theta')|&= \frac{1}{2}\bigsmile{\norm{\theta^*-\theta}_{A(\overline{e}_{T_\ell})}^2-\norm{\theta^*-\theta'}_{A(\overline{e}_{T_\ell})}^2}\\
    &= \frac{1}{2}\bigsmile{(2\theta^*-\theta-\theta')^\T A(\overline{e}_{T_\ell}) (\theta-\theta')}\\
    &= \frac{1}{2T_\ell}\sum_{t=1}^{T_\ell}\bigsmile{(2\theta^*-\theta-\theta')^\T x_tx_t^\T (\theta-\theta')}\\
    &\leq \Delta_{\max} \max_{x\in\X}x^\T (\theta-\theta')\\
    &\leq \Delta_{\max} \max_{x\in\X}\norm{x}_2 \norm{\theta-\theta'}_2\\
    &\leq L\Delta_{\max}\norm{\theta-\theta'}_2.
\end{align*}
Then, there exists $\delta>0$ such that 
\[\norm{\theta-\theta'}_2<\delta\Rightarrow |W_{T_\ell}(\theta)-W_{T_\ell}(\theta')|<\epsilon.\]
Then, we take a $\delta$-cover of $\Theta$ with $\norm{\cdot}_2$, and intersect them with $\tilde{\Theta}$, and denote the resulting cover as $\mathcal{O}$. Then, $\tilde{\theta}_\ell\in O$ for some $O\in\mathcal{O}$. Since we know that $\operatorname{Vol}(O)>0$ for any $O\in\mathcal{O}$, we have
\begin{align*}
    \iota_\ell \geq \int_O \exp \bigsmile{-{T_\ell} W_{T_\ell}(\theta)} d \theta \geq \operatorname{Vol}(O) \exp \left(-{T_\ell}\left(W_{T_\ell}\left(\tilde{\theta}_\ell\right)-\epsilon\right)\right).
\end{align*}
Taking logarithm on both sides implies that 
\begin{align*}
    \frac{1}{{T_\ell}} \log \left(\iota_\ell\right)+W_{T_\ell}\left(\tilde{\theta}_\ell\right) \geq \frac{\operatorname{Vol}(O)}{{T_\ell}}-\epsilon \rightarrow-\epsilon.
\end{align*}
Since we choose $\epsilon>0$ arbitrarily, we have 
\begin{align*}
    \liminf_{\ell\to\infty}\frac{1}{T_\ell} \log \left(\iota_\ell\right)+W_{T_\ell}\left(\tilde{\theta}_\ell\right) \geq 0.
\end{align*}
Therefore, $\lim_{\ell\to\infty}\frac{1}{T_\ell} \log \left(\iota_\ell\right)+W_{T_\ell}\left(\tilde{\theta}_\ell\right)=0$ and the statement follows.
\end{proof}

\section{Supplementary plots}\label{sec:supp_plots}

In this section, we present more supplementary plots. All experiments in the main text and supplement are run on a computing cluster with 64 AMD EPYC 7302 16-Core Processor (1500 MHz) with 1TB of RAM. For LinGame, LinGapE, and Oracle algorithms, we directly use the existing implementation from \cite{tirinzoni2022elimination} with the open-source GitHub link: https://github.com/AndreaTirinzoni/bandit-elimination. 

\begin{figure}[!htb]
    \centering
    \includegraphics[scale=0.25]{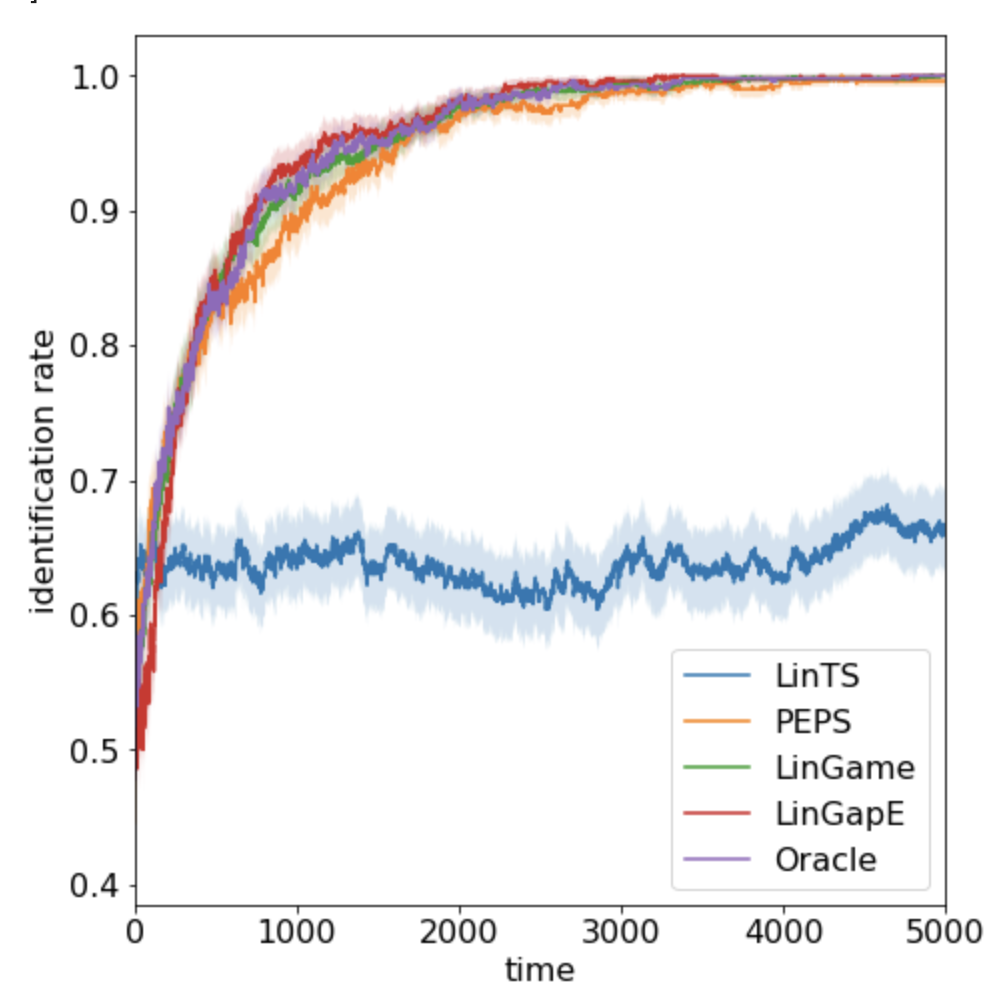}
    \includegraphics[scale=0.25]{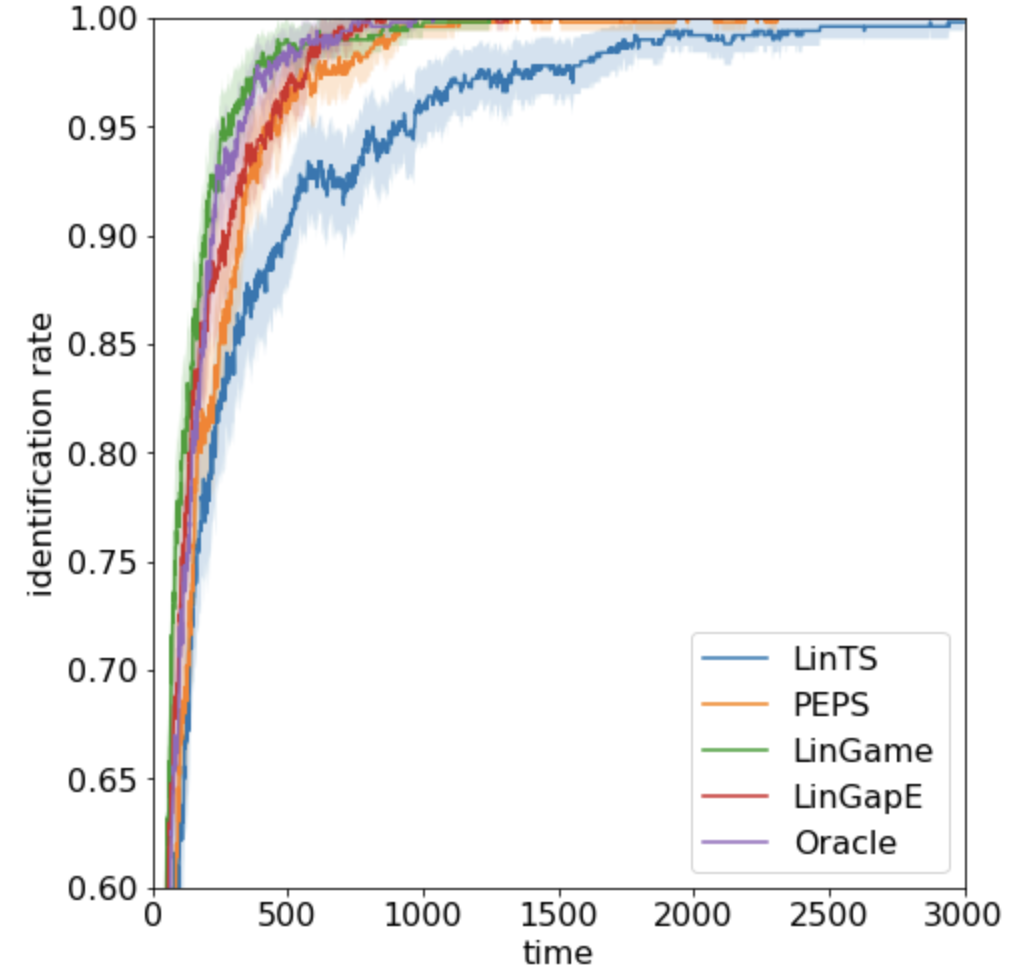}
    \includegraphics[scale=0.22]{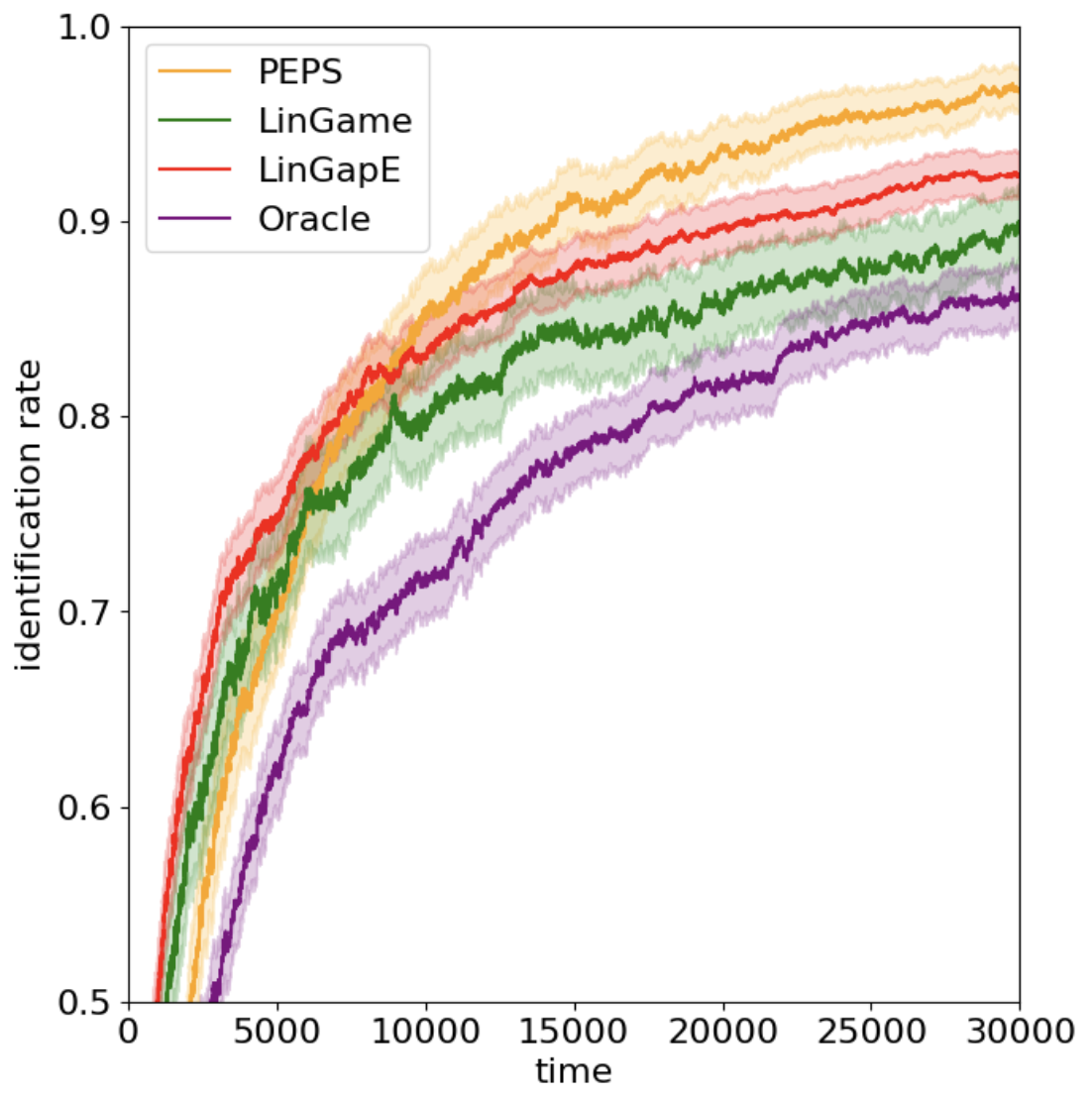}
    \caption{Best-arm identification rate for PEPS, LinGame \cite{degenne2020gamification}, LinGapE \cite{xu2018fully}, Thompson sampling, and fixed weight strategy under three instances: Soare instance with $\omega=0.1$, sphere instance with $d=6$ and $|\X|=20$, and Top-k instance with $d=12$ and $k=3$, with 500 repetitions for each instance. Confidence intervals with plus or minus two standard errors are shown.}
    \label{fig:t_full}
\end{figure}

We first provide a more extensive comparison of the performance for our algorithm by adding another benchmark \textsf{LinGapE} and an oracle strategy that pulls arms from the allocation derived from the lower bound. We run each instance for 500 repetitions and compare the following algorithms: Thompson sampling, PEPS, LinGame \cite{degenne2020gamification}, LinGapE \cite{xu2018fully}, and the fixed weights strategy where arms are pulled from the optimal allocation $\lambda^*$ obtained from $\tau^*$. We plot identification rates and arm pull probabilities for the above algorithms. Figure~\ref{fig:t_full} presents the identification rates for different algorithms respectively. Table~\ref{tab:id_rates_full} presents the number of samples needed to reach a $1-\delta$ idenfication rate for various $\delta$ values. We can see that our algorithm achieves a comparable performance to all benchmarks while beating LinTS on the Soare and Sphere instance, while outperforming all benchmark algorithms on the Top-k instance.  



\begin{table}[!htb]
\centering
\begin{tabular}{|l|lll|lll|lll|}
\hline
        & \multicolumn{3}{c|}{Soare's instance \cite{soare2014best}}                                                                             & \multicolumn{3}{c|}{Sphere}                                 & \multicolumn{3}{c|}{TopK}                                                                   \\ \hline
$\delta$   & \multicolumn{1}{l|}{0.1}                & \multicolumn{1}{l|}{0.05}               & 0.01               & \multicolumn{1}{l|}{0.1} & \multicolumn{1}{l|}{0.05} & 0.01 & \multicolumn{1}{l|}{0.2}   & \multicolumn{1}{l|}{0.1}                 & 0.05                \\ \hline
PEPS    & \multicolumn{1}{l|}{1027}               & \multicolumn{1}{l|}{1606}               & 3284               & \multicolumn{1}{l|}{294} & \multicolumn{1}{l|}{476}  & 794  & \multicolumn{1}{l|}{7326}  & \multicolumn{1}{l|}{14188}               & 22518               \\ \hline
LinGame & \multicolumn{1}{l|}{828}                & \multicolumn{1}{l|}{1500}               & 2688               & \multicolumn{1}{l|}{186} & \multicolumn{1}{l|}{282}  & 638  & \multicolumn{1}{l|}{8838}  & \multicolumn{1}{l|}{29963} & \textgreater{}30000 \\ \hline
LinGapE & \multicolumn{1}{l|}{708}                & \multicolumn{1}{l|}{1141}               & 2281               & \multicolumn{1}{l|}{316} & \multicolumn{1}{l|}{433}  & 690  & \multicolumn{1}{l|}{7096}  & \multicolumn{1}{l|}{20570}               & \textgreater{}30000 \\ \hline
Oracle  & \multicolumn{1}{l|}{766}                & \multicolumn{1}{l|}{1232}               & 2576               & \multicolumn{1}{l|}{243} & \multicolumn{1}{l|}{328}  & 473  & \multicolumn{1}{l|}{17363} & \multicolumn{1}{l|}{\textgreater{}30000} & \textgreater{}30000 \\ \hline
TS      & \multicolumn{1}{l|}{\textgreater{}5000} & \multicolumn{1}{l|}{\textgreater{}5000} & \textgreater{}5000 & \multicolumn{1}{l|}{431} & \multicolumn{1}{l|}{1046} & 2176 & \multicolumn{1}{l|}{N/A}   & \multicolumn{1}{l|}{N/A}                 & N/A                 \\ \hline
\end{tabular}
\medskip
\caption{The number of samples needed for $\P_{\theta\sim\pi_\ell}(\hat{z}_\ell=z_*)>1-\delta$ for various algorithms}
\label{tab:id_rates_full}
\end{table}

\begin{figure}[!htb]
    \centering
    \includegraphics[scale=0.3]{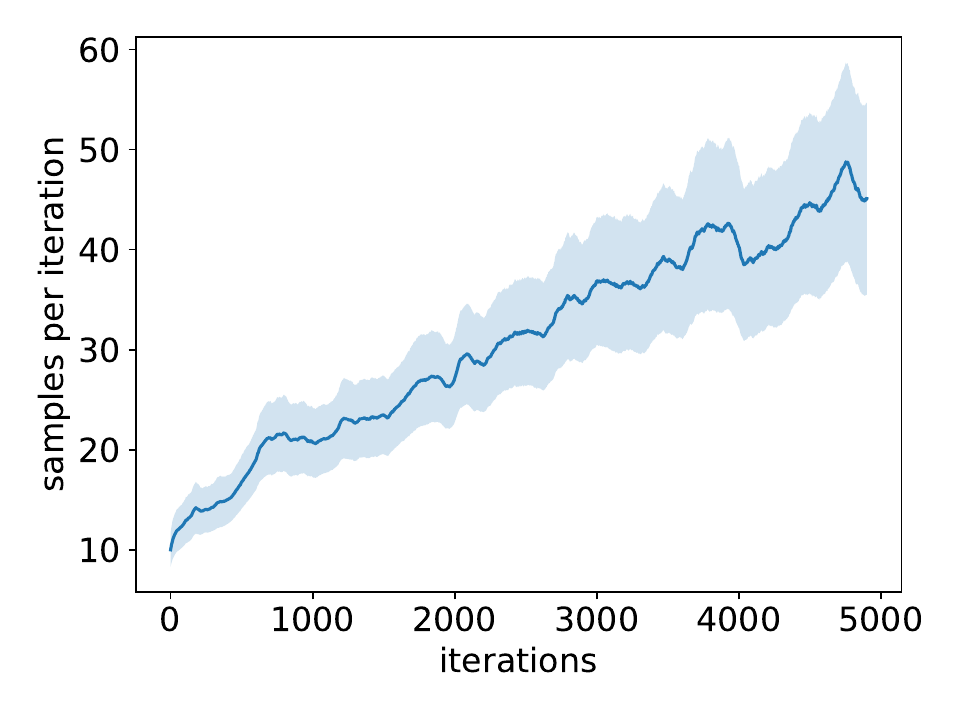}
    \includegraphics[scale=0.3]{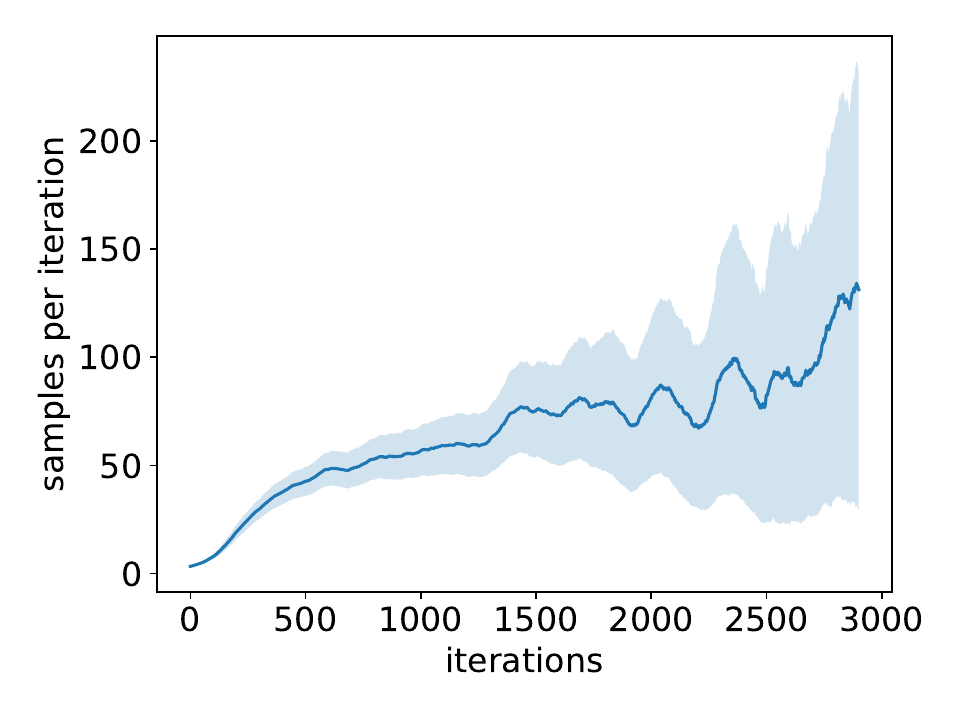}
    \includegraphics[scale=0.3]{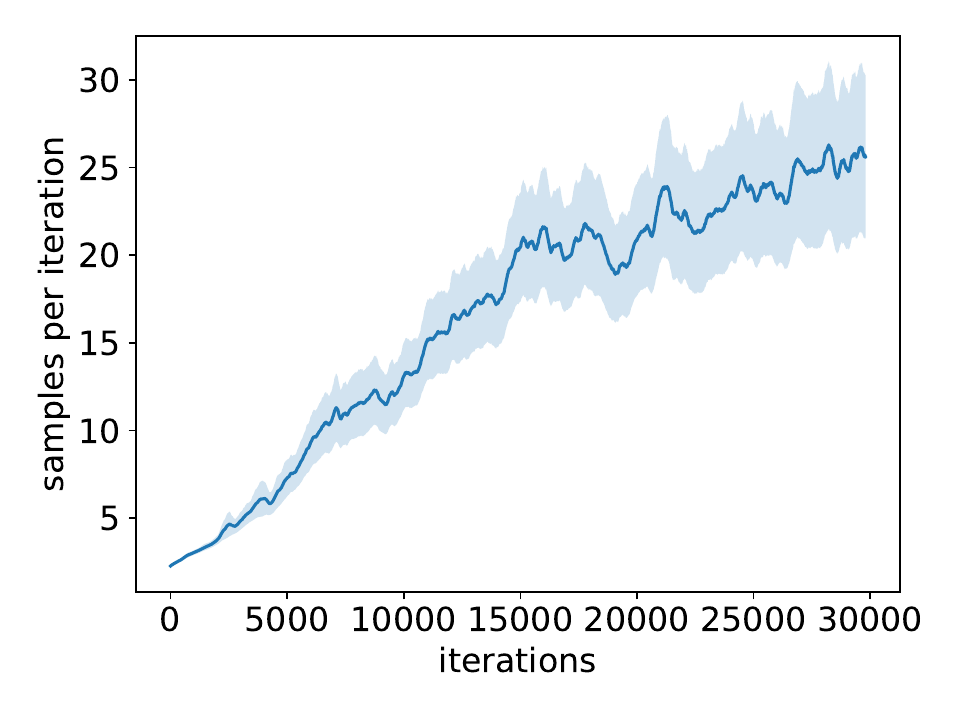}
    \caption{Average number of rejection samples taken until finding some $\theta\in\Theta_{\hat{z}_t}^c$}
    \label{fig:avg_samples_til_alt}
\end{figure}

\begin{figure}[!htb]
    \centering
    \includegraphics[scale=0.3]{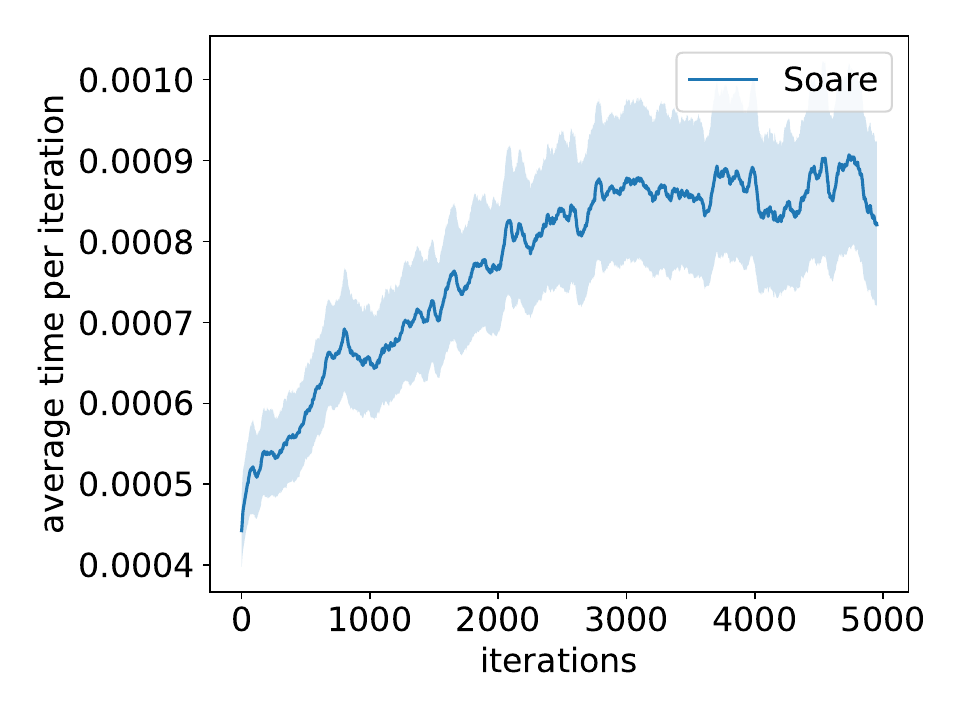}
    \includegraphics[scale=0.3]{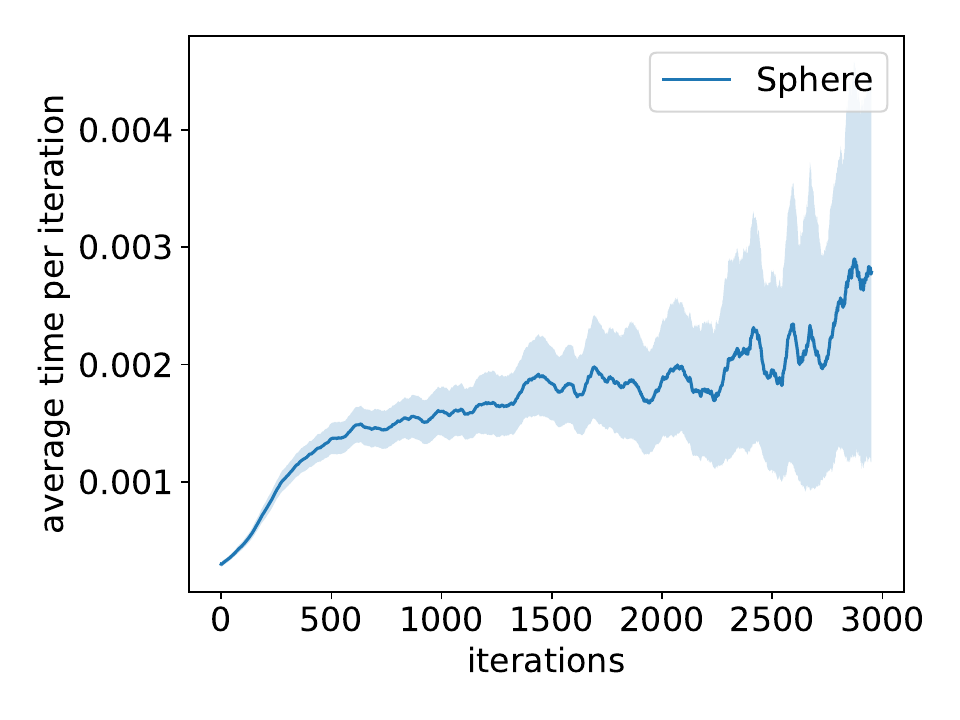}
    \includegraphics[scale=0.3]{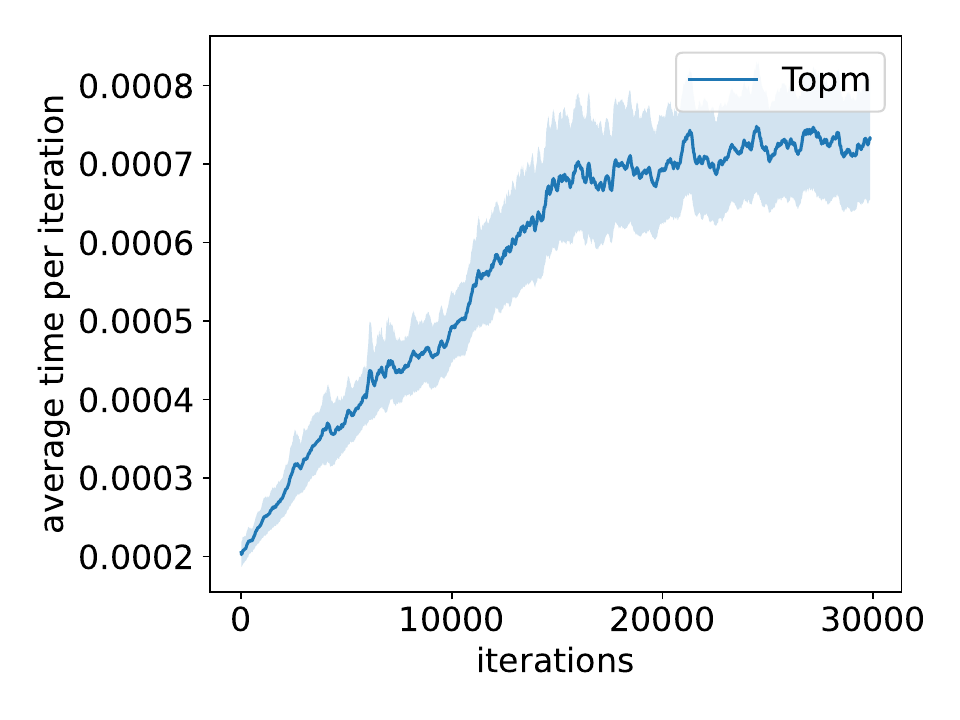}
    \caption{Average clock time per iteration for \textsf{PEPS} under three scenarios}
    \label{fig:time_per_iter}
\end{figure}

We then demonstrate that the computational cost of our algorithm is not heavy. We first plot the average number of rejection samples taken to get some $\theta\in\Theta_{\hat{z}_t}^c$ in the alternative and the running time for our algorithm to demonstrate the computation cost rejection sampling takes. Figures~\ref{fig:avg_samples_til_alt} and \ref{fig:time_per_iter} show the result. By comparing Figure~\ref{fig:avg_samples_til_alt} with Figure~\ref{fig:t_full}, we see that the number of rejection samples needed to get some $\theta\in\Theta_{\hat{z}_t}^c$ is generally less than 30 until $\delta<0.01$. This shows that the computational burden for rejection sampling is generally not large unless we have basically solved the problem. Also, we can see from Figure~\ref{fig:time_per_iter} that the running time per iteration is generally very small, which means our algorithm runs very fast. 

To make a clear comparison of the sampling part in our method with computing the best alternative step in LinGame, we implemented our algorithm, \textsf{PEPS}, in Julia and compared its clock time to existing LinGame implementations on a sphere instance with varying arm numbers, denoted as $K$. We run both algorithms for a fixed budget of 1000 iterations across 100 trials and compute the average clock time per iteration. We assessed both methods for $K=50, 200, 1000, 5000, 10000, 20000$, with results presented in milliseconds. Table~\ref{tab:avg_clock_time_comp} shows the results. We can see that our method consistently running faster than the benchmark LinGame, particularly as the number of arms increases. This distinction becomes especially significant when $K=10000$ and $K=20000$, which corresponds to the case that calculating the best alternative is expensive. Therefore, our method maintains efficiency even in scenarios when computing the alternative is really expensive.
\begin{table}[!htb]
\centering
\begin{tabular}{|l|l|l|l|l|l|l|}
\hline
        & $K=50$  & $K=200$  & $K=1000$ & $K=5000$ &
        $K=10000$ &
        $K=20000$ \\ \hline
PEPS    & 0.132 & 0.484 & 0.681 & 3.770 & 6.710 & 17.110 \\ \hline
LinGame & 0.152 & 0.596 & 3.265 & 18.610 & 46.762 & 126.683 \\ \hline
\end{tabular}
\caption{Average clock time per iteration for \textsf{PEPS} and \textsf{LinGame} under the sphere instance with $d=6$ and various number of arms $K$. Numbers are displayed in milliseconds.}
\label{tab:avg_clock_time_comp}
\end{table}

\end{document}